\PassOptionsToPackage{table}{xcolor}
\PassOptionsToPackage{inline}{enumitem}
\documentclass[sn-mathphys-num]{sn-jnl}

\usepackage{amsmath}
\usepackage{graphicx}%
\usepackage{multirow}%
\usepackage{amsmath,amssymb,amsfonts}%
\usepackage{amsthm}%
\usepackage{mathrsfs}%
\usepackage[title]{appendix}%
\usepackage{xcolor}%
\usepackage{textcomp}%
\usepackage{manyfoot}%
\usepackage{booktabs}%
\usepackage{algorithm}%
\usepackage{algorithmicx}%
\usepackage{algpseudocode}%
\usepackage{listings}%
\usepackage[inline]{enumitem}%
\usepackage{subcaption}

\theoremstyle{thmstyleone}%
\theoremstyle{thmstyletwo}%

\theoremstyle{thmstylethree}%

\newcommand{\footnoteref}[1]{\textsuperscript{\ref{#1}}}
\newcommand{\mbb}[1]{\mathbf{#1}}

\begin{document}

\title[MomentsNeRF: Leveraging Orthogonal Moments for Few-Shot Neural Rendering]{MomentsNeRF: Leveraging Orthogonal Moments for Few-Shot Neural Rendering}


\author*[1]{\fnm{Ahmad} \sur{AlMughrabi}}\email{sci.mughrabi@gmail.com}

\author[1,2]{\fnm{Ricardo} \sur{Marques}}\email{ricardo.marques@ub.edu}

\author[1,3]{\fnm{Petia} \sur{Radeva}}\email{petia.ivanova@ub.edu}

\affil*[1]{\orgdiv{Mathematics and Computer Science}, \orgname{Universitat de Barcelona}, \orgaddress{\street{Gran Via de les Corts Catalanes, 585}, \city{Barcelona}, \postcode{08007},  \country{Spain}}}

\affil[2]{\orgdiv{Computer Vision Center}, \orgname{Generalitat de Catalunya and the Universitat Autònoma de Barcelona}, \orgaddress{\street{Campus UAB, Edifici O, s/n}, \city{Barcelona}, \postcode{08193}, \country{Spain}}}

\affil[3]{\orgdiv{Institut de Neurosciències}, \orgname{University of Barcelona}, \orgaddress{\street{Passeig de la Vall d’Hebron, 171}, \city{City}, \postcode{08035}, \country{Barcelona}}}


\abstract{
We propose MomentsNeRF, a novel framework for one- and few-shot neural rendering that predicts a neural representation of a 3D scene using Orthogonal Moments. Our architecture offers a new transfer learning method to train on multi-scenes and incorporate a per-scene optimization using one or a few images at test time. Our approach is the first to successfully harness features extracted from Gabor and Zernike moments, seamlessly integrating them into the NeRF architecture. We show that MomentsNeRF performs better in synthesizing images with complex textures and shapes, achieving a significant noise reduction, artifact elimination, and completing the missing parts compared to the recent one- and few-shot neural rendering frameworks. Extensive experiments on the DTU and Shapenet datasets show that  MomentsNeRF improves the state-of-the-art by {3.39\;dB\;PSNR}, 11.1\% SSIM, 17.9\% LPIPS, and 8.3\% DISTS metrics. Moreover, it outperforms state-of-the-art performance for both novel view synthesis and single-image 3D view reconstruction. The source code is accessible at: \footnote{https://amughrabi.github.io/momentsnerf/}. 
}


\maketitle

\section{Introduction}
\label{sec:intro}
Neural fields \cite{xie2022neural} leverage the power of deep neural networks to represent 2D images or 3D scenes as continuous functions. The groundbreaking work in this area is Neural Radiance Fields (NeRF) \cite{mildenhall2021nerf}, which has been extensively researched and applied in various domains, such as novel view synthesis \cite{isaac2023exact, jain2022zero, martin2021nerf, mildenhall2021nerf}, 3D generation \cite{cai2022pix2nerf, dogaru2023sphere, jain2022zero, li2023neuralangelo, lin2023vision, poole2022dreamfusion}, deformation \cite{park2021nerfies, pumarola2021d, wang2023flow}, neural dynamic \cite{geng2023learning, jayasundara2023flexnerf, li2023dynibar}, depth-supervised methods \cite{deng2022depth, roessle2022dense, uy2023scade}, fast neural rendering \cite{chen2023mobilenerf, geng2023learning, wang2023f, muller2022instant, wan2023learning}, scene understanding \cite{ rivasmanzaneque2023, ruzzi2023gazenerf, siddiqui2023panoptic, zhang2023beyond, zhang2023nerflets}, controllable scene synthesis \cite{bao2023sine, xu2023discoscene, zhang2023ref, zheng2023editablenerf}, one-shot rendering \cite{deng2023nerdi, li20233d, pavllo2023shape, yin2023nerfinvertor}, zero-shot rendering \cite{jain2022zero, lin2023magic3d, liu2023stylerf, metzer2023latent, poole2022dreamfusion}, and more.
Despite notable advances, NeRF still requires a substantial number of input images to achieve high-fidelity scene representations. Unfortunately, it struggles to synthesize new views using one or a small number of input views, 
which significantly limits its practical utility in real-world applications.

NeRF \cite{mildenhall2021nerf} has attracted significant interest in the field of 3D computer vision and computer graphics for its remarkable capacity to produce high-quality novel views. NeRF has emerged as a promising method for learning how to represent 3D objects and scenes from a set of 2D images. During the training, NeRF can map 5D input coordinates (consisting of a 3D position within the scene and a 2D spherical direction indicating the radiance flow direction) to scene properties, such as color and density, encoded in the radiance field. Using direct volume rendering techniques, this radiance field information can be leveraged to generate new views of the learned 3D scene.
\begin{figure}[t]
    \centering\setlength{\tabcolsep}{1pt} 
    \begin{tabular}{cccc}
    \begin{subfigure}[b]{.132\linewidth}
         \centering
         \includegraphics[width=\textwidth]{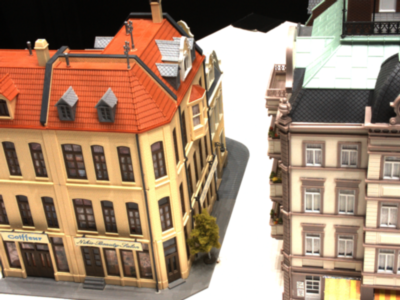}
         \\
          \includegraphics[width=\textwidth]{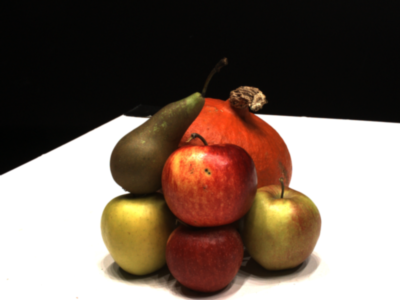}
          \\
         \includegraphics[width=\textwidth]{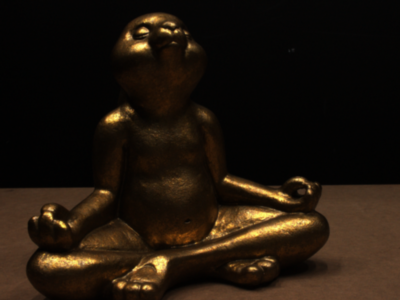}
         \caption{Input}
         \label{fig:scene55_3v_in}
     \end{subfigure}
      &
      \begin{subfigure}[b]{.28\linewidth}
      \captionsetup[figure]{font=footnotesize,labelfont=footnotesize}
         \centering
         \includegraphics[width=\textwidth]{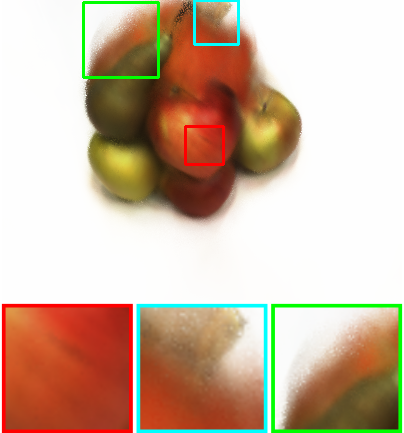}
         \caption{PixelNeRF}
         \label{fig:scene55_3v_a}
     \end{subfigure} 
     &
     \begin{subfigure}[b]{.28\linewidth}
         \centering
         \includegraphics[width=\textwidth]{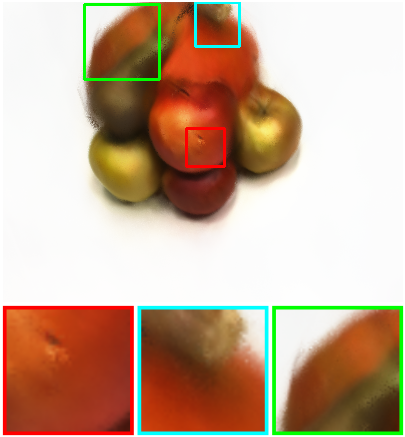}
         \caption{Ours}
         \label{fig:scene55_3v_b}
     \end{subfigure} 
     &
     \begin{subfigure}[b]{.28\linewidth}
         \centering
         \includegraphics[width=\textwidth]{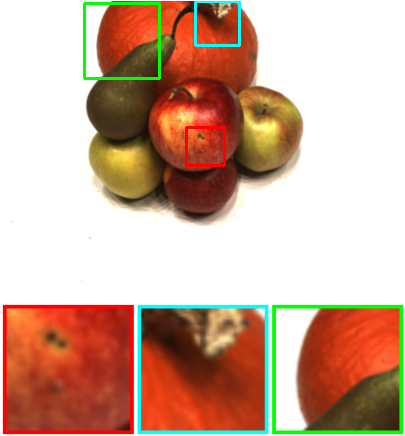}
         \caption{Reference}
         \label{fig:scene55_3v_c}
     \end{subfigure}  
\end{tabular}
        \vspace{-5px}
        \caption{Qualitative comparison on DTU dataset using 3 views settings. We show novel views rendered by PixelNeRF, Ours, and the Reference image. Our model performs better by showing the texture details, recovering artifacts, handling missing data, and better color adjustment.
        }
        \label{fig:qualitative_comparison_on_DTU}
\end{figure}
Despite all its benefits, NeRF is susceptible to over-fit to the training views and encounters difficulties in synthesizing novel views when presented with limited inputs. The issue of generating views from sparse inputs is called {\em few-shot neural rendering problem}\cite{yu2021pixelnerf}. Several NeRF-based approaches have been proposed that aim to enhance the process of view synthesis by addressing the problem 
 of \textbf{challenging input data}.  
For example, NeRF-W \cite{martin2021nerf} uses a learned low-dimensional latent space to model per-image appearance variations, such as exposure, lighting, weather, and post-processing. This allows NeRF-W to explain differences between images and provides control over the appearance of output renderings. NeRF-W models the scene as a combination of shared and image-dependent elements, dividing the scene content into "static" and "transient" components. 
RobustNeRF \cite{sabour2023robustnerf} simplifies the NeRF framework,  eliminating \textit{distractors} from the training data by identifying them as outliers in an optimization procedure. It improves scene baselines such as NeRF-W on synthetic and real-world scenes without prior knowledge of distractors. To deal with the  \textbf{noisy camera poses}, SPARF \cite{truong2023sparf} introduces a novel-view synthesis with only a few input images  and noisy camera poses. It learns and refines camera joint poses using multi-view geometry constraints. 

The final results quality of NeRF techniques directly depends on their capacity 
of choosing appropriate 
\textbf{feature extraction}.
Convolutional Neural Networks (CNNs) \cite{huang2004classification} have shown great potential in extracting and encoding visual features from images. However, their advancements have mainly focused on tasks that involve grid-like structures such as images \cite{gu2018recent}. This poses a challenge when dealing with geometry processing problems on curved surfaces, as CNNs struggle to generalize effectively \cite{gu2018recent}. The main obstacles lie in the absence of essential components like a canonical grid representation, consistent orientation, and a compatible local topology throughout the domain \cite{gu2018recent, qi2021survey}.  


While feature extraction discerns and extracts important patterns and attributes from image representation, image representation denotes having a ``meaningful representation'' of a digital image with readily apparent semantic characteristics \cite{qi2021survey}. This process is crucial in many image analysis problems.
Mathematically, a basic idea of image representation is an abstraction of projecting an image function upon a space formed by a set of specially designed basis functions, thereby producing a corresponding feature vector \cite{qi2021survey}. In the pre-CNN era, image representations and feature engineering have contributed to computer vision and pattern recognition development. According to \cite{qi2021survey}, image representation methods can be divided into four main categories:  
\begin{enumerate*}
    \item \textbf{Frequency transformations}, like Fourier Transform, Walsh Hadamard Transform, and Wavelet Transform;
    \item \textbf{Texture representation}, like Scale Invariant Feature Transform, Gradient Location and Orientation Histogram, and Local Binary Patterns;
    \item \textbf{Dimensionality reduction}, like Principal Component Analysis and Locally Linear Embedding;
    \item \textbf{Moments and moment invariants representation}, like Zernike Moments, Legendre Moments, and Polar Harmonic Transforms.
\end{enumerate*}
Image moments, such as Gabor and Zernike moments, have been proved to be very good candidates to extract representative and compact image features \cite{qi2021survey}. 

Leveraging the power of CNN-based methods to properly extract meaningful features from images,  PixelNeRF \cite{yu2021pixelnerf} and SRF \cite{chibane2021stereo} achieved robust feature extraction from sparse input views. These methods generalize NeRF to render novel scenes during test time. The ViT-NeRF \cite{lin2023vision} approach focuses on feature extraction for sparse input views using Transformers. For the remaining techniques, Pix2NeRF \cite{cai2022pix2nerf} couples $\pi$-GAN generator with the encoder to form auto-encoder without the need for 3D, multi-view, or pose-supervision. Notably, many NeRFs leverage the \textbf{latent space} as a feature bank, such as Latent-NeRF \cite{metzer2023latent}, Magic3D \cite{lin2023magic3d}, and Dreamfusion \cite{poole2022dreamfusion} to leverage the text-to-3D translation. Using a latent space has the advantage of \textbf{pose-free learning} \cite{lin2023magic3d, metzer2023latent, poole2022dreamfusion}.   



Since all the mentioned techniques, especially, one- and few-shot neural rendering techniques, rely on feature extraction, the image representation significantly impacts the quality and efficiency of feature extraction \cite{qi2021survey}. Appropriateness of the representation is critical to align extracted features with underlying image semantics \cite{qi2021survey}. Different representations vary in computational requirements and dimensionality, affecting the richness and complexity of the extracted features \cite{qi2021survey}. Moreover, most of the above-mentioned techniques rely on differentiable volumetric rendering to combine the colors and densities of discrete points along viewing rays to generate novel views.  Our hypothesis is that paying special attention on image representation can significantly increase the quality of the learned feature representations and the robustness of the geometric invariance in NeRF, especially for one- and few-shot neural rendering in sparse input views. In this paper, for first time, we propose MomentsNeRF, a new neural radiance field framework for reliable feature representation that reduces the learning complexity, integrating orthogonal moments, particularly Gabor and Zernike polynomials, to the CNN encoder, being capable in this way to produce very realistic renderings, as shown in Fig.~\ref{fig:qualitative_comparison_on_DTU}. We extensively evaluate and compare our framework on  challenging synthetic and real-world scene datasets \cite{jensen2014large} to show how MomentsNeRF outperforms current state-of-the-art.

\section{Related Work}
\label{sec:related_work}
Various approaches have been proposed to tackle the few-shot neural rendering challenge. Some methods employ transfer learning, such as PixelNerf \cite{yu2021pixelnerf}, SRF \cite{chibane2021stereo}, and MVSNeRF \cite{chen2021mvsnerf}, by pre-training on large-scale multi-view datasets and fine-tuning on a per-scene basis during testing. Other methods utilize depth-supervised learning \cite{deng2022depth, roessle2022dense}, which incorporates estimated depth as an external supervisory signal, leading to a more complex training process. Additionally, patch-based regularization methods impose various types of regularization on rendered patches, including semantic consistency \cite{jain2021putting}, geometry \cite{niemeyer2022regnerf}, and appearance regularization \cite{niemeyer2022regnerf}. However, these approaches have a computational overhead as they require additional patches during training. Moreover, FreeNeRF \cite{yang2023freenerf} introduces two types of regularization techniques. The first is frequency regularization, stabilizing the learning process and preventing catastrophic over-fitting by directly regularizing the visible frequency bands of NeRF's inputs. The second is occlusion regularization, which penalizes the density fields close to the camera that results in ``floaters," a standard failure mode in one- and few-shot neural rendering problems.

Regarding the task of improving image representation used by neural networks, the work in \cite{alekseev2019gabornet} proposed GaborNet, a neural network for image recognition, where a Gabor Layer is used as  first layer. 
 The Gabor Layer is defined  as a convolutional layer where the filters are constrained to fit Gabor functions (Gabor filters) \cite{alekseev2019gabornet}. The filters' parameters are initialized from a filter bank and updated using a standard back-propagation algorithm during the training process. Thus, the Gabor Layer aims to enhance the robustness of the learned feature representations and reduce the training complexity of the neural network. Note that the Gabor Layer can easily be integrated into any deep CNN architecture, since it is implemented using basic elements of CNNs \cite{alekseev2019gabornet, hu2020gabor}. 

In general, Gabor filters \cite{gabor1946theory} are widely utilized in computer vision due to their ability to extract spatial frequency structures from images based on a sinusoidal plane wave with specific frequency and orientation. In addition, they are suitable for tasks such as texture representation and face detection \cite{jain1997object}. 
The approach in \cite{kwolek2005face} involves using Gabor filters as a preprocessing tool to generate Gabor features, which are then used as input to a CNN. In contrast, the approach in \cite{wen2020gcsba} sets the first or second layer of the CNN as a fixed Gabor filter bank, reducing the number of trainable parameters in the network. The work in \cite{luan2018gabor} introduces Convolutional Gabor Orientation Filters as a specialized structure that modulates convolutional layers with learnable parameters by a non-learnable Gabor filter bank, although the filter parameters were not integrated into the back-propagation algorithm.

 Similarly to GaborNet, ZerNet \cite{sun2020zernet} employs Zernike polynomials \cite{zernike1934diffraction} to enable a rigorous and practical mathematical extension of CNNs to surfaces that are not limited to traditional planar or cylindrical shapes \cite{lakshminarayanan2011zernike}.
The Global-Gabor-Zernike network \cite{fathi2016new} incorporates a novel global feature descriptor that is multi-scale and rotation-invariant. This is achieved by utilizing Zernike moments on the outputs of Gabor filters. The resulting global feature is then combined with a histogram of oriented gradient, 
being an efficient local feature to improve the overall performance of the recognition system. 

Regarding the feature extraction addressed by NeRF techniques, PixelNeRF \cite{yu2021pixelnerf} uses a CNN encoder to extract the features from the given input views. 
In the  learning  phase, PixelNeRF uses the CNN features as input to feed a Deep Neural Network. 
Notably, we decided to use PixelNeRF as the basis for our experiments for three reasons: firstly, it has been efficiently used in one- and few-shot neural rendering \cite{yu2021pixelnerf}; secondly, CNN forms the PixelNeRF's core components and, thus, represents efficient image feature extraction and finally, it generalizes NeRFs across multiple scenes. 

Our main contributions  are  as follows:
\begin{enumerate}
  \item We build a novel CNN architecture for one- and few-shot Neural Rendering leveraging orthogonal moments for NeRF, employing a combination of Gabor Filters and Zernike polynomials.
  \item We introduce a new quality metric, DISTS \cite{ding2020image} to measure the details of the textures synthesized.
  \item We perform an extensive series of experiments on 3 real and synthetic image datasets to assess the effectiveness and flexibility of our framework.
  \item Our framework outperforms the state-of-the-art performance in one- and few-shots neural rendering for both novel view synthesis and single-image 3D view reconstruction.
  \item Our framework is 4 times faster than the baseline's \cite{yu2021pixelnerf} training. 
\end{enumerate}

The rest of the paper is structured as follows: We present the theoretical background in Sec.~\ref{sec:methodology}. We define our proposed methodology in Sec.~\ref{sec:momentsnerf}. A thorough set of experimental results is presented in Sec.~\ref{sec:results}. Finally, we present our conclusions and future work in Sec.~\ref{sec:conclusion}.

\section{Proposed Methodology}
\label{sec:methodology}
Our study focuses on improving the NeRF feature extraction to improve the rendered views of NeRF-based models.
\begin{figure}[htb]
\centering
\includegraphics[width=1.\linewidth]{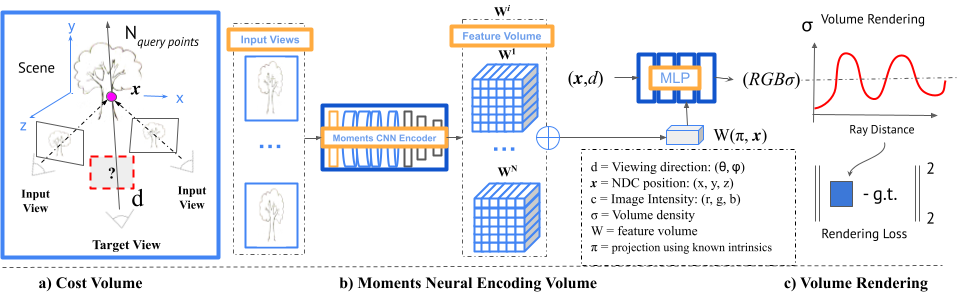}
   \caption{
   The MomentsNeRF design for the multi-view scenario involves a multi-stage process. (a) Cost Volume: given a query point \(x\) on a target camera ray with a specified view direction \(d\), an image feature corresponding to this point is extracted from the feature volume \(W^i\) through projection and interpolation operations. The Moments CNN encoder constructs the feature volume for every input during the (b) Moments Neural Encoding Volume stage. Subsequently, this feature is fed into the NeRF network MLP together with its spatial coordinates. The output of this network is represented by RGB and density values that are utilized in a (c) Volume Rendering stage, resulting in a synthesized image. 
   }
\label{fig:short}
\end{figure}
\subsection{Overview}
\label{sec:overview}



Our architecture, MomentsNeRF, aims to achieve high quality photo-realistic view synthesis in a context of one- and few-shot rendering.
To achieve it, our architecture heavily relies on the latent space backbone, comprising three distinct phases:
\begin{enumerate*}
    \item Cost Volume, 
    \item Moments Neural Encoding Volume, and 
    \item Volume Rendering.
\end{enumerate*}

Fig.~\ref{fig:short} shows a diagram representing the general overview of our proposed approach. In the first phase (Fig.~\ref{fig:short}(a)), MomentsNeRF performs ray casting and applies bilinear interpolation on pixel-wise features to extract a moment feature vector. The next phase (Fig.~\ref{fig:short}(b)) takes the pixel coordinates and viewing direction associated with the moment feature vector and feeds them into a MultiLayer Perception (MLP). This ultimately leads to the "Volume Rendering" phase (Fig.~\ref{fig:short}(c)), where the color generated by sampling MLP outcomes from multiple rays is used in the loss function against the ground truth color. 
The input of our architecture is a set of $i$ sparse input views $I$, with $1 \leq i \leq 9$ that is used to extract the orthogonal moments features for each input view.
Consequently, a neural encoding volume phase accepts a query point \(x\), a viewing direction \(d\), and the projected features from different feature volumes \(W\) that are fed to the MLP. Lastly, the Volume Rendering accepts the output of the MLP: an RGB color value $c$ and a density value $\sigma$, which results in a synthesized image for the target view.

\subsection{Preliminaries: NeRF for View Synthesis}
The \textbf{NeRF} \cite{mildenhall2021nerf} method can be conceptualized as a functional model $f$ which takes as input a continuous 5D coordinate as a given input and generates the corresponding color and density values at that location. This functional mapping is 
formalized as:
\begin{equation}
f:(\mbb{x}, \mbb{d}) \rightarrow  (\mbb{c}, \sigma),
\label{NeRF_equation}
\end{equation}
where $\mbb{x} = (x,y,z)$ is a 3D spatial location, $\mbb{d}=(\theta,\Phi)$ represents a spherical direction, $\mbb{c}=(r,g,b)$ is the obtained color of the 5D input coordinate $(\mbb{x}, \mbb{d})$ and $\sigma$ is the corresponding volume density.

Typically, NeRFs are modeled as MLPs capable of capturing the volumetric radiance information of a 3D scene in a 5D space. This 5D information, $(x,d)$ is then projected onto the image plane of a virtual camera to synthesize a new, unseen photo-realistic image. The standard approach to accomplish this involves direct volume rendering, which utilizes a virtual ray, $r$, originating from the virtual camera's position, $\mbb{o}$, and directed towards the scene. The number of rays cast is equal to the number of pixels present in the synthetic image, with each ray's direction, $d$, set to ensure that it passes through the center of the corresponding image pixel. The position along a given ray is determined by the ray parameter, $t \geq 0$, with $\mbb{r}(t) = \mbb{o} + t\,\mbb{d}$. The color of each pixel is then computed by traversing its associated ray and collecting color ($c$) and volume density ($\sigma$) values at multiple discrete positions. Finally, these collected values are integrated to determine the final color value $\hat{C}(\mbb{r})$ of each pixel. This entire process can be precisely formulated as follows \cite{mildenhall2021nerf}:
\begin{equation}
   \hat{C}(\mbb{r}) = \sum_{i=1}^{l} T_i \left(1-\exp(-\sigma_i \delta_i)\right) \, \mbb{c}_i,
 \;\; T_i = \exp \left ( -\sum_{j=1}^{i-1}\sigma_j\delta_j \right ), 
\end{equation}
where \(l\) denotes the total number of sampled points $(\mbb{x}_1, \ldots, \mbb{x}_l)$, with $\mbb{x}_i = \mbb{r}(t_i)$, evenly spaced along the ray $\mbb{r}$,   \(\delta_i = t_{i + 1} - t_i \) represents the distance between two adjacent samples along the ray.  The variables  \(c_i\) and \(\sigma_i\) denote the per-point radiance and density values, respectively, acquired by evaluating the NeRF model, $f$ (Eq.~\eqref{NeRF_equation}) at each sampled point, $\mbb{x}_i$ and with ray direction, $\mbb{d}$. Additionally, \(T_i\) represents the accumulated transmittance computed at the current sample point $\mbb{x}_i$, taking into account the density values obtained at previous points along the ray.

NeRF enables the formulation of a loss function that facilitates training by minimizing the mean-squared error (MSE) between the predicted renderings and their respective ground-truth colors. The NeRF loss function is defined as follows \cite{mildenhall2021nerf}:
\begin{equation}
\mathcal{L}_{MSE} = \frac{1}{|\mathcal{R}|}\sum_{\mbb{r} \in \mathcal{R}} \left \|  \hat{C}(\mbb{r}) - C(\mbb{r}))\right \|^2,
\end{equation}
where \(\mathcal{R}\) denotes the set of randomly sampled rays that belong to one or more training images and the ground truth and estimated color values of a given ray \(\mbb{r}\) are represented by $C(\mbb{r})$ and $\hat{C}(\mbb{r})$, respectively. 

\subsection{
One- and Few-shot Neural Rendering}
To begin the process for a given scene captured in an input image \(I\), PixelNeRF \cite{yu2021pixelnerf} initially extracts a feature volume \(W=E(I)\), where $E$ is a CNN encoder function that receives the input view image and returns the feature volumes $W$  containing the feature vectors.
Subsequently, for a point $\mbb{x}$ located along a camera ray, PixelNeRF obtains the corresponding image feature by projecting $\mbb{x}$ onto the image plane, and using 
the camera intrinsics, more precisely, the optical center and focal length of the camera, computes the corresponding image coordinates, \(\pi(\mbb{x})\). PixelNeRF then employs bilinear interpolation on the pixel-wise features to extract the feature vector, \(W^{(i)}(\pi(\mbb{x}))\). Afterwards, the image features, in conjunction with the position and view direction (both specified in the input view coordinate system), are fed into the NeRF network as follows:
\begin{equation}
    f(\gamma(\mbb{x}), d, W^{(i)}(\pi(\mbb{x}))) \rightarrow  (\mbb{c}, \sigma) \, ,
\end{equation}
where \(\gamma(.)\) is a positional encoding on \(\mbb{x}\) with  
6 exponentially increasing frequencies presented in NeRF \cite{mildenhall2021nerf}. 

 PixelNeRF independently processes the coordinates and their corresponding features in each view's coordinate frame, subsequently aggregating the results across the views within the NeRF network. It refers to the initial layers of the NeRF network as \(f_1\), which operate on the inputs in each input view space individually, and the final layers as \(f_2\), which handle the aggregated views.
 The feature volume, \(W^{(i)} = E(I^{(i)})\) is used to encode each input image. For the view-space point \(\mbb{x}^{(i)}\), the corresponding image feature is extracted from the feature volume, \(W^{(i)}\) at the projected image coordinate, \(\mbb{x}^{(i)}\). These inputs are then passed through \(f_1\) to obtain intermediate vectors:
\begin{equation}
V^{(i)} = f_1\left ( \gamma(\mbb{x}^{(i)}), \mbb{d}^{(i)}; W^{(i)} \left ( \pi(\mbb{x}^{(i)}) \right ) \right ),
\end{equation}
where the intermediate vectors \(V^{(i)}\) are aggregated using the average pooling operator \(\psi\) and fed into the final layers, denoted as \(f_2\), in order to produce the estimated density and color values:
\begin{equation}
(\sigma, \mbb{c}) = f_2\left ( \psi(V^{(1)}, ..., V^{(n)})  \right ).
\label{eq:sigma_c}
\end{equation}
Hence, in the single-view special case, Eq.~\eqref{eq:sigma_c} with \(f = f_2 \circ f_1\) constructs the view space as the world space. 

\section{Our Proposal: MomentsNeRF}
\label{sec:momentsnerf}
Our framework aims to propose a rich and expressive feature volume $W$ for one- and few-shot neural rendering. 
In this section, we present the formal definition of Zernike polynomials,  and introduce the discretization of the continuous form of the Zernike convolution to define Zernike convolutional layers \cite{lakshminarayanan2011zernike, sun2020zernet, zernike1934diffraction}. Secondly, we briefly define Gabor filters, and apply discrete 2D Gabor functions to define the Gabor convolutional layers. Finally, we present our feature volumetric representation by introducing Moments Encoder based on  linearly linked Zernike and Gabor convolutional layers. 

\subsection{Zernike Polynomials}
Zernike polynomials are a unique orthogonal function set defined over the unit circle. 
Zernike polynomials have a nice property forming a complete set, meaning they can represent any complex shape if enough terms are used in the expansion. Let us consider an arbitrary radial function, $f_z(\bar r, \bar \theta)$, where $\bar r$ is the radial distance such that $0 \leq \bar r \leq 1$, and $0 \leq \bar \theta \leq 2\pi$ is the azimuthal angle. Then, the expansion of $f_z(\bar r, \bar \theta)$ in terms of Zernike polynomials is given by \cite{sun2020zernet}: 
\begin{equation}\label{eq:input_representation}
f_{z}(\bar r, \bar\theta) = \sum_{n=0}^{\infty} \sum_{m=-n}^{n} \widetilde{\alpha}_{n}^{m} Z_n^m(\bar r, \bar \theta),
\end{equation}
where $Z_n^m$ are the Zernike polynomials, $\tilde{\alpha}_n^m$ are the corresponding coefficients, and $n$ and $m$ are positive integers with $n \geq m \geq 0$.
Zernike polynomials represent orthogonal basis functions. Consequently, Zernike moments are able to represent, without redundancies, the features of a given image. This property is one of the reasons which makes Zernike polynomials appealing basis functions for our approach.

The  formal definition of Zernike polynomials is split into odd and even sequences referred to as $Z_{n}^{m}(\bar r,\bar\theta)$ and $Z_{n}^{-m}(\bar r, \bar \theta)$, respectively \cite{sun2020zernet}:
\begin{align}
Z_{n}^{m}(\bar r, \bar \theta) = R_{n}^{m}(\bar r) \cos(m\bar\theta), Z_{n}^{-m}(\bar r, \bar\theta) = R_{n}^{m}(\bar r) \sin(m\bar\theta),
\end{align}
where the Zernike radial polynomial $R_n^m$ is defined as \cite{sun2020zernet}:
\begin{equation}
R_{n}^{m}(\bar r)=
\left\{
\begin{matrix}
 R_{e}(n, m, \bar r), & \text{if} \: n-m \: \text{ is even} \\ 
0, & \text{otherwise}.
\end{matrix}
\right.
\end{equation}
$R_{e}(n,m, \bar r)$ is formulated as \cite{sun2020zernet}:
\begin{equation}
    R_e(n, m, \bar r) = \sum_{k=0}^{\frac{n-m}{2}} \frac{(-1)^k(n-k)!}{k!(\frac{n+m}{2} - k)!(\frac{n-m}{2}-k)!}\bar r^{n-2k},
\end{equation}
while \(R_{n}^{m}(\bar r)\) have a value for an even number of \(n-m\) and $0$ for an odd number of \(n-m\), a special value is \(R_{n}^{m}(1) = 1\) \cite{sun2020zernet}. 
Fig.~\ref{fig:zernike_even_odd_ploy_3d} shows some examples of Zernike polynomials with their classical names in 2D and 3D representations \footnote{Additional illustrations can be found in the supplementary material \label{illu}}. 
\newcommand{\w}{.15}
\begin{figure}[htb]
     \centering\setlength{\tabcolsep}{1pt}
    \begin{tabular}{ccccc}
      \begin{subfigure}[t]{\w\linewidth}
         \centering
         \includegraphics[width=\textwidth]{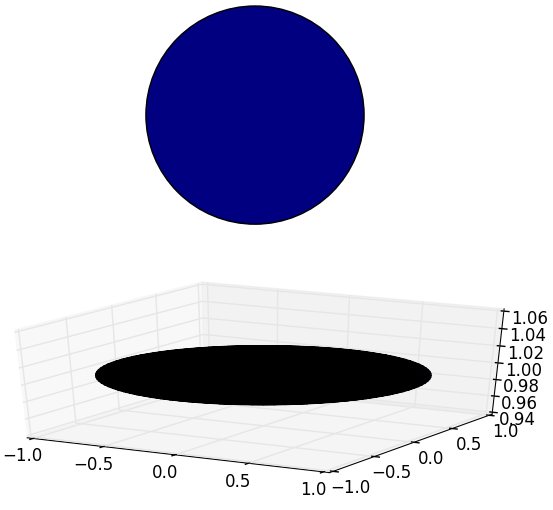}    
         \caption{\(Z_{0}^{0}\)(Even)\\Piston}
     \end{subfigure} 
     &
     \begin{subfigure}[t]{\w\linewidth}
         \centering
         \includegraphics[width=\textwidth]{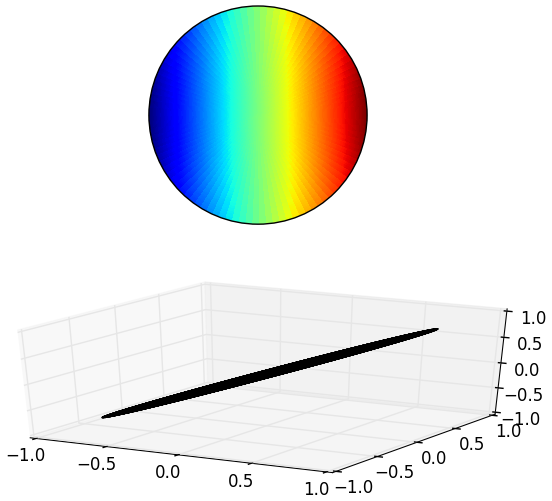}  
         \caption{ \(Z_{1}^{1} (Even)\)\\Vertical tilt}
     \end{subfigure} 
     &
     \begin{subfigure}[t]{\w\linewidth}
         \centering
         \includegraphics[width=\textwidth]{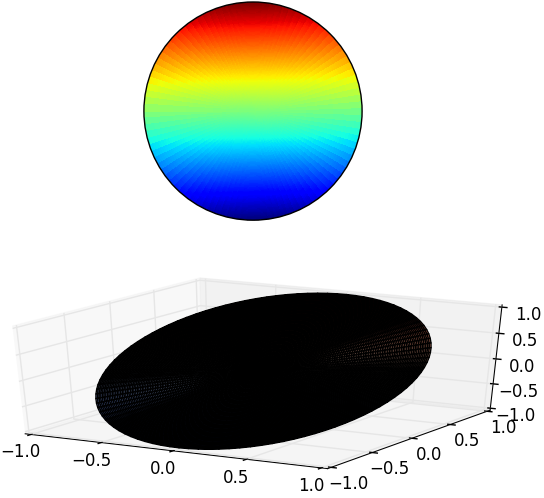}   
         \caption{\(Z_{1}^{1} (Odd)\)\\Horizontal tilt}
     \end{subfigure} 
     &
    ...
     &
     \begin{subfigure}[t]{\w\linewidth}
         \centering
         \includegraphics[width=\textwidth]{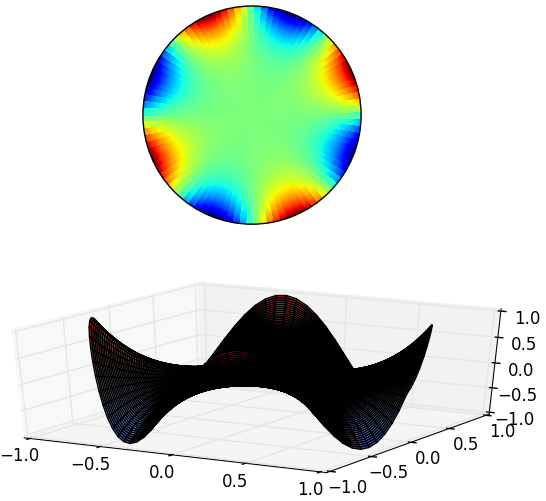}
         \setcounter{subfigure}{16}%
         \caption{\(Z_{4}^{4} (Odd)\)\\Vertical quadrafoil}
     \end{subfigure} 
\end{tabular}
        \caption{Even (on the left) and Odd (on the right) 2D (up) \& 3D (down) plots for four Zernike polynomials, also labeled with their classical names. The complete set of Zernike Polynomials (i.e., 15 Polynomials) can be found in the supplements.}

        
        \label{fig:zernike_even_odd_ploy_3d}
\end{figure}

\subsection{Convolutional Zernike Layer}
In traditional convolution, a filter is slid over an image in two-dimensional Cartesian space, and the degree of match between the filter and the image is measured. To leverage this concept to the manifold 
setting by similarly defining convolution, the filter tangent is kept to the manifold while sliding it around and locally parameterizing it using the tangent space to perform the usual convolution operation\footnoteref{illu}. Thus, let $g: T_pM \rightarrow \mathbb{R}$ be our ‘filter’, where $T_p$ is the tangent space and $I: M \rightarrow \mathbb{R}$ be our ‘image’ defined on a manifold $M$, $p$ is a point from the manifold $p \in M$, and $\mathbb{R}$ defines the radius of the local neighborhood around $p$, respectively \cite{sun2020zernet}: 
\begin{equation}
\label{eqn:conv}
(I * g)(p) = \sum_{i} \alpha_{I}^{i}\alpha_{g}^{i},
\end{equation}
which  denotes a vector dot product between the Zernike coefficient vectors, $\alpha_{g}$ and $\alpha_{I}$ at point $p$. 
Interestingly, the pixels mapping relies on the calculation method for unit disk on a polar coordinates (\(\bar\theta, \bar r\)) system, commonly known as polar pixel tiling \cite{camacho2014high, mukundan1995fast}.

\subsection{Gabor Filter and Gabor Convolutional Layer}
The 2D Gabor function was originally introduced by \cite{daugman1987image} and since then it has been extensively used in the analysis of visual information \cite{alekseev2019gabornet, fathi2016new, luan2018gabor}. In the context of the Fourier transform, the 2D Gabor function corresponds to the short-time Fourier function with a Gaussian window \cite{alekseev2019gabornet, hu2020gabor}:
\begin{equation}
G\left(x, y; \Omega\right)=
e^{-\frac{x’^2+\gamma^2y’^2}{2\sigma^2}} e^{i\left(2\mathrm\pi\frac{x’}{\lambda}+\psi\right)},
\end{equation}
where $x$ and $y$ are the 2D spatial coordinates and $\Omega = (\lambda,\theta,\psi,\sigma,\gamma)$, 
$\lambda$ is the sine function wavelength, $\psi$ is the phase shift, $\sigma$ is the Gaussian function standard deviation, $\gamma$ is the space aspect ratio, $x'=x\hspace{0.25em}\cos\hspace{0.25em}\theta+y\hspace{0.25em}\sin\hspace{0.25em}\theta$, $y'=-x\hspace{0.25em}\sin\hspace{0.25em}\theta+y\hspace{0.25em}\cos\hspace{0.25em}\theta$, and $\theta$ is the direction of the Gabor
kernel function, $0\leq \theta \leq 2\pi$. Its functional expression  represents the 2D Gabor function real part, which has stronger filtering ability and feature extraction ability, as compared with the imaginary part, respectively  \cite{alekseev2019gabornet, hu2020gabor}:
\begin{equation}
G_{\mathrm{real}}\left(x, y; \Omega\right)=
e^{-\frac{x’^2+\gamma^2y’^2}{2\sigma^2}} \cos\left(2\mathrm\pi\frac{x’}{\lambda}+\psi\right).
\end{equation}

Numerous experiments have revealed that the shape of the Gabor filter's convolution kernel plays a crucial role in feature extraction. Specifically, when the scale, direction, central position, phase, and structure type of both the Gabor filter and image content are aligned, the optimal feature response can be achieved. To extract complex and rich feature information, the variables $k_1$ to $k_5$ are introduced to adjust the real part of the Gabor convolution kernel. As a result, the following equation is obtained \cite{alekseev2019gabornet, hu2020gabor}: 
\begin{equation}
G_{comp}\left(x, y; \Omega\right)=
e^{-\frac{x'^2+\gamma^2y’^2}{2\sigma^2}}\cos\left(2\mathrm\pi\frac{\left(k_1\cdot x’^{k_2}+y’^{k_3}+k_4\right)^{ k_5}}{\lambda}+\psi\right),
\end{equation} 
where $K$ is represented by 5 learnable by the neural network parameters, $K = (k_1,k_2,k_3,k_4,k_5)$; 
 $k_2$, $k_3$, and $k_5$  determine the shape of Gabor filter, and parameters  $k_1$ and $k_4$  determine the direction and phase of the convolutional kernel. 

\begin{figure}[tbh]
    \centering
    \includegraphics[width=1.\linewidth]{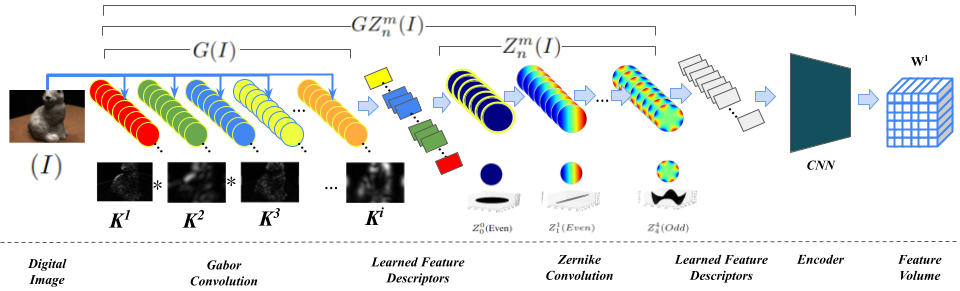}
    \caption{
    Detailed illustration of the Moments CNN Encoder. }
    \label{fig:moments-cnn-encoder}
\end{figure}

We apply the filter $G_{comp}$ on the red, green, and blue channels of the color image that leads us to the Gabor convolutional layer output, $G_r$, $G_g$, and $G_b$.  
Note that the final output of the Gabor convolutional layer is given by the sum of the three color output components: $G (I) = G_r + G_g + G_b$ (see Fig.~\ref{fig:moments-cnn-encoder}). 

\subsection{Moments CNN Encoder
}
To define the MomentsNeRF, we first extend the Gabor filter through a convolution. We define a Gabor convolutional layer. We use a similar notation to that of Eq.~\eqref{eqn:conv}, $\alpha_{I}$ and $\alpha_{g}$ denote the image and Gabor filter coefficients, where \(g\) is our filter and \(I\) is our image.
Interestingly, by applying the defined Zernike convolution on the filtered image of Gabor convolution, we can obtain the Gabor and Zernike convolutional layers: 
\begin{equation}
    GZ_{n}^{m}(I) = Z_{n}^{m}(G(I)),
\end{equation}
where $Z_{n}^{m}$ is the Zernike convolutional and $G$ denotes to Gabor convolutional. We use Gabor and Zernike convolutional layers as a feature extractor of our MomentsNeRF in our One- and Few-shot Neural Rendering architecture; furthermore, the encoder $W^{(i)} = GZ_{n}^{m}(I)$ to build and generate the feature volume $W^{(i)}$, 
for the given input image $I$.


Fig.~\ref{fig:moments-cnn-encoder} shows a detailed representation of our Moments CNN encoder. At the beginning of the process, a digital image is fed into the encoder, and its features are extracted using Gabor convolutional layers. The resulting feature descriptors are then given as input into Zernike convolutional layers, which continues the feature extraction across the 15 Zernike Layers (see Fig. \ref{fig:zernike_even_odd_ploy_3d}). Finally, the encoded features are processed through a CNN encoder, resulting in a Feature Volume $W$. Then the Feature Volume $W$ is substituted to the main design, as shown in Fig. ~\ref{fig:short}. 
We additionally change the MLP activation function to use Super-Gaussian Activation Function (AF) \cite{ramasinghe2022beyond} over ReLU. Moreover, we tweak the Positional Encoding (PE) similar to NeRF \cite{mildenhall2021nerf} implementation \footnoteref{sup_desc}.


\section{Experimental Results}
\label{sec:results}
We followed the evaluation protocol of \cite{yu2021pixelnerf} on the same datasets. More precisely, our framework is evaluated on three public datasets: a multifaceted dataset, DTU MVS \cite{jensen2014large} and two subsets for ShapeNet \cite{chang2015shapenet}\footnote{For a detailed description, kindly refer to the supplementary material. \label{sup_desc}}. Following the validation methodology of NeRFs, we apply four error metrics: PSNR, SSIM \cite{wang2004image}, LPIP \cite{zhang2018unreasonable}, and DISTS \cite{ding2020image}. 

\begin{table}[t]
\setlength{\tabcolsep}{1pt}
\scriptsize
\centering
\caption{A quantitative comparison of our model with the SOTA on DTU dataset. Best results in red, second-best in orange, and third-best in yellow. $\dagger$ denotes to PixelNeRF with Resnet152.}

\label{table:avg_comparasion}
\begin{tabular}{@{\extracolsep\fill}ccccccccccccccccc}
\toprule
                 & \multicolumn{4}{{@{}c@{}}}{1 View}    & \multicolumn{4}{{@{}c@{}}}{3 Views}    & \multicolumn{4}{{@{}c@{}}}{6 Views}    & \multicolumn{4}{{@{}c@{}}}{9 Views}  
                 \\\cmidrule{2-5} \cmidrule{6-9} \cmidrule{10-13} \cmidrule{14-17}
                 
                  & \tiny{PSNR↑}   & \tiny{SSIM↑}  & \tiny{LPIPS↓} & \tiny{DISTS↓}  & \tiny{PSNR↑}   & \tiny{SSIM↑}  & \tiny{LPIPS↓} & \tiny{DISTS↓} & \tiny{PSNR↑}   & \tiny{SSIM↑}  & \tiny{LPIPS↓} & \tiny{DISTS↓} & \tiny{PSNR↑}   & \tiny{SSIM↑}  & \tiny{LPIPS↓} & \tiny{DISTS↓} \\
\midrule
\cite{chibane2021stereo} & - & - & - & -  & 16.06  & 0.55  & 0.431 & -     & 16.060 & 0.657 & 0.353 & -     & 19.970 & 0.678 & 0.325 & -     \\
\cite{chen2021mvsnerf}& - & - & - & -  & 16.26  & 0.601 & 0.384 & -     & 18.220 & 0.694 & 0.319 & -     & 20.320 & 0.736 & 0.278 & -     \\
\cite{barron2021mip} & - & - & - & - & 7.640  & 0.227 & 0.655 & -     & 14.330 & 0.568 & 0.394 & -     & 20.710 & 0.799 & 0.209 & -     \\
\cite{jain2021putting} & - & - & - & - & 10.01  & 0.354 & 0.574 & -     & 18.700 & 0.668 & 0.336 & -     & 22.160 & 0.740 & 0.277 & -     \\
\cite{niemeyer2022regnerf} & - & - & - & -  & 15.33  & 0.621 & 0.341 & -     & 19.100 & 0.757 & 0.233 & -  & 22.300 & 0.823 & 0.184 & -     \\
\cite{yang2023freenerf} & - & - & - & - & 18.02  & \cellcolor{yellow!25} 0.68  & \cellcolor{yellow!25} 0.318 & -     & \cellcolor{yellow!25} 22.390 & \cellcolor{yellow!25} 0.779 & \cellcolor{yellow!25} 0.24  & -  & 24.200 & \cellcolor{yellow!25} 0.833 & \cellcolor{yellow!25} 0.187 & -     \\
\cite{yu2021pixelnerf} & \cellcolor{yellow!25}15.311 & \cellcolor{orange!25}0.523 & \cellcolor{yellow!25}0.555 & \cellcolor{yellow!25}0.339 & \cellcolor{yellow!25} 18.990 & 0.678 & 0.395 & \cellcolor{yellow!25} 0.251 & 19.962 & 0.713 & 0.347 & \cellcolor{yellow!25} 0.233 & \cellcolor{yellow!25} 20.471 & 0.734 & 0.307 & \cellcolor{yellow!25} 0.210 \\
\cite{yu2021pixelnerf}$^\dagger$ & \cellcolor{orange!25}17.284 & \cellcolor{yellow!25}0.422 &\cellcolor{orange!25}0.338 & \cellcolor{orange!25}0.296 & \cellcolor{orange!25}24.114 & \cellcolor{orange!25}0.824 & \cellcolor{orange!25}0.141 &\cellcolor{orange!25}0.167 &\cellcolor{orange!25}24.402 &\cellcolor{orange!25}0.847 &\cellcolor{orange!25}0.129 &\cellcolor{red!25}0.171 &\cellcolor{orange!25}24.776 &\cellcolor{orange!25}0.860 &\cellcolor{orange!25}0.125 &\cellcolor{red!25}0.167 \\
\footnotesize{Ours}   & \cellcolor{red!25}21.991 & \cellcolor{red!25}0.747 & \cellcolor{red!25}0.175 & \cellcolor{red!25}0.175 & \cellcolor{red!25} 24.507 & \cellcolor{red!25}0.847 & \cellcolor{red!25}0.131 & \cellcolor{red!25}0.167 & \cellcolor{red!25}25.002 & \cellcolor{red!25} 0.867 & \cellcolor{red!25}0.125 & \cellcolor{orange!25}0.177 & \cellcolor{red!25}25.320 & \cellcolor{red!25}0.875 & \cellcolor{red!25}0.118 & \cellcolor{orange!25}0.172 \\
\botrule
\end{tabular}
\end{table}

\subsection{Implementation settings}
\label{sec:resource_limitation}
We used two GPUs, GeForce GTX 1080 Ti/12G and RTX 3090/24G, to run the experiments. For MomentsNeRF encoders, we use Resnet34 backbone ImageNet. All encoders end with four pooling layers; before them, features are extracted, upsampled using bilinear interpolation, and concatenated to form latent vectors of size 512 aligned to each pixel. The positional encoding has 6 exponentially increasing frequencies. Our model used \(3\) as a batch size, a learning rate \(1 \times 10^{-4}\), and 600k iterations.

\subsection{MomentsNeRF Results}
\label{sec:results_mipnerf_360}
\begin{figure}[htb]
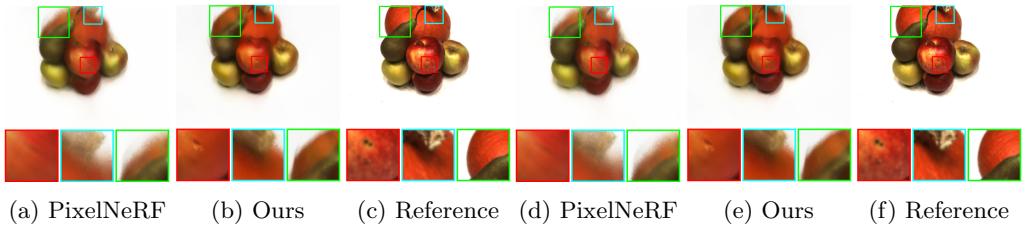

    \centering\setlength{\tabcolsep}{1pt}
      \renewcommand{\arraystretch}{.5}
      \begin{tabular}{cccccc}
         \begin{subfigure}[t]{.166\linewidth}
         \centering
         \includegraphics[width=\textwidth]{comparisons_scan63_6v_fig_sample_a.png} 
         \caption{PixelNeRF}
         \end{subfigure} 
         &
         \begin{subfigure}[t]{.166\linewidth}
             \centering
             \includegraphics[width=\textwidth]{comparisons_scan63_6v_fig_sample_b.png}
             \caption{Ours}
         \end{subfigure} 
         &
         \begin{subfigure}[t]{.166\linewidth}
             \centering
             \includegraphics[width=\textwidth]{comparisons_scan63_6v_fig_sample_c.png}
             \caption{Reference}
         \end{subfigure} 
         &
     \begin{subfigure}[t]{.166\linewidth}
         \centering
         \includegraphics[width=\textwidth]{comparisons_scan63_6v_fig_sample_a.png}
         \caption{PixelNeRF}%
     \end{subfigure} 
     &
     \begin{subfigure}[t]{.166\linewidth}
         \centering
         \includegraphics[width=\textwidth]{comparisons_scan63_6v_fig_sample_b.png}
         \caption{Ours}%
     \end{subfigure} 
     &
     \begin{subfigure}[t]{.166\linewidth}
         \centering
         \includegraphics[width=\textwidth]{comparisons_scan63_6v_fig_sample_c.png}
         \caption{Reference}%
     \end{subfigure} 
\end{tabular}
\caption{Comparison to the Reference and PixelNeRF for 1 \& 9 views settings on the DTU dataset. 
}
\label{fig:1_9v_settings_dtu_results}
\end{figure}

\begin{figure}[htb]
     \centering\setlength{\tabcolsep}{1pt}
     \small
    \begin{tabular}{@{\extracolsep\fill}cccccccccccccccc}
\toprule
    \multicolumn{8}{@{}c@{}}{One-shot} & \multicolumn{8}{@{}c@{}}{Two-shot}
    \\\cmidrule{2-8}\cmidrule{9-16}
       \multicolumn{2}{@{}c@{}}{Input} & \multicolumn{2}{@{}c@{}}{PixelNeRF} &  \multicolumn{2}{@{}c@{}}{Ours} & \multicolumn{2}{@{}c@{}}{Reference} &       \multicolumn{2}{@{}c@{}}{Input} & \multicolumn{2}{c}{PixelNeRF} &  \multicolumn{2}{c}{Ours} & \multicolumn{2}{c}{Reference}       
        \\
        \midrule
        \multicolumn{2}{c}{
           \begin{subfigure}[t]{0.06\linewidth}
             \centering
             \includegraphics[trim={1cm 0.75cm 1cm 1cm},clip,width=0.9\textwidth]{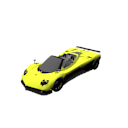}
            \end{subfigure}       
        }
     &
     \begin{subfigure}[t]{0.06\linewidth}
         \centering
         \includegraphics[trim={1cm 0.75cm 1cm 1cm},clip,width=0.9\textwidth]{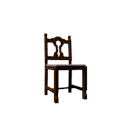}
     \end{subfigure} 
     &
     \begin{subfigure}[t]{0.06\linewidth}
         \centering
         \includegraphics[trim={1cm 0.75cm 1cm 1cm},clip,width=0.9\textwidth]{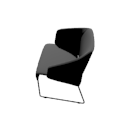}
     \end{subfigure} 
     &
      \begin{subfigure}[t]{0.06\linewidth}
         \centering
         \includegraphics[trim={1cm 0.75cm 1cm 1cm},clip,width=0.9\textwidth]{17e916fc863540ee3def89b32cef8e45_gt_000000.png}
     \end{subfigure}
     &
     \begin{subfigure}[t]{0.06\linewidth}
         \centering
         \includegraphics[trim={1cm 0.75cm 1cm 1cm},clip,width=0.9\textwidth]{174019d47144a9462fa77440dcb93214_gt_000131.png}
     \end{subfigure}
     &
     \begin{subfigure}[t]{0.06\linewidth}
         \centering
         \includegraphics[trim={1cm 0.75cm 1cm 1cm},clip,width=0.9\textwidth]{17e916fc863540ee3def89b32cef8e45_gt_000000.png}
     \end{subfigure} 
     &
     \begin{subfigure}[t]{0.06\linewidth}
         \centering
         \includegraphics[trim={1cm 0.75cm 1cm 1cm},clip,width=0.9\textwidth]{174019d47144a9462fa77440dcb93214_gt_000131.png}
     \end{subfigure} 
    &
          \begin{subfigure}[t]{0.06\linewidth}
         \centering
         \includegraphics[trim={1cm 0.65cm 1cm 1cm},clip,width=1\textwidth]{144d0880f61d813ef7b860bd772a37_gt_000064.png}
     \end{subfigure} 
     &
    \begin{subfigure}[t]{0.06\linewidth}
         \centering
         \includegraphics[trim={1cm 0.75cm 1cm 1cm},clip,width=1\textwidth]{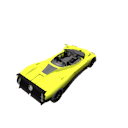}
     \end{subfigure} 
     &
     \begin{subfigure}[t]{0.06\linewidth}
         \centering
         \includegraphics[trim={1cm 0.75cm 1cm 1cm},clip,width=1\textwidth]{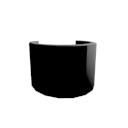}
     \end{subfigure} 
     &
     \begin{subfigure}[t]{0.06\linewidth}
         \centering
         \includegraphics[trim={1cm 0.75cm 1cm 1cm},clip,width=1\textwidth]{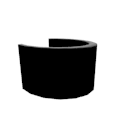}
     \end{subfigure} 
     &
    \begin{subfigure}[t]{0.06\linewidth}
         \centering
         \includegraphics[trim={1cm 0.75cm 1cm 1cm},clip,width=1\textwidth]{103b75dfd146976563ed57e35c972b4b_gt_000037.png}
     \end{subfigure} 
     &
     \begin{subfigure}[t]{0.06\linewidth}
         \centering
         \includegraphics[trim={1cm 0.75cm 1cm 1cm},clip,width=1\textwidth]{103b75dfd146976563ed57e35c972b4b_gt_000048.png}
     \end{subfigure} 
     &
      \begin{subfigure}[t]{0.06\linewidth}
         \centering
         \includegraphics[trim={1cm 0.75cm 1cm 1cm},clip,width=1\textwidth]{103b75dfd146976563ed57e35c972b4b_gt_000037.png}
     \end{subfigure}
     &
     \begin{subfigure}[t]{0.06\linewidth}
         \centering
         \includegraphics[trim={1cm 0.75cm 1cm 1cm},clip,width=1\textwidth]{103b75dfd146976563ed57e35c972b4b_gt_000048.png}
     \end{subfigure}
     \\
    \multicolumn{2}{c}{
        \begin{subfigure}[t]{0.06\linewidth}
             \centering
             \includegraphics[trim={1cm 1cm 1cm 1cm},clip,width=1\textwidth]{144d0880f61d813ef7b860bd772a37_gt_000064.png}
         \end{subfigure} 
     }
     &
     \begin{subfigure}[t]{0.06\linewidth}
         \centering
         \includegraphics[trim={1cm 1cm 1cm 1cm},clip,width=1\textwidth]{17e916fc863540ee3def89b32cef8e45_gt_000000.png}
     \end{subfigure} 
     &
     \begin{subfigure}[t]{0.06\linewidth}
         \centering
         \includegraphics[trim={1cm 1cm 1cm 1cm},clip,width=1\textwidth]{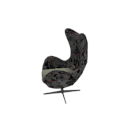}
     \end{subfigure} 
     &
      \begin{subfigure}[t]{0.06\linewidth}
         \centering
         \includegraphics[trim={1cm 1cm 1cm 1cm},clip,width=1\textwidth]{17e916fc863540ee3def89b32cef8e45_gt_000000.png}
     \end{subfigure}
     &
     \begin{subfigure}[t]{0.06\linewidth}
         \centering
         \includegraphics[trim={1cm 1cm 1cm 1cm},clip,width=1\textwidth]{19ce953da9aa8065d747a43c11e738e9_gt_000056.png}
     \end{subfigure}
     &
     \begin{subfigure}[t]{0.06\linewidth}
         \centering
         \includegraphics[trim={1cm 1cm 1cm 1cm},clip,width=1\textwidth]{17e916fc863540ee3def89b32cef8e45_gt_000000.png}
     \end{subfigure} 
     &
     \begin{subfigure}[t]{0.06\linewidth}
         \centering
         \includegraphics[trim={1cm 1cm 1cm 1cm},clip,width=1\textwidth]{19ce953da9aa8065d747a43c11e738e9_gt_000056.png}
     \end{subfigure}
    &
         \begin{subfigure}[t]{0.06\linewidth}
         \centering
         \includegraphics[trim={1cm 0.55cm 1cm 1cm},clip,width=0.9\textwidth]{144d0880f61d813ef7b860bd772a37_gt_000064.png}
     \end{subfigure} 
     &
      \begin{subfigure}[t]{0.06\linewidth}
         \centering
         \includegraphics[trim={1cm 0.55cm 1cm 1cm},clip,width=0.9\textwidth]{144d0880f61d813ef7b860bd772a37_gt_000104.png}
     \end{subfigure} 
     &
     \begin{subfigure}[t]{0.06\linewidth}
         \centering
         \includegraphics[trim={1cm 0.55cm 1cm 1cm},clip,width=0.9\textwidth]{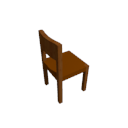}
     \end{subfigure} 
     &
     \begin{subfigure}[t]{0.06\linewidth}
         \centering
         \includegraphics[trim={1cm 0.55cm 1cm 1cm},clip,width=0.9\textwidth]{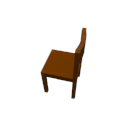}
     \end{subfigure} 
     &
      \begin{subfigure}[t]{0.06\linewidth}
         \centering
         \includegraphics[trim={1cm 0.55cm 1cm 1cm},clip,width=0.9\textwidth]{112cee32461c31d1d84b8ba651dfb8ac_gt_000100.png}
     \end{subfigure}
     &
     \begin{subfigure}[t]{0.06\linewidth}
         \centering
         \includegraphics[trim={1cm 0.55cm 1cm 1cm},clip,width=0.9\textwidth]{112cee32461c31d1d84b8ba651dfb8ac_gt_000129.png}
     \end{subfigure}
     &
     \begin{subfigure}[t]{0.06\linewidth}
         \centering
         \includegraphics[trim={1cm 0.55cm 1cm 1cm},clip,width=0.9\textwidth]{112cee32461c31d1d84b8ba651dfb8ac_gt_000100.png}
     \end{subfigure} 
     &
     \begin{subfigure}[t]{0.06\linewidth}
         \centering
         \includegraphics[trim={1cm 0.55cm 1cm 1cm},clip,width=0.9\textwidth]{112cee32461c31d1d84b8ba651dfb8ac_gt_000129.png}
     \end{subfigure} 
     
     \\
     \multicolumn{2}{c}{
     \begin{subfigure}[t]{0.06\linewidth}
         \centering
         \includegraphics[trim={0.55cm 0.75cm 0.75cm 0.75cm},clip,width=0.9\textwidth]{144d0880f61d813ef7b860bd772a37_gt_000064.png}
     \end{subfigure} 
     }
     &
     \begin{subfigure}[t]{0.06\linewidth}
         \centering
         \includegraphics[trim={0.55cm 0.75cm 0.75cm 0.75cm},clip,width=0.9\textwidth]{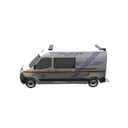}
     \end{subfigure} 
     &
     \begin{subfigure}[t]{0.06\linewidth}
         \centering
         \includegraphics[trim={0.55cm 0.75cm 0.5cm 0.75cm},clip,width=0.9\textwidth]{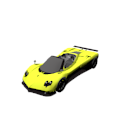}
     \end{subfigure} 
     &
      \begin{subfigure}[t]{0.06\linewidth}
         \centering
         \includegraphics[trim={0.55cm 0.75cm 0.75cm 0.75cm},clip,width=0.9\textwidth]{125a4780c2d5095d19454008aa267bf_gt_000054.png}
     \end{subfigure}
     &
     \begin{subfigure}[t]{0.06\linewidth}
         \centering
         \includegraphics[trim={0.55cm 0.75cm 0.5cm 0.75cm},clip,width=0.9\textwidth]{144d0880f61d813ef7b860bd772a37_gt_000065.png}
     \end{subfigure}
     &
     \begin{subfigure}[t]{0.06\linewidth}
         \centering
         \includegraphics[trim={0.55cm 0.75cm 0.75cm 0.75cm},clip,width=0.9\textwidth]{125a4780c2d5095d19454008aa267bf_gt_000054.png}
     \end{subfigure} 
     &
     \begin{subfigure}[t]{0.06\linewidth}
         \centering
         \includegraphics[trim={0.55cm 0.75cm 0.5cm 0.75cm},clip,width=0.9\textwidth]{144d0880f61d813ef7b860bd772a37_gt_000065.png}
     \end{subfigure}
     &

          \begin{subfigure}[t]{0.06\linewidth}
         \centering
         \includegraphics[trim={0.55cm 0.75cm 0.5cm 0.75cm},clip,width=0.9\textwidth]{144d0880f61d813ef7b860bd772a37_gt_000064.png}
     \end{subfigure} 
     &
     \begin{subfigure}[t]{0.06\linewidth}
         \centering
         \includegraphics[trim={0.55cm 0.75cm 0.5cm 0.75cm},clip,width=0.9\textwidth]{144d0880f61d813ef7b860bd772a37_gt_000104.png}
     \end{subfigure} 
     &
     \begin{subfigure}[t]{0.06\linewidth}
         \centering
         \includegraphics[trim={0.55cm 0.75cm 0.5cm 0.75cm},clip,width=0.9\textwidth]{144d0880f61d813ef7b860bd772a37_gt_000065.png}
     \end{subfigure} 
     &
     \begin{subfigure}[t]{0.06\linewidth}
         \centering
         \includegraphics[trim={0.3cm 0.3cm 0.3cm 0.3cm},clip,width=0.9\textwidth]{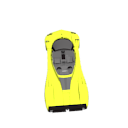}
     \end{subfigure} 
     &
      \begin{subfigure}[t]{0.06\linewidth}
         \centering
         \includegraphics[trim={0.55cm 0.75cm 0.5cm 0.75cm},clip,width=0.9\textwidth]{144d0880f61d813ef7b860bd772a37_gt_000065.png}
     \end{subfigure}
     &
     \begin{subfigure}[t]{0.06\linewidth}
         \centering
         \includegraphics[trim={0.3cm 0.3cm 0.3cm 0.3cm},clip,width=0.9\textwidth]{144d0880f61d813ef7b860bd772a37_gt_000192.png}
     \end{subfigure}
     &
     \begin{subfigure}[t]{0.06\linewidth}
         \centering
         \includegraphics[trim={0.55cm 0.75cm 0.5cm 0.75cm},clip,width=0.9\textwidth]{144d0880f61d813ef7b860bd772a37_gt_000065.png}
     \end{subfigure} 
     &
     \begin{subfigure}[t]{0.06\linewidth}
         \centering
         \includegraphics[trim={0.3cm 0.3cm 0.3cm 0.3cm},clip,width=0.9\textwidth]{144d0880f61d813ef7b860bd772a37_gt_000192.png}
     \end{subfigure} 

     \\
     \multicolumn{2}{c}{
         \begin{subfigure}[t]{0.06\linewidth}
             \centering
             \includegraphics[trim={0.55cm 0.75cm 0.75cm 1cm},clip,width=1\textwidth]{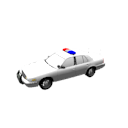}
         \end{subfigure} 
     }
     &
     \begin{subfigure}[t]{0.06\linewidth}
         \centering
         \includegraphics[trim={0.55cm 0.75cm 1cm 1cm},clip,width=1\textwidth]{11a96098620b2ebac2f9fb5458a091d1_gt_000060.png}
     \end{subfigure} 
     &
     \begin{subfigure}[t]{0.06\linewidth}
         \centering
         \includegraphics[trim={0.55cm 1cm 0.75cm 1cm},clip,width=1\textwidth]{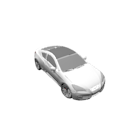}
     \end{subfigure} 
     &
      \begin{subfigure}[t]{0.06\linewidth}
         \centering
         \includegraphics[trim={0.55cm 1cm 0.75cm 1cm},clip,width=1\textwidth]{11a96098620b2ebac2f9fb5458a091d1_gt_000060.png}
     \end{subfigure}
     &
     \begin{subfigure}[t]{0.06\linewidth}
         \centering
         \includegraphics[trim={0.55cm 1cm 0.75cm 1cm},clip,width=1\textwidth]{1079efee042629d4ce28f0f1b509eda_gt_000081.png}
     \end{subfigure}
     &
     \begin{subfigure}[t]{0.06\linewidth}
         \centering
         \includegraphics[trim={0.55cm 1cm 0.75cm 1cm},clip,width=1\textwidth]{11a96098620b2ebac2f9fb5458a091d1_gt_000060.png}
     \end{subfigure} 
     &
     \begin{subfigure}[t]{0.06\linewidth}
         \centering
         \includegraphics[trim={0.55cm 1cm 0.75cm 1cm},clip,width=1\textwidth]{1079efee042629d4ce28f0f1b509eda_gt_000081.png}
     \end{subfigure} 
      
    &


     \begin{subfigure}[t]{0.06\linewidth}
         \centering
         \includegraphics[trim={0.55cm 0.75cm 0.5cm 0.75cm},clip,width=1\textwidth]{144d0880f61d813ef7b860bd772a37_gt_000064.png}
     \end{subfigure} 
     &
     \begin{subfigure}[t]{0.06\linewidth}
         \centering
         \includegraphics[trim={0.55cm 0.75cm 0.5cm 0.75cm},clip,width=1\textwidth]{144d0880f61d813ef7b860bd772a37_gt_000104.png}
     \end{subfigure} 
     &
     \begin{subfigure}[t]{0.06\linewidth}
         \centering
         \includegraphics[trim={0.55cm 0.75cm 0.5cm 0.75cm},clip,width=1\textwidth]{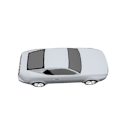}
     \end{subfigure} 
     &
     \begin{subfigure}[t]{0.06\linewidth}
         \centering
         \includegraphics[trim={0.3cm 0.3cm 0.3cm 0.3cm},clip,width=1\textwidth]{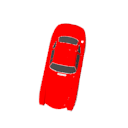}
     \end{subfigure} 
     &
      \begin{subfigure}[t]{0.06\linewidth}
         \centering
         \includegraphics[trim={0.55cm 0.75cm 0.5cm 0.75cm},clip,width=1\textwidth]{158a95b4da25aa1fa37f3fc191551700_gt_000172.png}
     \end{subfigure}
     &
     \begin{subfigure}[t]{0.06\linewidth}
         \centering
         \includegraphics[trim={0.3cm 0.3cm 0.3cm 0.3cm},clip,width=1\textwidth]{1523402e11400b75becf71e2e014ff6f_gt_000228.png}
     \end{subfigure}
     &
     \begin{subfigure}[t]{0.06\linewidth}
         \centering
         \includegraphics[trim={0.55cm 0.75cm 0.5cm 0.75cm},clip,width=1\textwidth]{144d0880f61d813ef7b860bd772a37_gt_000104.png}
     \end{subfigure} 
     &
     \begin{subfigure}[t]{0.06\linewidth}
         \centering
         \includegraphics[trim={0.55cm 0.75cm 0.5cm 0.75cm},clip,width=1\textwidth]{158a95b4da25aa1fa37f3fc191551700_gt_000172.png}
     \end{subfigure}
     \\
\botrule
\end{tabular}
        \caption{Category-specific one- and two-shot neural rendering benchmark.
        }
        \label{fig:2v_cars_chairs_results_main}
\end{figure}

Following the validation scheme of PixelNeRF, we extensively evaluated our model in two experimental categories:
\begin{enumerate*}
\item the real scenes of the DTU MVS dataset, and
\item the ShapeNet benchmark for category-specific view synthesis.
\end{enumerate*}
  For the experiments on the DTU dataset, we compared to PixelNeRF, training our model on sparse views, similar to it. We reproduced the results of PixelNeRF using the same quality metrics. Table~\ref{table:avg_comparasion} presents PSNR, SSIM, LPIP, DISTS, and the number of views values for the considered NeRF-like methods for the 15 scenes\footnoteref{sup_desc}.
  Briefly, we calculated quality metrics for each rendered image in each considered scene to ensure consistency with previous works. We then determine the quality metrics at scene level, by averaging the quality of all rendered images of the same scene. Next, we average the quality metrics across all scenes to compute the final quality values per method. This process is repeated for 1, 3, 6, and 9 views settings. Similarly, for the ShapeNet benchmark presented in Table. \ref{table:cars_chairs_comparsions}, we compared it to PixelNeRF using one and two view settings. 
Notably, our model requires only one day to complete training, which is 4x faster than the baseline. The qualitative results on the DTU dataset are shown in Fig.~\ref{fig:1_9v_settings_dtu_results}. Additionally,  Fig.~\ref{fig:2v_cars_chairs_results_main} shows the qualitative results on the two category-specific from the ShapeNet dataset. The figures show that our model excels in texture details, artifact correction, missing data handling, and color adjustment across different scene parts, surpassing other models.  Table~\ref{table:cars_chairs_comparsions} shows the quantitative results of our comparison. As it can be seen in the table, our model shows better performance in novel rendering in different views settings. 

\begin{table}[t]
\setlength{\tabcolsep}{1pt}
\scriptsize
\centering
\caption{Quantitative comparison of our model with PixelNeRF on the two category-specific datasets called chairs and cars.}
\begin{tabular}{@{\extracolsep\fill}lllcccc}
\toprule
                & &  & \multicolumn{2}{@{}c@{}}{1 View} & \multicolumn{2}{@{}c@{}}{2 Views}  \\\cmidrule{4-4}\cmidrule{5-7}
                 & & & PSNR↑   & SSIM↑  & PSNR↑ & SSIM↑ \\
\midrule
\multirow{2}{*}{Chairs}& \vline  & PixelNeRF & 23.72 & 0.91 & 26.20 & 0.94 \\
       & \vline & Ours & \textbf{27.79} & \textbf{0.93} & \textbf{27.42}	& \textbf{0.95} \\
\midrule
\multirow{2}{*}{Cars} & \vline  & PixelNeRF & 23.17 & 0.90 & 25.66 & 0.94 \\
   & \vline  & Ours & \textbf{28.09} & \textbf{0.93} & \textbf{27.24} & \textbf{0.95} \\
\botrule
\end{tabular}

\label{table:cars_chairs_comparsions}
\end{table}


\subsection{Ablation study}
In order to understand the factors contributing to the success of MomentsNeRF, we conducted a study on the impact of removing various components. The results of this analysis can be found in Table \ref{table:ablation_comparasion}. 
 The study indicates that Zernike polynomials mainly enhance texture details, particularly on LPIPS and DISTS metrics for all views.
In contrast, Gabor filters impact mainly on the image structure, with notable improvements on the PSNR and SSIM across all views. 
The study also shows the effect of the Positional Encoding and the Activation Function 
 settings to fit the baseline settings showing a slight degradation in the performance.
In Fig.~\ref{fig:ablation_study}, our study showcases the impact of view count on Moments Neural Rendering. The findings show that the texture is more prominently displayed in cases with fewer views, such as 1 or 3.  
Moreover, Table ~\ref{table:ablation_comparasion_randomly_views_settings} shows slight performance degradation with having bad exposures compared with the results in Table ~\ref{table:avg_comparasion}. On test time, we randomly changed the view settings for both ours and the baseline models; we used fixed settings for 1, 3, 6, and 9 views \footnoteref{illu}. Table \ref{table:ablation_comparasion_randomly_views_settings} shows a slight degradation in overall performance \footnoteref{illu}. The supplemental material includes additional quantitative results, qualitative comparisons with varying input views, and discussions.

\begin{table}[htb]
\setlength{\tabcolsep}{1pt}
\scriptsize
\centering
\caption{Ablation study of our model on DTU dataset by removing various components (best results in red, second-best in orange, and third-best in yellow).
}
\begin{tabular}{@{\extracolsep\fill}lllllllllllllllll}
\toprule
                 & \multicolumn{4}{@{}c@{}}{1 View}    & 
                 \multicolumn{4}{@{}c@{}}{3 Views}    & \multicolumn{4}{@{}c@{}}{6 Views}    & \multicolumn{4}{@{}c@{}}{9 Views} 
                 \\\cmidrule{2-5} \cmidrule{6-9} \cmidrule{10-13} \cmidrule{14-17}
                  & \tiny{PSNR↑}   & \tiny{SSIM↑}  & \tiny{LPIPS↓} & \tiny{DISTS↓}  & \tiny{PSNR↑}   & \tiny{SSIM↑}  & \tiny{LPIPS↓} & \tiny{DISTS↓} & \tiny{PSNR↑}   & \tiny{SSIM↑}  & \tiny{LPIPS↓} & \tiny{DISTS↓} & \tiny{PSNR↑}   & \tiny{SSIM↑}  & \tiny{LPIPS↓} & \tiny{DISTS↓} \\
\midrule
Ours   & \cellcolor{red!25}21.543 & \cellcolor{red!25}0.729 & \cellcolor{red!25}0.186 & \cellcolor{red!25}0.178 & \cellcolor{orange!25}23.810 & \cellcolor{red!25}0.828 & \cellcolor{red!25}0.138 & \cellcolor{red!25}0.169 & \cellcolor{yellow!25}24.443 & \cellcolor{red!25}0.847 & \cellcolor{red!25}0.131 & \cellcolor{red!25}0.173 & \cellcolor{red!25}24.655 & \cellcolor{red!25}0.855 & \cellcolor{red!25}0.127 & \cellcolor{red!25}0.174 \\
Ours\textsuperscript{1}  & \cellcolor{orange!25}17.337	& \cellcolor{orange!25}0.386 & 0.733 & 0.459 &  \cellcolor{yellow!25}23.794 &  \cellcolor{yellow!25}0.808 & 0.511 & 0.312 & \cellcolor{orange!25}24.445 &  \cellcolor{yellow!25}0.836 & 0.435 & 0.280 & 24.565 & 0.842 & 0.385 & 0.252 \\
Ours\textsuperscript{2} & 17.107 & 0.370 & \cellcolor{yellow!25} 0.366 & \cellcolor{yellow!25} 0.321 &  23.419 & 0.795 & \cellcolor{yellow!25} 0.156 & \cellcolor{yellow!25} 0.181 & 24.214 & 0.834 & \cellcolor{yellow!25}0.137 & \cellcolor{yellow!25}0.182 & \cellcolor{yellow!25}24.597 &  \cellcolor{yellow!25}0.846 & \cellcolor{yellow!25}0.132 &  \cellcolor{yellow!25}0.183 \\
Ours\textsuperscript{3} & \cellcolor{yellow!25}17.263 & \cellcolor{yellow!25}0.376 & \cellcolor{orange!25}0.356 & \cellcolor{orange!25}0.317 &  \cellcolor{red!25}23.992 &  \cellcolor{orange!25}0.818 &  \cellcolor{orange!25}0.145 &  \cellcolor{orange!25}0.174 &  \cellcolor{red!25}24.540 &  \cellcolor{orange!25}0.845 &  \cellcolor{orange!25}0.134 & \cellcolor{orange!25} 0.181 &  \cellcolor{orange!25}24.641 & \cellcolor{orange!25} 0.850 &  \cellcolor{orange!25}0.131 &  \cellcolor{orange!25}0.179 \\
\botrule
\end{tabular}
\footnotetext{\textsuperscript{(1)}w/o Zernike; \textsuperscript{(2)}w/o Gabor; \textsuperscript{(3)} w/o PE \& AF}

\label{table:ablation_comparasion}
\vspace{-1in}
\end{table}






\begin{table}[htb]
\setlength{\tabcolsep}{1pt}
\scriptsize
\centering
\caption{On the top is an ablation study after randomly changing the view settings. On the bottom is an ablation study by keeping the bad exposure. Both ablations are done on the DTU dataset on test time (the best results are in red, and the second-best are in orange).
}
\begin{tabular}{@{\extracolsep\fill}ccccccccccccccccc}
\toprule
                 & \multicolumn{4}{@{}c@{}}{1 View}    & \multicolumn{4}{@{}c@{}}{3 Views}    & \multicolumn{4}{@{}c@{}}{6 Views}    & \multicolumn{4}{@{}c@{}}{9 Views}                     \\\cmidrule{2-5} \cmidrule{6-9} \cmidrule{10-13} \cmidrule{14-17}    
                  & \tiny{PSNR↑}   & \tiny{SSIM↑}  & \tiny{LPIPS↓} & \tiny{DISTS↓}  & \tiny{PSNR↑}   & \tiny{SSIM↑}  & \tiny{LPIPS↓} & \tiny{DISTS↓} & \tiny{PSNR↑}   & \tiny{SSIM↑}  & \tiny{LPIPS↓} & \tiny{DISTS↓} & \tiny{PSNR↑}   & \tiny{SSIM↑}  & \tiny{LPIPS↓} & \tiny{DISTS↓} \\
\midrule
\footnotesize{PixelNeRF}  & \cellcolor{orange!25}15.142 & \cellcolor{orange!25}0.497 & \cellcolor{orange!25}0.200 & \cellcolor{orange!25}0.191 & \cellcolor{orange!25}18.729 & \cellcolor{orange!25}0.677 & \cellcolor{red!25}0.127 & \cellcolor{red!25}0.169 & \cellcolor{orange!25}19.852 & \cellcolor{orange!25}0.716 & \cellcolor{red!25}0.113 & \cellcolor{red!25}0.167 & \cellcolor{orange!25}20.455 & \cellcolor{orange!25}0.737 & \cellcolor{red!25}0.105 & \cellcolor{red!25}0.163 \\
\footnotesize{Ours}  & \cellcolor{red!25}21.543 & \cellcolor{red!25}0.729 & \cellcolor{red!25}0.186 & \cellcolor{red!25}0.178 & \cellcolor{red!25}23.810 & \cellcolor{red!25}0.828 & \cellcolor{orange!25}0.138 & \cellcolor{red!25}0.169 & \cellcolor{red!25}24.443 & \cellcolor{red!25}0.847 & \cellcolor{orange!25}0.131 & \cellcolor{orange!25}0.173 & \cellcolor{red!25}24.655 & \cellcolor{red!25}0.855 & \cellcolor{orange!25}0.127 & \cellcolor{orange!25}0.174 \\
\midrule

\footnotesize{PixelNeRF*} & \cellcolor{orange!25}15.226 & \cellcolor{orange!25}0.497 & - & - & \cellcolor{orange!25}18.729 &  \cellcolor{orange!25}0.677 & - & - & \cellcolor{orange!25}18.949 &  \cellcolor{orange!25}0.691 & - & - & \cellcolor{orange!25}19.430 & \cellcolor{orange!25}0.708 & - & - \\
\footnotesize{Ours*}  & \cellcolor{red!25}21.291 & \cellcolor{red!25}0.717 &- & - & \cellcolor{red!25}23.210 & \cellcolor{red!25}0.810  & - & - & \cellcolor{red!25}24.143 & \cellcolor{red!25}0.807  &  - & - & \cellcolor{red!25}24.155 &  \cellcolor{red!25}0.805 &-&- \\
\botrule
\end{tabular}

\label{table:ablation_comparasion_randomly_views_settings}
\end{table}

 \begin{figure}[htb]
    \centering\setlength{\tabcolsep}{1pt}
    \begin{tabular}{ccccc}
     1 view & 3 views & 6 views & 9 views & Reference \\ 
     \begin{subfigure}[t]{.19\linewidth}
         \includegraphics[width=\textwidth]{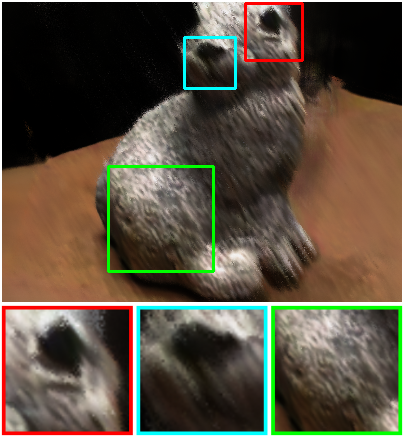} 
     \end{subfigure} 
     &
     \begin{subfigure}[t]{.19\linewidth}
         \includegraphics[width=\textwidth]{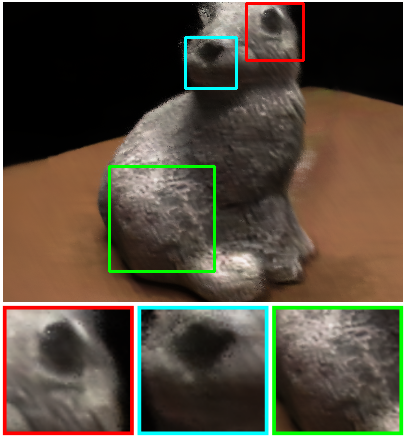} 
     \end{subfigure}
     &
     \begin{subfigure}[t]{.19\linewidth}
         \includegraphics[width=\textwidth]{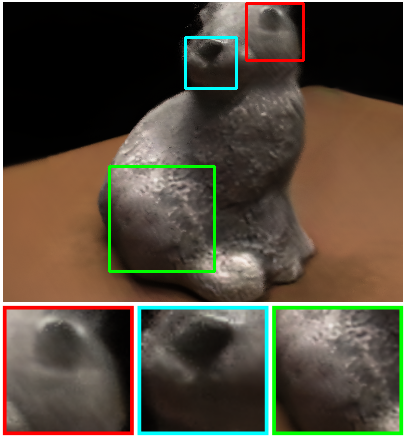} 
     \end{subfigure}
     &
     \begin{subfigure}[t]{.19\linewidth}
         \includegraphics[width=\textwidth]{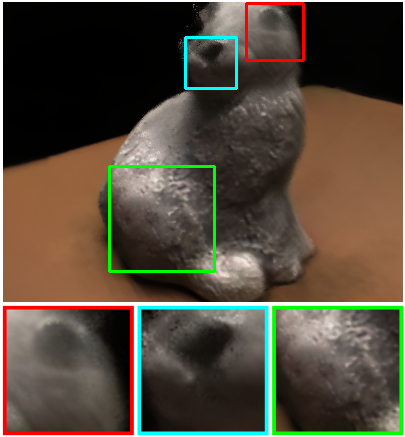} 
     \end{subfigure}
     &
     \begin{subfigure}[t]{.19\linewidth}
         \includegraphics[width=\textwidth]{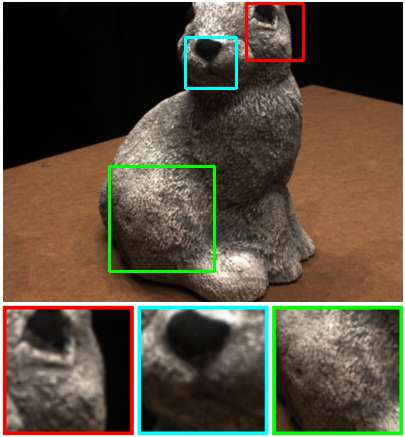} 
     \end{subfigure}
    \\
    \begin{subfigure}[t]{.19\linewidth}
         \includegraphics[width=\textwidth]{ablation_scan55_fig_sample_1v.png} 
     \end{subfigure} 
     &
     \begin{subfigure}[t]{.19\linewidth}
         \includegraphics[width=\textwidth]{ablation_scan55_fig_sample_3v.png} 
     \end{subfigure}
     &
     \begin{subfigure}[t]{.19\linewidth}
         \includegraphics[width=\textwidth]{ablation_scan55_fig_sample_6v.png} 
     \end{subfigure}
     &
     \begin{subfigure}[t]{.19\linewidth}
         \includegraphics[width=\textwidth]{ablation_scan55_fig_sample_9v.png} 
     \end{subfigure}
     &
     \begin{subfigure}[t]{.19\linewidth}
         \includegraphics[width=\textwidth]{ablation_scan55_fig_sample_gt.png} 
     \end{subfigure}
\end{tabular}
        \caption{Ablation study for the same view from the same scene, but in different view settings. On the top, scan55 rendering on different view settings is shown, while on the bottom, scan114 is illustrated in different view settings.}
        \label{fig:ablation_study}
\end{figure}

\paragraph{Limitations.}
Similar to the baseline, the limitations of our method are presented, respectively:
\begin{enumerate*}
    \item Our method may not photo-realistically reconstruct non-rigidly deforming scenes captured casually from mobile phones; 
    \item Using a different number of views in testing than in training can impair model performance\footnoteref{sup_desc};
    \item During testing, altering input view settings leads to varying outcomes;
    \item Training on uncorrelated scenes data results in a decline in performance.
\end{enumerate*}

\section{Conclusions and Future Work}
\label{sec:conclusion}
We have presented MomentsNeRF, a novel framework incorporating orthogonal moments in NeRF, enhancing the one- and few-shot neural rendering.  Our model outperforms the performance of the majority of NeRF-like frameworks (Table. \ref{table:avg_comparasion}). Our findings shed important light on the stability of NeRF-based models. 
Still, the very essential and interesting question of—how the MomentsNeRF for a 360º real scene impacts NeRF's robustness—is still open and is part of our future work. Moreover, we are exploring other Moments categories  (see the supplementary material) to extend our framework and achieve interesting properties for the MomentsNeRF.
\bmhead{Supplementary information}
For a detailed description, kindly refer to the supplementary material.
\bmhead{Acknowledgement}
This work was partially funded by the EU project MUSAE (No. 01070421), 2021-SGR-01094 (AGAUR), Icrea Academia’2022 (Generalitat de Catalunya), Robo STEAM (2022-1-BG01-KA220-VET000089434, Erasmus+ EU), DeepSense (ACE053/22/000029, ACCIÓ), CERCA Programme/Generalitat de Catalunya, and Grants PID2022141566NB-I00 (IDEATE), PDC2022-133642-I00 (DeepFoodVol), and CNS2022-135480 (A-BMC) funded by MICIU/AEI/10.13039/501100 011033, by FEDER (UE), and by European Union NextGenerationEU/ PRTR. R. Marques acknowledges the support of the Serra Húnter Programme. A. AlMughrabi acknowledges the support of FPI Becas, MICINN, Spain.

\bibliography{sn-article}


\begin{thebibliography}{71}
\ifx \bisbn   \undefined \def \bisbn  #1{ISBN #1}\fi
\ifx \binits  \undefined \def \binits#1{#1}\fi
\ifx \bauthor  \undefined \def \bauthor#1{#1}\fi
\ifx \batitle  \undefined \def \batitle#1{#1}\fi
\ifx \bjtitle  \undefined \def \bjtitle#1{#1}\fi
\ifx \bvolume  \undefined \def \bvolume#1{\textbf{#1}}\fi
\ifx \byear  \undefined \def \byear#1{#1}\fi
\ifx \bissue  \undefined \def \bissue#1{#1}\fi
\ifx \bfpage  \undefined \def \bfpage#1{#1}\fi
\ifx \blpage  \undefined \def \blpage #1{#1}\fi
\ifx \burl  \undefined \def \burl#1{\textsf{#1}}\fi
\ifx \doiurl  \undefined \def \doiurl#1{\url{https://doi.org/#1}}\fi
\ifx \betal  \undefined \def \betal{\textit{et al.}}\fi
\ifx \binstitute  \undefined \def \binstitute#1{#1}\fi
\ifx \binstitutionaled  \undefined \def \binstitutionaled#1{#1}\fi
\ifx \bctitle  \undefined \def \bctitle#1{#1}\fi
\ifx \beditor  \undefined \def \beditor#1{#1}\fi
\ifx \bpublisher  \undefined \def \bpublisher#1{#1}\fi
\ifx \bbtitle  \undefined \def \bbtitle#1{#1}\fi
\ifx \bedition  \undefined \def \bedition#1{#1}\fi
\ifx \bseriesno  \undefined \def \bseriesno#1{#1}\fi
\ifx \blocation  \undefined \def \blocation#1{#1}\fi
\ifx \bsertitle  \undefined \def \bsertitle#1{#1}\fi
\ifx \bsnm \undefined \def \bsnm#1{#1}\fi
\ifx \bsuffix \undefined \def \bsuffix#1{#1}\fi
\ifx \bparticle \undefined \def \bparticle#1{#1}\fi
\ifx \barticle \undefined \def \barticle#1{#1}\fi
\bibcommenthead
\ifx \bconfdate \undefined \def \bconfdate #1{#1}\fi
\ifx \botherref \undefined \def \botherref #1{#1}\fi
\ifx \url \undefined \def \url#1{\textsf{#1}}\fi
\ifx \bchapter \undefined \def \bchapter#1{#1}\fi
\ifx \bbook \undefined \def \bbook#1{#1}\fi
\ifx \bcomment \undefined \def \bcomment#1{#1}\fi
\ifx \oauthor \undefined \def \oauthor#1{#1}\fi
\ifx \citeauthoryear \undefined \def \citeauthoryear#1{#1}\fi
\ifx \endbibitem  \undefined \def \endbibitem {}\fi
\ifx \bconflocation  \undefined \def \bconflocation#1{#1}\fi
\ifx \arxivurl  \undefined \def \arxivurl#1{\textsf{#1}}\fi
\csname PreBibitemsHook\endcsname

\bibitem[\protect\citeauthoryear{Xie et~al.}{2022}]{xie2022neural}
\begin{bchapter}
\bauthor{\bsnm{Xie}, \binits{Y.}},
\bauthor{\bsnm{Takikawa}, \binits{T.}},
\bauthor{\bsnm{Saito}, \binits{S.}},
\bauthor{\bsnm{Litany}, \binits{O.}},
\bauthor{\bsnm{Yan}, \binits{S.}},
\bauthor{\bsnm{Khan}, \binits{N.}},
\bauthor{\bsnm{Tombari}, \binits{F.}},
\bauthor{\bsnm{Tompkin}, \binits{J.}},
\bauthor{\bsnm{Sitzmann}, \binits{V.}},
\bauthor{\bsnm{Sridhar}, \binits{S.}}:
\bctitle{Neural fields in visual computing and beyond}.
In: \bbtitle{Computer Graphics Forum},
vol. \bseriesno{41},
pp. \bfpage{641}--\blpage{676}
(\byear{2022}).
\bcomment{Wiley Online Library}
\end{bchapter}
\endbibitem

\bibitem[\protect\citeauthoryear{Mildenhall et~al.}{2021}]{mildenhall2021nerf}
\begin{barticle}
\bauthor{\bsnm{Mildenhall}, \binits{B.}},
\bauthor{\bsnm{Srinivasan}, \binits{P.P.}},
\bauthor{\bsnm{Tancik}, \binits{M.}},
\bauthor{\bsnm{Barron}, \binits{J.T.}},
\bauthor{\bsnm{Ramamoorthi}, \binits{R.}},
\bauthor{\bsnm{Ng}, \binits{R.}}:
\batitle{Nerf: Representing scenes as neural radiance fields for view synthesis}.
\bjtitle{Communications of the ACM}
\bvolume{65}(\bissue{1}),
\bfpage{99}--\blpage{106}
(\byear{2021})
\end{barticle}
\endbibitem

\bibitem[\protect\citeauthoryear{Isaac-Medina et~al.}{2023}]{isaac2023exact}
\begin{bchapter}
\bauthor{\bsnm{Isaac-Medina}, \binits{B.K.}},
\bauthor{\bsnm{Willcocks}, \binits{C.G.}},
\bauthor{\bsnm{Breckon}, \binits{T.P.}}:
\bctitle{Exact-nerf: An exploration of a precise volumetric parameterization for neural radiance fields}.
In: \bbtitle{Proceedings of the IEEE/CVF Conference on Computer Vision and Pattern Recognition},
pp. \bfpage{66}--\blpage{75}
(\byear{2023})
\end{bchapter}
\endbibitem

\bibitem[\protect\citeauthoryear{Jain et~al.}{2022}]{jain2022zero}
\begin{bchapter}
\bauthor{\bsnm{Jain}, \binits{A.}},
\bauthor{\bsnm{Mildenhall}, \binits{B.}},
\bauthor{\bsnm{Barron}, \binits{J.T.}},
\bauthor{\bsnm{Abbeel}, \binits{P.}},
\bauthor{\bsnm{Poole}, \binits{B.}}:
\bctitle{Zero-shot text-guided object generation with dream fields}.
In: \bbtitle{Proceedings of the IEEE/CVF Conference on Computer Vision and Pattern Recognition},
pp. \bfpage{867}--\blpage{876}
(\byear{2022})
\end{bchapter}
\endbibitem

\bibitem[\protect\citeauthoryear{Martin-Brualla et~al.}{2021}]{martin2021nerf}
\begin{bchapter}
\bauthor{\bsnm{Martin-Brualla}, \binits{R.}},
\bauthor{\bsnm{Radwan}, \binits{N.}},
\bauthor{\bsnm{Sajjadi}, \binits{M.S.}},
\bauthor{\bsnm{Barron}, \binits{J.T.}},
\bauthor{\bsnm{Dosovitskiy}, \binits{A.}},
\bauthor{\bsnm{Duckworth}, \binits{D.}}:
\bctitle{Nerf in the wild: Neural radiance fields for unconstrained photo collections}.
In: \bbtitle{Proceedings of the IEEE/CVF Conference on Computer Vision and Pattern Recognition},
pp. \bfpage{7210}--\blpage{7219}
(\byear{2021})
\end{bchapter}
\endbibitem

\bibitem[\protect\citeauthoryear{Cai et~al.}{2022}]{cai2022pix2nerf}
\begin{bchapter}
\bauthor{\bsnm{Cai}, \binits{S.}},
\bauthor{\bsnm{Obukhov}, \binits{A.}},
\bauthor{\bsnm{Dai}, \binits{D.}},
\bauthor{\bsnm{Van~Gool}, \binits{L.}}:
\bctitle{Pix2nerf: Unsupervised conditional p-gan for single image to neural radiance fields translation}.
In: \bbtitle{Proceedings of the IEEE/CVF Conference on Computer Vision and Pattern Recognition},
pp. \bfpage{3981}--\blpage{3990}
(\byear{2022})
\end{bchapter}
\endbibitem

\bibitem[\protect\citeauthoryear{Dogaru et~al.}{2023}]{dogaru2023sphere}
\begin{bchapter}
\bauthor{\bsnm{Dogaru}, \binits{A.}},
\bauthor{\bsnm{Ardelean}, \binits{A.-T.}},
\bauthor{\bsnm{Ignatyev}, \binits{S.}},
\bauthor{\bsnm{Zakharov}, \binits{E.}},
\bauthor{\bsnm{Burnaev}, \binits{E.}}:
\bctitle{Sphere-guided training of neural implicit surfaces}.
In: \bbtitle{Proceedings of the IEEE/CVF Conference on Computer Vision and Pattern Recognition},
pp. \bfpage{20844}--\blpage{20853}
(\byear{2023})
\end{bchapter}
\endbibitem

\bibitem[\protect\citeauthoryear{Li et~al.}{2023}]{li2023neuralangelo}
\begin{bchapter}
\bauthor{\bsnm{Li}, \binits{Z.}},
\bauthor{\bsnm{M{\"u}ller}, \binits{T.}},
\bauthor{\bsnm{Evans}, \binits{A.}},
\bauthor{\bsnm{Taylor}, \binits{R.H.}},
\bauthor{\bsnm{Unberath}, \binits{M.}},
\bauthor{\bsnm{Liu}, \binits{M.-Y.}},
\bauthor{\bsnm{Lin}, \binits{C.-H.}}:
\bctitle{Neuralangelo: High-fidelity neural surface reconstruction}.
In: \bbtitle{Proceedings of the IEEE/CVF Conference on Computer Vision and Pattern Recognition},
pp. \bfpage{8456}--\blpage{8465}
(\byear{2023})
\end{bchapter}
\endbibitem

\bibitem[\protect\citeauthoryear{Lin et~al.}{2023}]{lin2023vision}
\begin{bchapter}
\bauthor{\bsnm{Lin}, \binits{K.-E.}},
\bauthor{\bsnm{Lin}, \binits{Y.-C.}},
\bauthor{\bsnm{Lai}, \binits{W.-S.}},
\bauthor{\bsnm{Lin}, \binits{T.-Y.}},
\bauthor{\bsnm{Shih}, \binits{Y.-C.}},
\bauthor{\bsnm{Ramamoorthi}, \binits{R.}}:
\bctitle{Vision transformer for nerf-based view synthesis from a single input image}.
In: \bbtitle{Proceedings of the IEEE/CVF Winter Conference on Applications of Computer Vision},
pp. \bfpage{806}--\blpage{815}
(\byear{2023})
\end{bchapter}
\endbibitem

\bibitem[\protect\citeauthoryear{Poole et~al.}{2022}]{poole2022dreamfusion}
\begin{botherref}
\oauthor{\bsnm{Poole}, \binits{B.}},
\oauthor{\bsnm{Jain}, \binits{A.}},
\oauthor{\bsnm{Barron}, \binits{J.T.}},
\oauthor{\bsnm{Mildenhall}, \binits{B.}}:
Dreamfusion: Text-to-3d using 2d diffusion.
arXiv preprint arXiv:2209.14988
(2022)
\end{botherref}
\endbibitem

\bibitem[\protect\citeauthoryear{Park et~al.}{2021}]{park2021nerfies}
\begin{bchapter}
\bauthor{\bsnm{Park}, \binits{K.}},
\bauthor{\bsnm{Sinha}, \binits{U.}},
\bauthor{\bsnm{Barron}, \binits{J.T.}},
\bauthor{\bsnm{Bouaziz}, \binits{S.}},
\bauthor{\bsnm{Goldman}, \binits{D.B.}},
\bauthor{\bsnm{Seitz}, \binits{S.M.}},
\bauthor{\bsnm{Martin-Brualla}, \binits{R.}}:
\bctitle{Nerfies: Deformable neural radiance fields}.
In: \bbtitle{Proceedings of the IEEE/CVF International Conference on Computer Vision},
pp. \bfpage{5865}--\blpage{5874}
(\byear{2021})
\end{bchapter}
\endbibitem

\bibitem[\protect\citeauthoryear{Pumarola et~al.}{2021}]{pumarola2021d}
\begin{bchapter}
\bauthor{\bsnm{Pumarola}, \binits{A.}},
\bauthor{\bsnm{Corona}, \binits{E.}},
\bauthor{\bsnm{Pons-Moll}, \binits{G.}},
\bauthor{\bsnm{Moreno-Noguer}, \binits{F.}}:
\bctitle{D-nerf: Neural radiance fields for dynamic scenes}.
In: \bbtitle{Proceedings of the IEEE/CVF Conference on Computer Vision and Pattern Recognition},
pp. \bfpage{10318}--\blpage{10327}
(\byear{2021})
\end{bchapter}
\endbibitem

\bibitem[\protect\citeauthoryear{Wang et~al.}{2023}]{wang2023flow}
\begin{bchapter}
\bauthor{\bsnm{Wang}, \binits{C.}},
\bauthor{\bsnm{MacDonald}, \binits{L.E.}},
\bauthor{\bsnm{Jeni}, \binits{L.A.}},
\bauthor{\bsnm{Lucey}, \binits{S.}}:
\bctitle{Flow supervision for deformable nerf}.
In: \bbtitle{Proceedings of the IEEE/CVF Conference on Computer Vision and Pattern Recognition},
pp. \bfpage{21128}--\blpage{21137}
(\byear{2023})
\end{bchapter}
\endbibitem

\bibitem[\protect\citeauthoryear{Geng et~al.}{2023}]{geng2023learning}
\begin{bchapter}
\bauthor{\bsnm{Geng}, \binits{C.}},
\bauthor{\bsnm{Peng}, \binits{S.}},
\bauthor{\bsnm{Xu}, \binits{Z.}},
\bauthor{\bsnm{Bao}, \binits{H.}},
\bauthor{\bsnm{Zhou}, \binits{X.}}:
\bctitle{Learning neural volumetric representations of dynamic humans in minutes}.
In: \bbtitle{Proceedings of the IEEE/CVF Conference on Computer Vision and Pattern Recognition},
pp. \bfpage{8759}--\blpage{8770}
(\byear{2023})
\end{bchapter}
\endbibitem

\bibitem[\protect\citeauthoryear{Jayasundara et~al.}{2023}]{jayasundara2023flexnerf}
\begin{bchapter}
\bauthor{\bsnm{Jayasundara}, \binits{V.}},
\bauthor{\bsnm{Agrawal}, \binits{A.}},
\bauthor{\bsnm{Heron}, \binits{N.}},
\bauthor{\bsnm{Shrivastava}, \binits{A.}},
\bauthor{\bsnm{Davis}, \binits{L.S.}}:
\bctitle{Flexnerf: Photorealistic free-viewpoint rendering of moving humans from sparse views}.
In: \bbtitle{Proceedings of the IEEE/CVF Conference on Computer Vision and Pattern Recognition},
pp. \bfpage{21118}--\blpage{21127}
(\byear{2023})
\end{bchapter}
\endbibitem

\bibitem[\protect\citeauthoryear{Li et~al.}{2023}]{li2023dynibar}
\begin{bchapter}
\bauthor{\bsnm{Li}, \binits{Z.}},
\bauthor{\bsnm{Wang}, \binits{Q.}},
\bauthor{\bsnm{Cole}, \binits{F.}},
\bauthor{\bsnm{Tucker}, \binits{R.}},
\bauthor{\bsnm{Snavely}, \binits{N.}}:
\bctitle{Dynibar: Neural dynamic image-based rendering}.
In: \bbtitle{Proceedings of the IEEE/CVF Conference on Computer Vision and Pattern Recognition},
pp. \bfpage{4273}--\blpage{4284}
(\byear{2023})
\end{bchapter}
\endbibitem

\bibitem[\protect\citeauthoryear{Deng et~al.}{2022}]{deng2022depth}
\begin{bchapter}
\bauthor{\bsnm{Deng}, \binits{K.}},
\bauthor{\bsnm{Liu}, \binits{A.}},
\bauthor{\bsnm{Zhu}, \binits{J.-Y.}},
\bauthor{\bsnm{Ramanan}, \binits{D.}}:
\bctitle{Depth-supervised nerf: Fewer views and faster training for free}.
In: \bbtitle{Proceedings of the IEEE/CVF Conference on Computer Vision and Pattern Recognition},
pp. \bfpage{12882}--\blpage{12891}
(\byear{2022})
\end{bchapter}
\endbibitem

\bibitem[\protect\citeauthoryear{Roessle et~al.}{2022}]{roessle2022dense}
\begin{bchapter}
\bauthor{\bsnm{Roessle}, \binits{B.}},
\bauthor{\bsnm{Barron}, \binits{J.T.}},
\bauthor{\bsnm{Mildenhall}, \binits{B.}},
\bauthor{\bsnm{Srinivasan}, \binits{P.P.}},
\bauthor{\bsnm{Nie{\ss}ner}, \binits{M.}}:
\bctitle{Dense depth priors for neural radiance fields from sparse input views}.
In: \bbtitle{Proceedings of the IEEE/CVF Conference on Computer Vision and Pattern Recognition},
pp. \bfpage{12892}--\blpage{12901}
(\byear{2022})
\end{bchapter}
\endbibitem

\bibitem[\protect\citeauthoryear{Uy et~al.}{2023}]{uy2023scade}
\begin{bchapter}
\bauthor{\bsnm{Uy}, \binits{M.A.}},
\bauthor{\bsnm{Martin-Brualla}, \binits{R.}},
\bauthor{\bsnm{Guibas}, \binits{L.}},
\bauthor{\bsnm{Li}, \binits{K.}}:
\bctitle{Scade: Nerfs from space carving with ambiguity-aware depth estimates}.
In: \bbtitle{Proceedings of the IEEE/CVF Conference on Computer Vision and Pattern Recognition},
pp. \bfpage{16518}--\blpage{16527}
(\byear{2023})
\end{bchapter}
\endbibitem

\bibitem[\protect\citeauthoryear{Chen et~al.}{2023}]{chen2023mobilenerf}
\begin{bchapter}
\bauthor{\bsnm{Chen}, \binits{Z.}},
\bauthor{\bsnm{Funkhouser}, \binits{T.}},
\bauthor{\bsnm{Hedman}, \binits{P.}},
\bauthor{\bsnm{Tagliasacchi}, \binits{A.}}:
\bctitle{Mobilenerf: Exploiting the polygon rasterization pipeline for efficient neural field rendering on mobile architectures}.
In: \bbtitle{Proceedings of the IEEE/CVF Conference on Computer Vision and Pattern Recognition},
pp. \bfpage{16569}--\blpage{16578}
(\byear{2023})
\end{bchapter}
\endbibitem

\bibitem[\protect\citeauthoryear{Wang et~al.}{2023}]{wang2023f}
\begin{botherref}
\oauthor{\bsnm{Wang}, \binits{P.}},
\oauthor{\bsnm{Liu}, \binits{Y.}},
\oauthor{\bsnm{Chen}, \binits{Z.}},
\oauthor{\bsnm{Liu}, \binits{L.}},
\oauthor{\bsnm{Liu}, \binits{Z.}},
\oauthor{\bsnm{Komura}, \binits{T.}},
\oauthor{\bsnm{Theobalt}, \binits{C.}},
\oauthor{\bsnm{Wang}, \binits{W.}}:
F2-nerf: Fast neural radiance field training with free camera trajectories.
arXiv preprint arXiv:2303.15951
(2023)
\end{botherref}
\endbibitem

\bibitem[\protect\citeauthoryear{M{\"u}ller et~al.}{2022}]{muller2022instant}
\begin{barticle}
\bauthor{\bsnm{M{\"u}ller}, \binits{T.}},
\bauthor{\bsnm{Evans}, \binits{A.}},
\bauthor{\bsnm{Schied}, \binits{C.}},
\bauthor{\bsnm{Keller}, \binits{A.}}:
\batitle{Instant neural graphics primitives with a multiresolution hash encoding}.
\bjtitle{ACM Transactions on Graphics (ToG)}
\bvolume{41}(\bissue{4}),
\bfpage{1}--\blpage{15}
(\byear{2022})
\end{barticle}
\endbibitem

\bibitem[\protect\citeauthoryear{Wan et~al.}{2023}]{wan2023learning}
\begin{bchapter}
\bauthor{\bsnm{Wan}, \binits{Z.}},
\bauthor{\bsnm{Richardt}, \binits{C.}},
\bauthor{\bsnm{Bo{\v{z}}i{\v{c}}}, \binits{A.}},
\bauthor{\bsnm{Li}, \binits{C.}},
\bauthor{\bsnm{Rengarajan}, \binits{V.}},
\bauthor{\bsnm{Nam}, \binits{S.}},
\bauthor{\bsnm{Xiang}, \binits{X.}},
\bauthor{\bsnm{Li}, \binits{T.}},
\bauthor{\bsnm{Zhu}, \binits{B.}},
\bauthor{\bsnm{Ranjan}, \binits{R.}}, \betal:
\bctitle{Learning neural duplex radiance fields for real-time view synthesis}.
In: \bbtitle{Proceedings of the IEEE/CVF Conference on Computer Vision and Pattern Recognition},
pp. \bfpage{8307}--\blpage{8316}
(\byear{2023})
\end{bchapter}
\endbibitem

\bibitem[\protect\citeauthoryear{Rivas-Manzaneque et~al.}{2023}]{rivasmanzaneque2023}
\begin{bchapter}
\bauthor{\bsnm{Rivas-Manzaneque}, \binits{F.}},
\bauthor{\bsnm{Sierra-Acosta}, \binits{J.}},
\bauthor{\bsnm{Penate-Sanchez}, \binits{A.}},
\bauthor{\bsnm{Moreno-Noguer}, \binits{F.}},
\bauthor{\bsnm{Ribeiro}, \binits{A.}}:
\bctitle{Nerflight: Fast and light neural radiance fields using a shared feature grid}.
In: \bbtitle{Proceedings of the IEEE/CVF Conference on Computer Vision and Pattern Recognition (CVPR)},
pp. \bfpage{12417}--\blpage{12427}
(\byear{2023})
\end{bchapter}
\endbibitem

\bibitem[\protect\citeauthoryear{Ruzzi et~al.}{2023}]{ruzzi2023gazenerf}
\begin{bchapter}
\bauthor{\bsnm{Ruzzi}, \binits{A.}},
\bauthor{\bsnm{Shi}, \binits{X.}},
\bauthor{\bsnm{Wang}, \binits{X.}},
\bauthor{\bsnm{Li}, \binits{G.}},
\bauthor{\bsnm{De~Mello}, \binits{S.}},
\bauthor{\bsnm{Chang}, \binits{H.J.}},
\bauthor{\bsnm{Zhang}, \binits{X.}},
\bauthor{\bsnm{Hilliges}, \binits{O.}}:
\bctitle{Gazenerf: 3d-aware gaze redirection with neural radiance fields}.
In: \bbtitle{Proceedings of the IEEE/CVF Conference on Computer Vision and Pattern Recognition},
pp. \bfpage{9676}--\blpage{9685}
(\byear{2023})
\end{bchapter}
\endbibitem

\bibitem[\protect\citeauthoryear{Siddiqui et~al.}{2023}]{siddiqui2023panoptic}
\begin{bchapter}
\bauthor{\bsnm{Siddiqui}, \binits{Y.}},
\bauthor{\bsnm{Porzi}, \binits{L.}},
\bauthor{\bsnm{Bul{\`o}}, \binits{S.R.}},
\bauthor{\bsnm{M{\"u}ller}, \binits{N.}},
\bauthor{\bsnm{Nie{\ss}ner}, \binits{M.}},
\bauthor{\bsnm{Dai}, \binits{A.}},
\bauthor{\bsnm{Kontschieder}, \binits{P.}}:
\bctitle{Panoptic lifting for 3d scene understanding with neural fields}.
In: \bbtitle{Proceedings of the IEEE/CVF Conference on Computer Vision and Pattern Recognition},
pp. \bfpage{9043}--\blpage{9052}
(\byear{2023})
\end{bchapter}
\endbibitem

\bibitem[\protect\citeauthoryear{Zhang et~al.}{2023a}]{zhang2023beyond}
\begin{bchapter}
\bauthor{\bsnm{Zhang}, \binits{M.}},
\bauthor{\bsnm{Zheng}, \binits{S.}},
\bauthor{\bsnm{Bao}, \binits{Z.}},
\bauthor{\bsnm{Hebert}, \binits{M.}},
\bauthor{\bsnm{Wang}, \binits{Y.-X.}}:
\bctitle{Beyond rgb: Scene-property synthesis with neural radiance fields}.
In: \bbtitle{Proceedings of the IEEE/CVF Winter Conference on Applications of Computer Vision},
pp. \bfpage{795}--\blpage{805}
(\byear{2023})
\end{bchapter}
\endbibitem

\bibitem[\protect\citeauthoryear{Zhang et~al.}{2023b}]{zhang2023nerflets}
\begin{bchapter}
\bauthor{\bsnm{Zhang}, \binits{X.}},
\bauthor{\bsnm{Kundu}, \binits{A.}},
\bauthor{\bsnm{Funkhouser}, \binits{T.}},
\bauthor{\bsnm{Guibas}, \binits{L.}},
\bauthor{\bsnm{Su}, \binits{H.}},
\bauthor{\bsnm{Genova}, \binits{K.}}:
\bctitle{Nerflets: Local radiance fields for efficient structure-aware 3d scene representation from 2d supervision}.
In: \bbtitle{Proceedings of the IEEE/CVF Conference on Computer Vision and Pattern Recognition},
pp. \bfpage{8274}--\blpage{8284}
(\byear{2023})
\end{bchapter}
\endbibitem

\bibitem[\protect\citeauthoryear{Bao et~al.}{2023}]{bao2023sine}
\begin{bchapter}
\bauthor{\bsnm{Bao}, \binits{C.}},
\bauthor{\bsnm{Zhang}, \binits{Y.}},
\bauthor{\bsnm{Yang}, \binits{B.}},
\bauthor{\bsnm{Fan}, \binits{T.}},
\bauthor{\bsnm{Yang}, \binits{Z.}},
\bauthor{\bsnm{Bao}, \binits{H.}},
\bauthor{\bsnm{Zhang}, \binits{G.}},
\bauthor{\bsnm{Cui}, \binits{Z.}}:
\bctitle{Sine: Semantic-driven image-based nerf editing with prior-guided editing field}.
In: \bbtitle{Proceedings of the IEEE/CVF Conference on Computer Vision and Pattern Recognition},
pp. \bfpage{20919}--\blpage{20929}
(\byear{2023})
\end{bchapter}
\endbibitem

\bibitem[\protect\citeauthoryear{Xu et~al.}{2023}]{xu2023discoscene}
\begin{bchapter}
\bauthor{\bsnm{Xu}, \binits{Y.}},
\bauthor{\bsnm{Chai}, \binits{M.}},
\bauthor{\bsnm{Shi}, \binits{Z.}},
\bauthor{\bsnm{Peng}, \binits{S.}},
\bauthor{\bsnm{Skorokhodov}, \binits{I.}},
\bauthor{\bsnm{Siarohin}, \binits{A.}},
\bauthor{\bsnm{Yang}, \binits{C.}},
\bauthor{\bsnm{Shen}, \binits{Y.}},
\bauthor{\bsnm{Lee}, \binits{H.-Y.}},
\bauthor{\bsnm{Zhou}, \binits{B.}}, \betal:
\bctitle{Discoscene: Spatially disentangled generative radiance fields for controllable 3d-aware scene synthesis}.
In: \bbtitle{Proceedings of the IEEE/CVF Conference on Computer Vision and Pattern Recognition},
pp. \bfpage{4402}--\blpage{4412}
(\byear{2023})
\end{bchapter}
\endbibitem

\bibitem[\protect\citeauthoryear{Zhang et~al.}{2023}]{zhang2023ref}
\begin{bchapter}
\bauthor{\bsnm{Zhang}, \binits{Y.}},
\bauthor{\bsnm{He}, \binits{Z.}},
\bauthor{\bsnm{Xing}, \binits{J.}},
\bauthor{\bsnm{Yao}, \binits{X.}},
\bauthor{\bsnm{Jia}, \binits{J.}}:
\bctitle{Ref-npr: Reference-based non-photorealistic radiance fields for controllable scene stylization}.
In: \bbtitle{Proceedings of the IEEE/CVF Conference on Computer Vision and Pattern Recognition},
pp. \bfpage{4242}--\blpage{4251}
(\byear{2023})
\end{bchapter}
\endbibitem

\bibitem[\protect\citeauthoryear{Zheng et~al.}{2023}]{zheng2023editablenerf}
\begin{bchapter}
\bauthor{\bsnm{Zheng}, \binits{C.}},
\bauthor{\bsnm{Lin}, \binits{W.}},
\bauthor{\bsnm{Xu}, \binits{F.}}:
\bctitle{Editablenerf: Editing topologically varying neural radiance fields by key points}.
In: \bbtitle{Proceedings of the IEEE/CVF Conference on Computer Vision and Pattern Recognition},
pp. \bfpage{8317}--\blpage{8327}
(\byear{2023})
\end{bchapter}
\endbibitem

\bibitem[\protect\citeauthoryear{Deng et~al.}{2023}]{deng2023nerdi}
\begin{bchapter}
\bauthor{\bsnm{Deng}, \binits{C.}},
\bauthor{\bsnm{Jiang}, \binits{C.}},
\bauthor{\bsnm{Qi}, \binits{C.R.}},
\bauthor{\bsnm{Yan}, \binits{X.}},
\bauthor{\bsnm{Zhou}, \binits{Y.}},
\bauthor{\bsnm{Guibas}, \binits{L.}},
\bauthor{\bsnm{Anguelov}, \binits{D.}}, \betal:
\bctitle{Nerdi: Single-view nerf synthesis with language-guided diffusion as general image priors}.
In: \bbtitle{Proceedings of the IEEE/CVF Conference on Computer Vision and Pattern Recognition},
pp. \bfpage{20637}--\blpage{20647}
(\byear{2023})
\end{bchapter}
\endbibitem

\bibitem[\protect\citeauthoryear{Li et~al.}{2023}]{li20233d}
\begin{bchapter}
\bauthor{\bsnm{Li}, \binits{S.}},
\bauthor{\bsnm{Weijer}, \binits{J.}},
\bauthor{\bsnm{Wang}, \binits{Y.}},
\bauthor{\bsnm{Khan}, \binits{F.S.}},
\bauthor{\bsnm{Liu}, \binits{M.}},
\bauthor{\bsnm{Yang}, \binits{J.}}:
\bctitle{3d-aware multi-class image-to-image translation with nerfs}.
In: \bbtitle{Proceedings of the IEEE/CVF Conference on Computer Vision and Pattern Recognition},
pp. \bfpage{12652}--\blpage{12662}
(\byear{2023})
\end{bchapter}
\endbibitem

\bibitem[\protect\citeauthoryear{Pavllo et~al.}{2023}]{pavllo2023shape}
\begin{bchapter}
\bauthor{\bsnm{Pavllo}, \binits{D.}},
\bauthor{\bsnm{Tan}, \binits{D.J.}},
\bauthor{\bsnm{Rakotosaona}, \binits{M.-J.}},
\bauthor{\bsnm{Tombari}, \binits{F.}}:
\bctitle{Shape, pose, and appearance from a single image via bootstrapped radiance field inversion}.
In: \bbtitle{Proceedings of the IEEE/CVF Conference on Computer Vision and Pattern Recognition},
pp. \bfpage{4391}--\blpage{4401}
(\byear{2023})
\end{bchapter}
\endbibitem

\bibitem[\protect\citeauthoryear{Yin et~al.}{2023}]{yin2023nerfinvertor}
\begin{bchapter}
\bauthor{\bsnm{Yin}, \binits{Y.}},
\bauthor{\bsnm{Ghasedi}, \binits{K.}},
\bauthor{\bsnm{Wu}, \binits{H.}},
\bauthor{\bsnm{Yang}, \binits{J.}},
\bauthor{\bsnm{Tong}, \binits{X.}},
\bauthor{\bsnm{Fu}, \binits{Y.}}:
\bctitle{Nerfinvertor: High fidelity nerf-gan inversion for single-shot real image animation}.
In: \bbtitle{Proceedings of the IEEE/CVF Conference on Computer Vision and Pattern Recognition},
pp. \bfpage{8539}--\blpage{8548}
(\byear{2023})
\end{bchapter}
\endbibitem

\bibitem[\protect\citeauthoryear{Lin et~al.}{2023}]{lin2023magic3d}
\begin{bchapter}
\bauthor{\bsnm{Lin}, \binits{C.-H.}},
\bauthor{\bsnm{Gao}, \binits{J.}},
\bauthor{\bsnm{Tang}, \binits{L.}},
\bauthor{\bsnm{Takikawa}, \binits{T.}},
\bauthor{\bsnm{Zeng}, \binits{X.}},
\bauthor{\bsnm{Huang}, \binits{X.}},
\bauthor{\bsnm{Kreis}, \binits{K.}},
\bauthor{\bsnm{Fidler}, \binits{S.}},
\bauthor{\bsnm{Liu}, \binits{M.-Y.}},
\bauthor{\bsnm{Lin}, \binits{T.-Y.}}:
\bctitle{Magic3d: High-resolution text-to-3d content creation}.
In: \bbtitle{Proceedings of the IEEE/CVF Conference on Computer Vision and Pattern Recognition},
pp. \bfpage{300}--\blpage{309}
(\byear{2023})
\end{bchapter}
\endbibitem

\bibitem[\protect\citeauthoryear{Liu et~al.}{2023}]{liu2023stylerf}
\begin{bchapter}
\bauthor{\bsnm{Liu}, \binits{K.}},
\bauthor{\bsnm{Zhan}, \binits{F.}},
\bauthor{\bsnm{Chen}, \binits{Y.}},
\bauthor{\bsnm{Zhang}, \binits{J.}},
\bauthor{\bsnm{Yu}, \binits{Y.}},
\bauthor{\bsnm{El~Saddik}, \binits{A.}},
\bauthor{\bsnm{Lu}, \binits{S.}},
\bauthor{\bsnm{Xing}, \binits{E.P.}}:
\bctitle{Stylerf: Zero-shot 3d style transfer of neural radiance fields}.
In: \bbtitle{Proceedings of the IEEE/CVF Conference on Computer Vision and Pattern Recognition},
pp. \bfpage{8338}--\blpage{8348}
(\byear{2023})
\end{bchapter}
\endbibitem

\bibitem[\protect\citeauthoryear{Metzer et~al.}{2023}]{metzer2023latent}
\begin{bchapter}
\bauthor{\bsnm{Metzer}, \binits{G.}},
\bauthor{\bsnm{Richardson}, \binits{E.}},
\bauthor{\bsnm{Patashnik}, \binits{O.}},
\bauthor{\bsnm{Giryes}, \binits{R.}},
\bauthor{\bsnm{Cohen-Or}, \binits{D.}}:
\bctitle{Latent-nerf for shape-guided generation of 3d shapes and textures}.
In: \bbtitle{Proceedings of the IEEE/CVF Conference on Computer Vision and Pattern Recognition},
pp. \bfpage{12663}--\blpage{12673}
(\byear{2023})
\end{bchapter}
\endbibitem

\bibitem[\protect\citeauthoryear{Yu et~al.}{2021}]{yu2021pixelnerf}
\begin{bchapter}
\bauthor{\bsnm{Yu}, \binits{A.}},
\bauthor{\bsnm{Ye}, \binits{V.}},
\bauthor{\bsnm{Tancik}, \binits{M.}},
\bauthor{\bsnm{Kanazawa}, \binits{A.}}:
\bctitle{pixelnerf: Neural radiance fields from one or few images}.
In: \bbtitle{Proceedings of the IEEE/CVF Conference on Computer Vision and Pattern Recognition},
pp. \bfpage{4578}--\blpage{4587}
(\byear{2021})
\end{bchapter}
\endbibitem

\bibitem[\protect\citeauthoryear{Sabour et~al.}{2023}]{sabour2023robustnerf}
\begin{bchapter}
\bauthor{\bsnm{Sabour}, \binits{S.}},
\bauthor{\bsnm{Vora}, \binits{S.}},
\bauthor{\bsnm{Duckworth}, \binits{D.}},
\bauthor{\bsnm{Krasin}, \binits{I.}},
\bauthor{\bsnm{Fleet}, \binits{D.J.}},
\bauthor{\bsnm{Tagliasacchi}, \binits{A.}}:
\bctitle{Robustnerf: Ignoring distractors with robust losses}.
In: \bbtitle{Proceedings of the IEEE/CVF Conference on Computer Vision and Pattern Recognition},
pp. \bfpage{20626}--\blpage{20636}
(\byear{2023})
\end{bchapter}
\endbibitem

\bibitem[\protect\citeauthoryear{Truong et~al.}{2023}]{truong2023sparf}
\begin{bchapter}
\bauthor{\bsnm{Truong}, \binits{P.}},
\bauthor{\bsnm{Rakotosaona}, \binits{M.-J.}},
\bauthor{\bsnm{Manhardt}, \binits{F.}},
\bauthor{\bsnm{Tombari}, \binits{F.}}:
\bctitle{Sparf: Neural radiance fields from sparse and noisy poses}.
In: \bbtitle{Proceedings of the IEEE/CVF Conference on Computer Vision and Pattern Recognition},
pp. \bfpage{4190}--\blpage{4200}
(\byear{2023})
\end{bchapter}
\endbibitem

\bibitem[\protect\citeauthoryear{Huang et~al.}{2004}]{huang2004classification}
\begin{bchapter}
\bauthor{\bsnm{Huang}, \binits{L.-L.}},
\bauthor{\bsnm{Shimizu}, \binits{A.}},
\bauthor{\bsnm{Kobatake}, \binits{H.}}:
\bctitle{Classification-based face detection using gabor filter features}.
In: \bbtitle{Sixth IEEE International Conference on Automatic Face and Gesture Recognition, 2004. Proceedings.},
pp. \bfpage{397}--\blpage{402}
(\byear{2004}).
\bcomment{IEEE}
\end{bchapter}
\endbibitem

\bibitem[\protect\citeauthoryear{Gu et~al.}{2018}]{gu2018recent}
\begin{barticle}
\bauthor{\bsnm{Gu}, \binits{J.}},
\bauthor{\bsnm{Wang}, \binits{Z.}},
\bauthor{\bsnm{Kuen}, \binits{J.}},
\bauthor{\bsnm{Ma}, \binits{L.}},
\bauthor{\bsnm{Shahroudy}, \binits{A.}},
\bauthor{\bsnm{Shuai}, \binits{B.}},
\bauthor{\bsnm{Liu}, \binits{T.}},
\bauthor{\bsnm{Wang}, \binits{X.}},
\bauthor{\bsnm{Wang}, \binits{G.}},
\bauthor{\bsnm{Cai}, \binits{J.}}, \betal:
\batitle{Recent advances in convolutional neural networks}.
\bjtitle{Pattern recognition}
\bvolume{77},
\bfpage{354}--\blpage{377}
(\byear{2018})
\end{barticle}
\endbibitem

\bibitem[\protect\citeauthoryear{Qi et~al.}{2021}]{qi2021survey}
\begin{barticle}
\bauthor{\bsnm{Qi}, \binits{S.}},
\bauthor{\bsnm{Zhang}, \binits{Y.}},
\bauthor{\bsnm{Wang}, \binits{C.}},
\bauthor{\bsnm{Zhou}, \binits{J.}},
\bauthor{\bsnm{Cao}, \binits{X.}}:
\batitle{A survey of orthogonal moments for image representation: theory, implementation, and evaluation}.
\bjtitle{ACM Computing Surveys (CSUR)}
\bvolume{55}(\bissue{1}),
\bfpage{1}--\blpage{35}
(\byear{2021})
\end{barticle}
\endbibitem

\bibitem[\protect\citeauthoryear{Chibane et~al.}{2021}]{chibane2021stereo}
\begin{bchapter}
\bauthor{\bsnm{Chibane}, \binits{J.}},
\bauthor{\bsnm{Bansal}, \binits{A.}},
\bauthor{\bsnm{Lazova}, \binits{V.}},
\bauthor{\bsnm{Pons-Moll}, \binits{G.}}:
\bctitle{Stereo radiance fields (srf): Learning view synthesis for sparse views of novel scenes}.
In: \bbtitle{Proceedings of the IEEE/CVF Conference on Computer Vision and Pattern Recognition},
pp. \bfpage{7911}--\blpage{7920}
(\byear{2021})
\end{bchapter}
\endbibitem

\bibitem[\protect\citeauthoryear{Jensen et~al.}{2014}]{jensen2014large}
\begin{bchapter}
\bauthor{\bsnm{Jensen}, \binits{R.}},
\bauthor{\bsnm{Dahl}, \binits{A.}},
\bauthor{\bsnm{Vogiatzis}, \binits{G.}},
\bauthor{\bsnm{Tola}, \binits{E.}},
\bauthor{\bsnm{Aan{\ae}s}, \binits{H.}}:
\bctitle{Large scale multi-view stereopsis evaluation}.
In: \bbtitle{Proceedings of the IEEE Conference on Computer Vision and Pattern Recognition},
pp. \bfpage{406}--\blpage{413}
(\byear{2014})
\end{bchapter}
\endbibitem

\bibitem[\protect\citeauthoryear{Chen et~al.}{2021}]{chen2021mvsnerf}
\begin{bchapter}
\bauthor{\bsnm{Chen}, \binits{A.}},
\bauthor{\bsnm{Xu}, \binits{Z.}},
\bauthor{\bsnm{Zhao}, \binits{F.}},
\bauthor{\bsnm{Zhang}, \binits{X.}},
\bauthor{\bsnm{Xiang}, \binits{F.}},
\bauthor{\bsnm{Yu}, \binits{J.}},
\bauthor{\bsnm{Su}, \binits{H.}}:
\bctitle{Mvsnerf: Fast generalizable radiance field reconstruction from multi-view stereo}.
In: \bbtitle{Proceedings of the IEEE/CVF International Conference on Computer Vision},
pp. \bfpage{14124}--\blpage{14133}
(\byear{2021})
\end{bchapter}
\endbibitem

\bibitem[\protect\citeauthoryear{Jain et~al.}{2021}]{jain2021putting}
\begin{bchapter}
\bauthor{\bsnm{Jain}, \binits{A.}},
\bauthor{\bsnm{Tancik}, \binits{M.}},
\bauthor{\bsnm{Abbeel}, \binits{P.}}:
\bctitle{Putting nerf on a diet: Semantically consistent few-shot view synthesis}.
In: \bbtitle{Proceedings of the IEEE/CVF International Conference on Computer Vision},
pp. \bfpage{5885}--\blpage{5894}
(\byear{2021})
\end{bchapter}
\endbibitem

\bibitem[\protect\citeauthoryear{Niemeyer et~al.}{2022}]{niemeyer2022regnerf}
\begin{bchapter}
\bauthor{\bsnm{Niemeyer}, \binits{M.}},
\bauthor{\bsnm{Barron}, \binits{J.T.}},
\bauthor{\bsnm{Mildenhall}, \binits{B.}},
\bauthor{\bsnm{Sajjadi}, \binits{M.S.}},
\bauthor{\bsnm{Geiger}, \binits{A.}},
\bauthor{\bsnm{Radwan}, \binits{N.}}:
\bctitle{Regnerf: Regularizing neural radiance fields for view synthesis from sparse inputs}.
In: \bbtitle{Proceedings of the IEEE/CVF Conference on Computer Vision and Pattern Recognition},
pp. \bfpage{5480}--\blpage{5490}
(\byear{2022})
\end{bchapter}
\endbibitem

\bibitem[\protect\citeauthoryear{Yang et~al.}{2023}]{yang2023freenerf}
\begin{botherref}
\oauthor{\bsnm{Yang}, \binits{J.}},
\oauthor{\bsnm{Pavone}, \binits{M.}},
\oauthor{\bsnm{Wang}, \binits{Y.}}:
Freenerf: Improving few-shot neural rendering with free frequency regularization.
arXiv preprint arXiv:2303.07418
(2023)
\end{botherref}
\endbibitem

\bibitem[\protect\citeauthoryear{Alekseev and Bobe}{2019}]{alekseev2019gabornet}
\begin{bchapter}
\bauthor{\bsnm{Alekseev}, \binits{A.}},
\bauthor{\bsnm{Bobe}, \binits{A.}}:
\bctitle{Gabornet: Gabor filters with learnable parameters in deep convolutional neural network}.
In: \bbtitle{2019 International Conference on Engineering and Telecommunication (EnT)},
pp. \bfpage{1}--\blpage{4}
(\byear{2019}).
\bcomment{IEEE}
\end{bchapter}
\endbibitem

\bibitem[\protect\citeauthoryear{Hu et~al.}{2020}]{hu2020gabor}
\begin{barticle}
\bauthor{\bsnm{Hu}, \binits{X.-d.}},
\bauthor{\bsnm{Wang}, \binits{X.-q.}},
\bauthor{\bsnm{Meng}, \binits{F.-j.}},
\bauthor{\bsnm{Hua}, \binits{X.}},
\bauthor{\bsnm{Yan}, \binits{Y.-j.}},
\bauthor{\bsnm{Li}, \binits{Y.-y.}},
\bauthor{\bsnm{Huang}, \binits{J.}},
\bauthor{\bsnm{Jiang}, \binits{X.-l.}}:
\batitle{Gabor-cnn for object detection based on small samples}.
\bjtitle{Defence Technology}
\bvolume{16}(\bissue{6}),
\bfpage{1116}--\blpage{1129}
(\byear{2020})
\end{barticle}
\endbibitem

\bibitem[\protect\citeauthoryear{Gabor}{1946}]{gabor1946theory}
\begin{barticle}
\bauthor{\bsnm{Gabor}, \binits{D.}}:
\batitle{Theory of communication. part 1: The analysis of information}.
\bjtitle{Journal of the Institution of Electrical Engineers-part III: radio and communication engineering}
\bvolume{93}(\bissue{26}),
\bfpage{429}--\blpage{441}
(\byear{1946})
\end{barticle}
\endbibitem

\bibitem[\protect\citeauthoryear{Jain et~al.}{1997}]{jain1997object}
\begin{barticle}
\bauthor{\bsnm{Jain}, \binits{A.K.}},
\bauthor{\bsnm{Ratha}, \binits{N.K.}},
\bauthor{\bsnm{Lakshmanan}, \binits{S.}}:
\batitle{Object detection using gabor filters}.
\bjtitle{Pattern recognition}
\bvolume{30}(\bissue{2}),
\bfpage{295}--\blpage{309}
(\byear{1997})
\end{barticle}
\endbibitem

\bibitem[\protect\citeauthoryear{Kwolek}{2005}]{kwolek2005face}
\begin{bchapter}
\bauthor{\bsnm{Kwolek}, \binits{B.}}:
\bctitle{Face detection using convolutional neural networks and gabor filters}.
In: \bbtitle{Artificial Neural Networks: Biological Inspirations--ICANN 2005: 15th International Conference, Warsaw, Poland, September 11-15, 2005. Proceedings, Part I 15},
pp. \bfpage{551}--\blpage{556}
(\byear{2005}).
\bcomment{Springer}
\end{bchapter}
\endbibitem

\bibitem[\protect\citeauthoryear{Wen et~al.}{2020}]{wen2020gcsba}
\begin{barticle}
\bauthor{\bsnm{Wen}, \binits{Z.}},
\bauthor{\bsnm{Feng}, \binits{R.}},
\bauthor{\bsnm{Liu}, \binits{J.}},
\bauthor{\bsnm{Li}, \binits{Y.}},
\bauthor{\bsnm{Ying}, \binits{S.}}:
\batitle{Gcsba-net: Gabor-based and cascade squeeze bi-attention network for gland segmentation}.
\bjtitle{IEEE Journal of Biomedical and Health Informatics}
\bvolume{25}(\bissue{4}),
\bfpage{1185}--\blpage{1196}
(\byear{2020})
\end{barticle}
\endbibitem

\bibitem[\protect\citeauthoryear{Luan et~al.}{2018}]{luan2018gabor}
\begin{barticle}
\bauthor{\bsnm{Luan}, \binits{S.}},
\bauthor{\bsnm{Chen}, \binits{C.}},
\bauthor{\bsnm{Zhang}, \binits{B.}},
\bauthor{\bsnm{Han}, \binits{J.}},
\bauthor{\bsnm{Liu}, \binits{J.}}:
\batitle{Gabor convolutional networks}.
\bjtitle{IEEE Transactions on Image Processing}
\bvolume{27}(\bissue{9}),
\bfpage{4357}--\blpage{4366}
(\byear{2018})
\end{barticle}
\endbibitem

\bibitem[\protect\citeauthoryear{Sun et~al.}{2020}]{sun2020zernet}
\begin{bchapter}
\bauthor{\bsnm{Sun}, \binits{Z.}},
\bauthor{\bsnm{Rooke}, \binits{E.}},
\bauthor{\bsnm{Charton}, \binits{J.}},
\bauthor{\bsnm{He}, \binits{Y.}},
\bauthor{\bsnm{Lu}, \binits{J.}},
\bauthor{\bsnm{Baek}, \binits{S.}}:
\bctitle{Zernet: Convolutional neural networks on arbitrary surfaces via zernike local tangent space estimation}.
In: \bbtitle{Computer Graphics Forum},
vol. \bseriesno{39},
pp. \bfpage{204}--\blpage{216}
(\byear{2020}).
\bcomment{Wiley Online Library}
\end{bchapter}
\endbibitem

\bibitem[\protect\citeauthoryear{Zernike}{1934}]{zernike1934diffraction}
\begin{barticle}
\bauthor{\bsnm{Zernike}, \binits{F.}}:
\batitle{Diffraction theory of the knife-edge test and its improved form, the phase-contrast method}.
\bjtitle{Monthly Notices of the Royal Astronomical Society, Vol. 94, p. 377-384}
\bvolume{94},
\bfpage{377}--\blpage{384}
(\byear{1934})
\end{barticle}
\endbibitem

\bibitem[\protect\citeauthoryear{Lakshminarayanan and Fleck}{2011}]{lakshminarayanan2011zernike}
\begin{barticle}
\bauthor{\bsnm{Lakshminarayanan}, \binits{V.}},
\bauthor{\bsnm{Fleck}, \binits{A.}}:
\batitle{Zernike polynomials: a guide}.
\bjtitle{Journal of Modern Optics}
\bvolume{58}(\bissue{7}),
\bfpage{545}--\blpage{561}
(\byear{2011})
\end{barticle}
\endbibitem

\bibitem[\protect\citeauthoryear{Fathi et~al.}{2016}]{fathi2016new}
\begin{barticle}
\bauthor{\bsnm{Fathi}, \binits{A.}},
\bauthor{\bsnm{Alirezazadeh}, \binits{P.}},
\bauthor{\bsnm{Abdali-Mohammadi}, \binits{F.}}:
\batitle{A new global-gabor-zernike feature descriptor and its application to face recognition}.
\bjtitle{Journal of Visual Communication and Image Representation}
\bvolume{38},
\bfpage{65}--\blpage{72}
(\byear{2016})
\end{barticle}
\endbibitem

\bibitem[\protect\citeauthoryear{Ding et~al.}{2020}]{ding2020image}
\begin{barticle}
\bauthor{\bsnm{Ding}, \binits{K.}},
\bauthor{\bsnm{Ma}, \binits{K.}},
\bauthor{\bsnm{Wang}, \binits{S.}},
\bauthor{\bsnm{Simoncelli}, \binits{E.P.}}:
\batitle{Image quality assessment: Unifying structure and texture similarity}.
\bjtitle{IEEE transactions on pattern analysis and machine intelligence}
\bvolume{44}(\bissue{5}),
\bfpage{2567}--\blpage{2581}
(\byear{2020})
\end{barticle}
\endbibitem

\bibitem[\protect\citeauthoryear{Camacho-Bello et~al.}{2014}]{camacho2014high}
\begin{barticle}
\bauthor{\bsnm{Camacho-Bello}, \binits{C.}},
\bauthor{\bsnm{Toxqui-Quitl}, \binits{C.}},
\bauthor{\bsnm{Padilla-Vivanco}, \binits{A.}},
\bauthor{\bsnm{B{\'a}ez-Rojas}, \binits{J.}}:
\batitle{High-precision and fast computation of jacobi--fourier moments for image description}.
\bjtitle{JOSA A}
\bvolume{31}(\bissue{1}),
\bfpage{124}--\blpage{134}
(\byear{2014})
\end{barticle}
\endbibitem

\bibitem[\protect\citeauthoryear{Mukundan and Ramakrishnan}{1995}]{mukundan1995fast}
\begin{barticle}
\bauthor{\bsnm{Mukundan}, \binits{R.}},
\bauthor{\bsnm{Ramakrishnan}, \binits{K.}}:
\batitle{Fast computation of legendre and zernike moments}.
\bjtitle{Pattern recognition}
\bvolume{28}(\bissue{9}),
\bfpage{1433}--\blpage{1442}
(\byear{1995})
\end{barticle}
\endbibitem

\bibitem[\protect\citeauthoryear{Daugman}{1987}]{daugman1987image}
\begin{bchapter}
\bauthor{\bsnm{Daugman}, \binits{J.G.}}:
\bctitle{Image analysis and compact coding by oriented 2d gabor primitives}.
In: \bbtitle{Image Understanding and the Man-Machine Interface},
vol. \bseriesno{758},
pp. \bfpage{19}--\blpage{30}
(\byear{1987}).
\bcomment{SPIE}
\end{bchapter}
\endbibitem

\bibitem[\protect\citeauthoryear{Ramasinghe and Lucey}{2022}]{ramasinghe2022beyond}
\begin{bchapter}
\bauthor{\bsnm{Ramasinghe}, \binits{S.}},
\bauthor{\bsnm{Lucey}, \binits{S.}}:
\bctitle{Beyond periodicity: Towards a unifying framework for activations in coordinate-mlps}.
In: \bbtitle{European Conference on Computer Vision},
pp. \bfpage{142}--\blpage{158}
(\byear{2022}).
\bcomment{Springer}
\end{bchapter}
\endbibitem

\bibitem[\protect\citeauthoryear{Chang et~al.}{2015}]{chang2015shapenet}
\begin{botherref}
\oauthor{\bsnm{Chang}, \binits{A.X.}},
\oauthor{\bsnm{Funkhouser}, \binits{T.}},
\oauthor{\bsnm{Guibas}, \binits{L.}},
\oauthor{\bsnm{Hanrahan}, \binits{P.}},
\oauthor{\bsnm{Huang}, \binits{Q.}},
\oauthor{\bsnm{Li}, \binits{Z.}},
\oauthor{\bsnm{Savarese}, \binits{S.}},
\oauthor{\bsnm{Savva}, \binits{M.}},
\oauthor{\bsnm{Song}, \binits{S.}},
\oauthor{\bsnm{Su}, \binits{H.}}, et al.:
Shapenet: An information-rich 3d model repository.
arXiv preprint arXiv:1512.03012
(2015)
\end{botherref}
\endbibitem

\bibitem[\protect\citeauthoryear{Wang et~al.}{2004}]{wang2004image}
\begin{barticle}
\bauthor{\bsnm{Wang}, \binits{Z.}},
\bauthor{\bsnm{Bovik}, \binits{A.C.}},
\bauthor{\bsnm{Sheikh}, \binits{H.R.}},
\bauthor{\bsnm{Simoncelli}, \binits{E.P.}}:
\batitle{Image quality assessment: from error visibility to structural similarity}.
\bjtitle{IEEE transactions on image processing}
\bvolume{13}(\bissue{4}),
\bfpage{600}--\blpage{612}
(\byear{2004})
\end{barticle}
\endbibitem

\bibitem[\protect\citeauthoryear{Zhang et~al.}{2018}]{zhang2018unreasonable}
\begin{bchapter}
\bauthor{\bsnm{Zhang}, \binits{R.}},
\bauthor{\bsnm{Isola}, \binits{P.}},
\bauthor{\bsnm{Efros}, \binits{A.A.}},
\bauthor{\bsnm{Shechtman}, \binits{E.}},
\bauthor{\bsnm{Wang}, \binits{O.}}:
\bctitle{The unreasonable effectiveness of deep features as a perceptual metric}.
In: \bbtitle{Proceedings of the IEEE Conference on Computer Vision and Pattern Recognition},
pp. \bfpage{586}--\blpage{595}
(\byear{2018})
\end{bchapter}
\endbibitem

\bibitem[\protect\citeauthoryear{Barron et~al.}{2021}]{barron2021mip}
\begin{bchapter}
\bauthor{\bsnm{Barron}, \binits{J.T.}},
\bauthor{\bsnm{Mildenhall}, \binits{B.}},
\bauthor{\bsnm{Tancik}, \binits{M.}},
\bauthor{\bsnm{Hedman}, \binits{P.}},
\bauthor{\bsnm{Martin-Brualla}, \binits{R.}},
\bauthor{\bsnm{Srinivasan}, \binits{P.P.}}:
\bctitle{Mip-nerf: A multiscale representation for anti-aliasing neural radiance fields}.
In: \bbtitle{Proceedings of the IEEE/CVF International Conference on Computer Vision},
pp. \bfpage{5855}--\blpage{5864}
(\byear{2021})
\end{bchapter}
\endbibitem

\end{thebibliography}

\clearpage
\title{Supplementary material: MomentsNeRF}
\setcounter{page}{1}
\maketitle

\section{Moments}
\begin{figure}[htb]
    \centering
    \includegraphics[width=.7\textwidth]{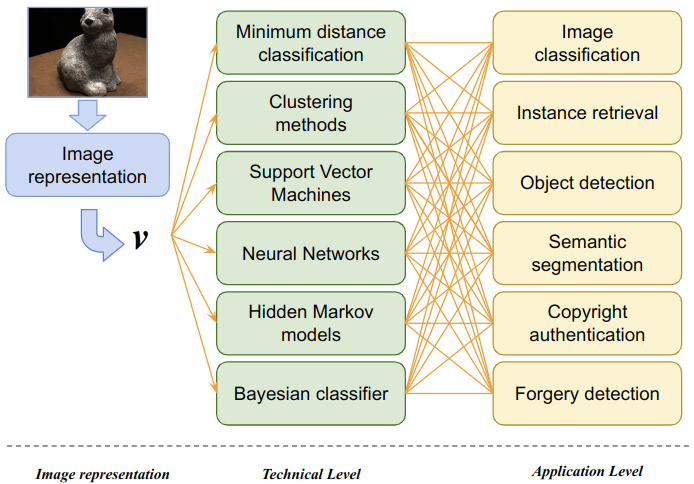}
    \caption{
   The representation of digital images plays a crucial role because a typical visual system is formed by creating a meaningful representation of the image, followed by applying techniques for knowledge extraction (at the technical level), and finally, achieving a high-level visual understanding (at the application level).}
    \label{fig:image_representation}
\end{figure}

In various applications of computer vision and pattern recognition, it is necessary to have a representation of a digital image that possesses meaningful, readily apparent semantic characteristics. This representation process, known as image representation, involves extracting features from the image and their translation into a format that machine learning algorithms can understand. The quality of the image representation can significantly affect the system's performance in tasks such as image classification, object detection, and image segmentation. Therefore, image representation is a fundamental aspect of computer vision and pattern recognition that requires careful consideration and optimization, as represented in Fig.~\ref{fig:image_representation}.

\begin{figure*}[htb]
    \centering
    \includegraphics[width=1.\textwidth]{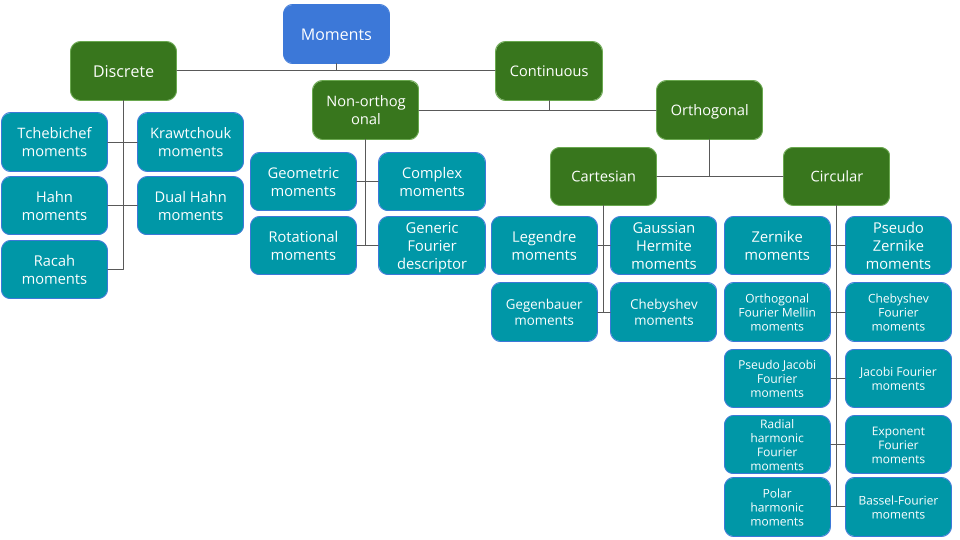}
    \caption{Image Moments Categories 
    }
    \label{fig:image_moments_family}
\end{figure*}

The inner product $\left \langle f, V_{nm}\right \rangle$ of the image function $f$ and the basis function $V_{nm}$ of $(n+m)$ order domain $D$ is represented as an image moment 
\begin{equation}
    \left \langle f, V_{nm} \right \rangle = \iint_{D}^{} V^*_{nm},(x,y)f(x,y)dxdy
\end{equation}
where the complex conjugate is referred by the asterisk $^*$ notation. The direct geometric interpretation of image moment set $\left \langle f, V_{nm} \right \rangle$ is that it is the projection of $f$ onto a subspace formed by a set of basis functions $\left \{ V_{nm}: (n,m) \in \mathbb{Z}^2 \right \}$. It is necessary to manually design a set of special basis functions $V_{nm}$ with valuable properties in $\left \langle f, V_{nm} \right \rangle$ which meet the semantic requirements because of the infinite number of basis function sets. The image moments family is divided into different categories due to the mathematical properties of basis functions, as shown in Fig.~\ref{fig:image_moments_family}.
Firstly, the image moments are orthogonal moments if the basis functions satisfy the \textit{orthogonality} property. The orthogonality defines that any two different basis functions $V_{nm}$ and $V_{n'm'}$ from the basis function set are “perpendicular” or they are uncorrelated in geometric concept, resulting to no redundancy in the moment set. More precisely, $V_{nm}$ and $V_{n'm'}$ are orthogonal when satisfying the following condition:
\begin{equation}
    \left \langle V_{nm}, V_{n'm'} \right \rangle = \iint_{D}^{} V_{nm}(x,y)V^*_{n'm'}(x,y)dxdy = \delta_{nn'}\delta_{mm'}
\end{equation}
where the Kronecker delta function ($\delta_{ij}$) is defined as:
\begin{equation}
\delta_{ij}=\left\{\begin{matrix}
0 & i \neq j, \\ 
1 & i = j.
\end{matrix}\right.
\end{equation}
The basis functions used for image reconstruction suffer from high information redundancy due to their non-orthogonality. This, in turn, leads to poor discriminability and robustness in image representation. Therefore, it is essential to ensure orthogonality when designing basis functions.
Secondly, image moments can be classified as continuous or discrete depending on the \textit{continuity} of the basis functions. Finally, image moments can be grouped into Cartesian and circular moments depending on the \textit{coordinate system} that defines the basis functions.

\subsection{Gabor Filter}
\begin{figure}[htb]
    \centering
    \includegraphics[width=.7\linewidth]{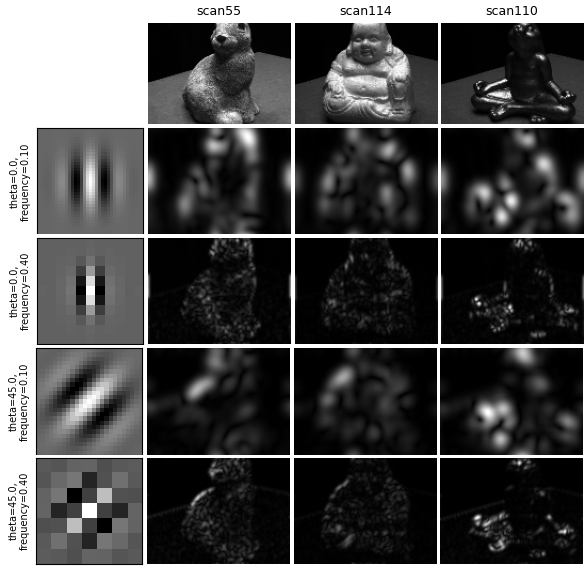}
    \caption{We demonstrated the impact of Gabor filters on various scenes from the DTU dataset by randomly selecting the frequency and the orientation representation for each image. On the left, we show the Gabor filter bank, while on the right we show the images after applying a Gabor filter on them.}
    \label{fig:gabor_filter_samples}
\end{figure}

In Fig.~\ref{fig:gabor_filter_samples}, we show the images before and after applying Gabor filters on three selected scenes from DTU MVS dataset. We randomly selected the frequency and the orientation representation for each image.  

\subsection{Zernike Polynomials} 
In Fig.~\ref{fig:zernike_even_odd_ploy_3d_complete}, we show the complete set of Zernike Polynomials with 3D representation and classical names. Notably, we do not show $Z_{2}^{0}$ and $Z_{4}^{0}$ polynomials since they are duplicated with $Z_{0}^{0}$.

\subsubsection{Coordinate Mapping:}
\begin{figure*}[htb]
    \centering
    \includegraphics[width=1.\textwidth]{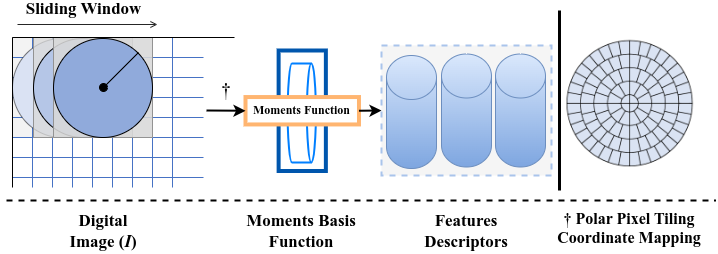}
    \caption{Coordinate mapping between the square region of the digital image and the unit disk to derive feature descriptors using a specific moment basis function. 
    }
    \label{fig:image_moments}
\end{figure*}

Zernike polynomials are mapped onto a pixel grid using polar pixel tilling coordinate mapping. These orthogonal functions are defined over a circular domain and are commonly used in optical and wavefront aberration analysis, as shown in Fig.~\ref{fig:image_moments}.

\begin{figure}[htb]
    \centering
    \includegraphics[width=1.\textwidth]{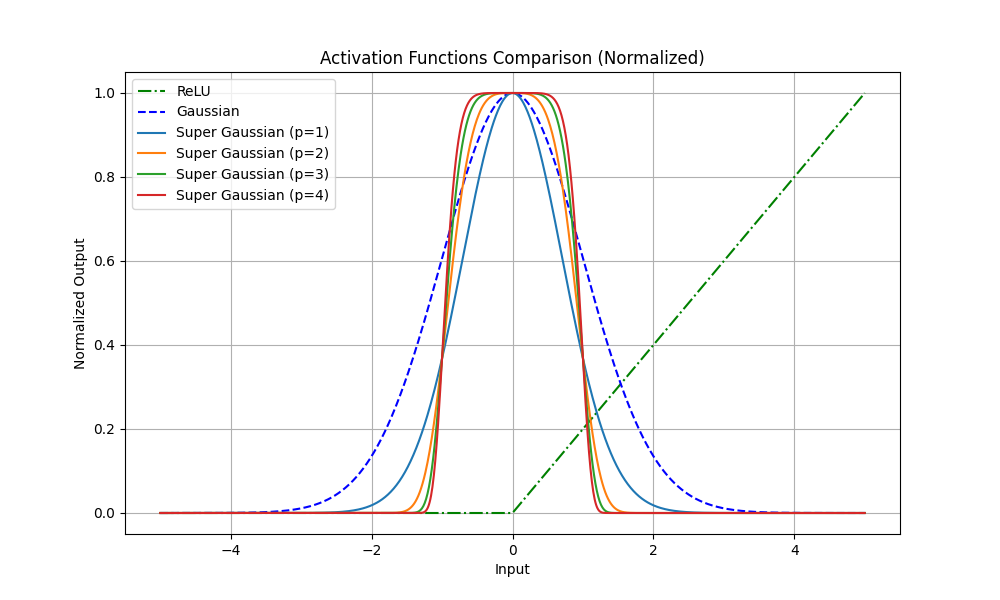}
    \caption{A comparison of RelU, Gaussian, and Super-Gaussian activation functions with normalized output for better visualizations.}
    \label{fig:normalized_activation_functions}
\end{figure}

\begin{figure}[htb]
     \centering\setlength{\tabcolsep}{1pt}
     \begin{center}
    \begin{tabular}{ccccc}
      \begin{subfigure}[t]{.2\linewidth}
         \centering
         \includegraphics[width=\textwidth]{zernike_3d_0-0-even.png}
         \setcounter{subfigure}{0}      
         \caption{\(Z_{0}^{0}\)(Even)\\Piston}
         \label{fig:z_0_0_3d_even}
     \end{subfigure} 
     &
     \begin{subfigure}[t]{.2\linewidth}
         \centering
         \includegraphics[width=\textwidth]{zernike_3d_1-1-even.png}
         \setcounter{subfigure}{1}     
         \caption{ \(Z_{1}^{1} (Even)\)\\Vertical tilt}
         \label{fig:z_1_1_3d_odd}
     \end{subfigure} 
     &
     \begin{subfigure}[t]{.2\linewidth}
         \centering
         \includegraphics[width=\textwidth]{zernike_3d_1-1-odd.png}
         \setcounter{subfigure}{2}     
         \caption{\(Z_{1}^{1} (Odd)\)\\Horizontal tilt}
         \label{fig:z_2_1_3d_odd}
     \end{subfigure} 
     &
      \begin{subfigure}[t]{.2\linewidth}
         \centering
         \includegraphics[width=\textwidth]{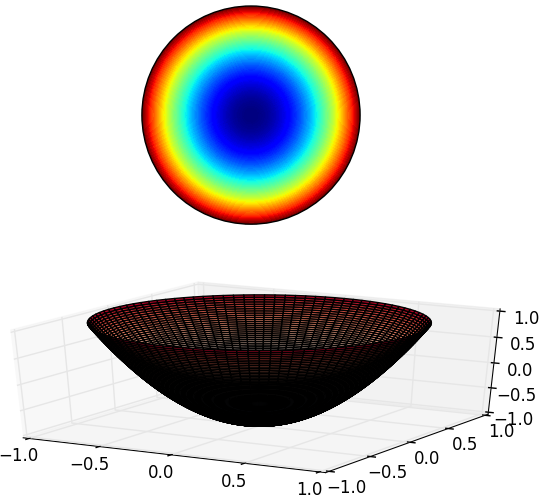}
         \setcounter{subfigure}{3}   
         \caption{\(Z_{2}^{0} (Even)\)\\Oblique astigmatism}
         \label{fig:z_2_0_3d_even}
     \end{subfigure}
     &
     \begin{subfigure}[t]{.2\linewidth}
         \centering
         \includegraphics[width=\textwidth]{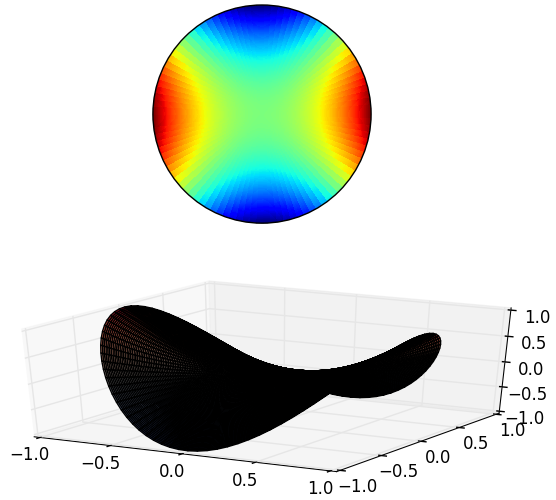}
         \setcounter{subfigure}{5}
         \caption{\(Z_{2}^{2} (Even)\)\\Defocus}
         \label{fig:z_2_2_3d_even}
     \end{subfigure}
     \\
     \begin{subfigure}[t]{.2\linewidth}
         \centering
         \includegraphics[width=\textwidth]{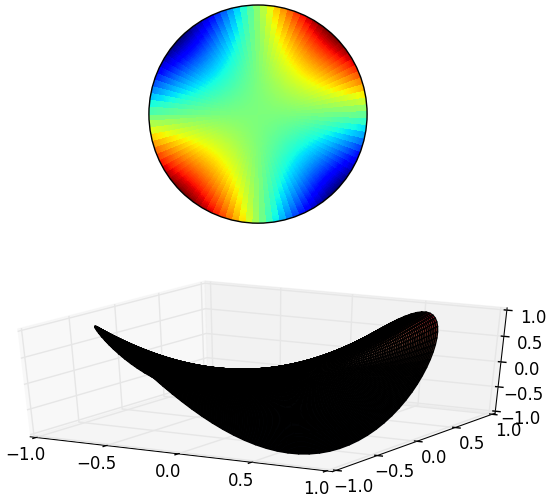}
         \setcounter{subfigure}{6}     
         \caption{ \(Z_{2}^{2} (Odd)\)\\Vertical astigmatism}
         \label{fig:z_2_2_3d}
     \end{subfigure} 
     &
     \begin{subfigure}[t]{.2\linewidth}
         \centering
         \includegraphics[width=\textwidth]{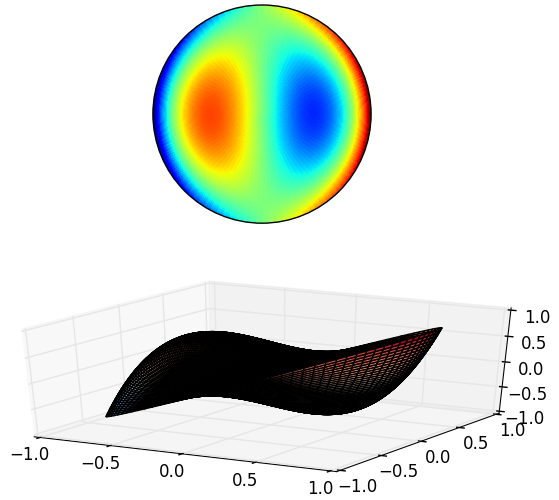}
         \setcounter{subfigure}{7}     
         \caption{\(Z_{3}^{1} (Even) \)\\Vertical trefoil}
         \label{fig:z_3_1_3d}
     \end{subfigure} 
     &
     \begin{subfigure}[t]{.2\linewidth}
         \centering
         \includegraphics[width=\textwidth]{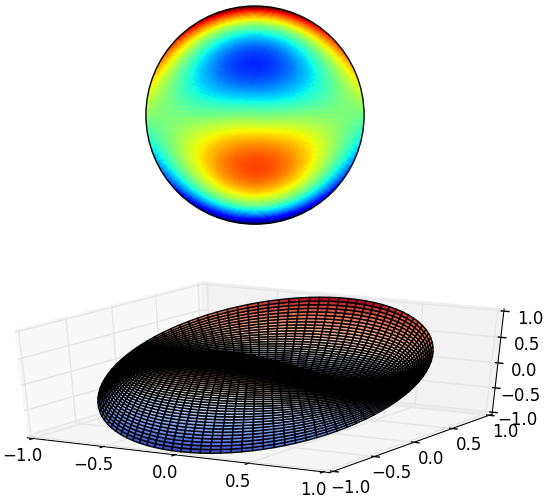}
         \setcounter{subfigure}{8}     
         \label{fig:z_3_1_3d}
         \caption{\(Z_{3}^{1} (Odd)\)\\Vertical coma}
     \end{subfigure}
     &
     \begin{subfigure}[t]{.2\linewidth}
         \centering
         \includegraphics[width=\textwidth]{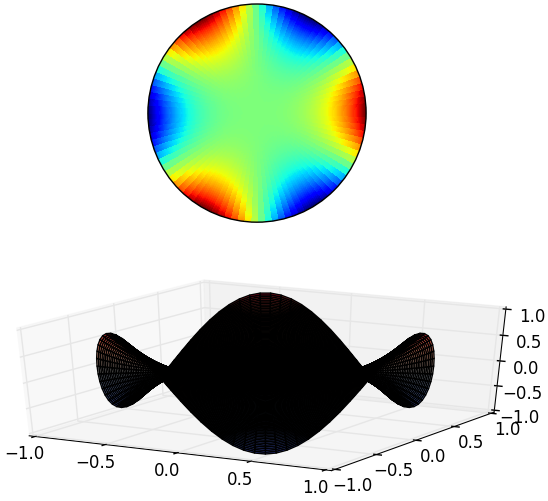}
         \setcounter{subfigure}{9}     
         \label{fig:z_3_3_3d}
         \caption{\(Z_{3}^{3} (Even)\)\\Horizontal coma}
     \end{subfigure} 
     &
     \begin{subfigure}[t]{.2\linewidth}
         \centering
         \includegraphics[width=\textwidth]{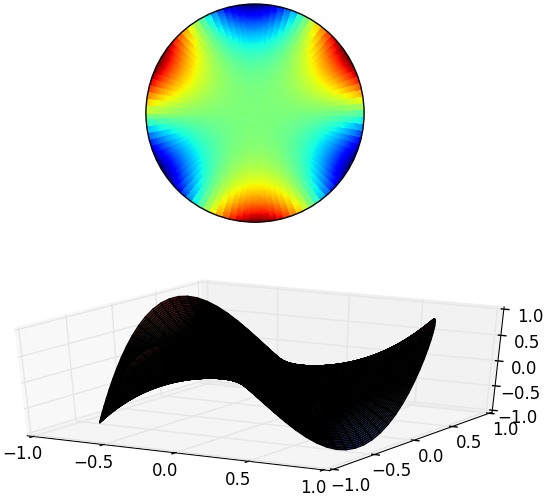}
         \setcounter{subfigure}{10}     
         \label{fig:z_3_3_3d}
         \caption{\(Z_{3}^{3} (Odd)\)\\Oblique trefoil}
     \end{subfigure} 
     \\
     \begin{subfigure}[t]{.2\linewidth}
         \centering
         \includegraphics[width=\textwidth]{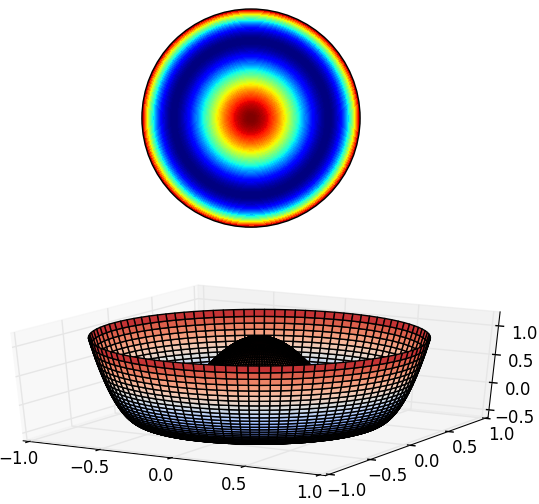}
         \setcounter{subfigure}{11}     
         \label{fig:z_4_0_3d}
         \caption{\(Z_{4}^{0} (Even)\)\\Oblique quadrafoil}
     \end{subfigure} 
     &
     \begin{subfigure}[t]{.20\linewidth}
         \centering
         \includegraphics[width=\textwidth]{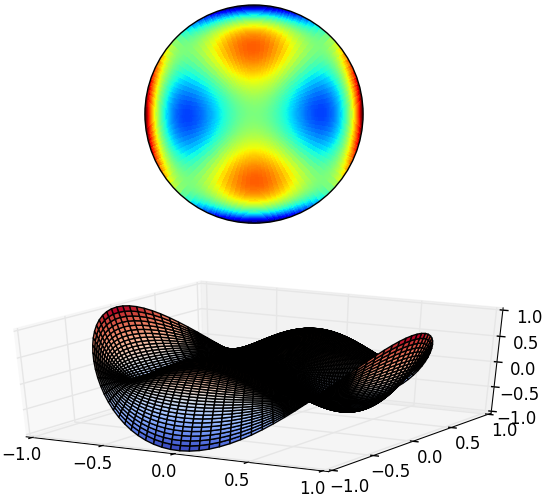}
         \setcounter{subfigure}{13}     
         \label{fig:z_4_2_3d}
         \caption{\(Z_{4}^{2} (Even)\)\\Oblique secondary astigmatism}
     \end{subfigure} 
     &
     \begin{subfigure}[t]{.20\linewidth}
         \centering
         \includegraphics[width=\textwidth]{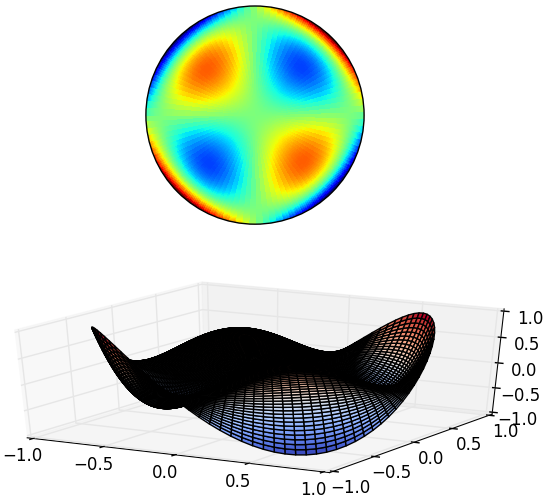}
         \setcounter{subfigure}{14}     
         \label{fig:z_4_2_3d}
         \caption{\(Z_{4}^{2} (Odd)\)\\Primary spherical}
     \end{subfigure}
     &
     \begin{subfigure}[t]{.20\linewidth}
         \centering
         \includegraphics[width=\textwidth]{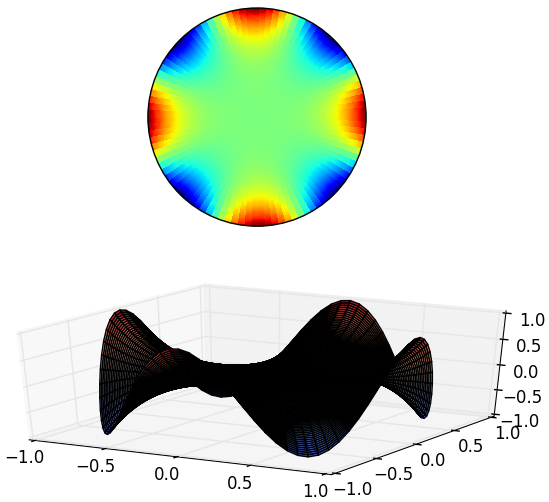}
         \setcounter{subfigure}{15}     
         \label{fig:z_4_4_3d}
         \caption{\(Z_{4}^{4} (Even)\)\\Vertical secondary astigmatism}
     \end{subfigure} 
     &
     \begin{subfigure}[t]{.20\linewidth}
         \centering
         \includegraphics[width=\textwidth]{zernike_3d_4-4-odd.png}
         \setcounter{subfigure}{16}     
         \label{fig:z_4_4_3d}
         \caption{\(Z_{4}^{4} (Odd)\)\\Vertical quadrafoil}
     \end{subfigure} 
\end{tabular}
        \caption{Odd and Even 2D \& 3D plots for each Zernike polynomials,  labelled with classical names. 
        } 
        \label{fig:zernike_even_odd_ploy_3d_complete}
        \end{center}
\end{figure}

\section{Results}
\label{sec:rationale}
\begin{figure}[htb]
    \centering
    \includegraphics[width=.9\textwidth]{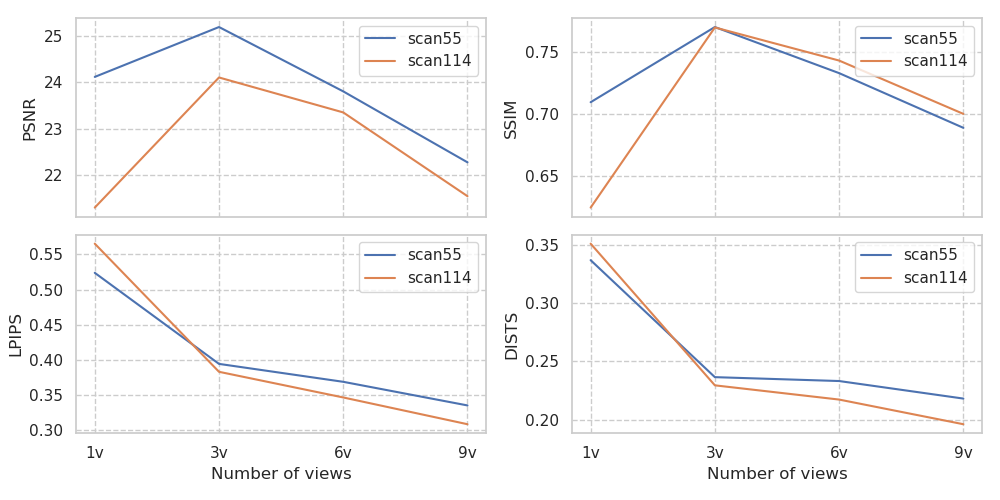}
    \caption{Results of Fig. 7 (main paper) scenes for each quality metric and how it depends on the number of views.}
    \label{fig:ablation_study_qualitative}
\end{figure}

We demonstrate more qualitative and quantitative results on the DTU dataset; in Fig. 7 (main paper)
, we present the ablation study, where we test the scene on different views settings. The associated qualitative results are shown in Fig.~\ref{fig:ablation_study_qualitative}. Moreover, Table.~\ref{table:figs_quantitive_res} shows the qualitative results of different figures in the main paper. Fig.~\ref{fig:1v_cars_chairs_results} shows more qualitative results for one-shot neural rendering on two category-specific datasets from ShapeNet. Similarly, Fig.~\ref{fig:2v_cars_chairs_results} shows two-shot neural rendering on the same datasets.

Similarly to PixelNeRF, our framework degrades PSNR and SSIM quality metrics performance. Unlike the bad exposure reason mentioned in PixelNeRF's evaluation protocol, we did a deeper analysis, and we found the degradation is related to sampling the corresponding image feature via projection and bilinear interpolation of the image features, as shown in Fig.~\ref{fig:ablation_study_qualitative}. 
Briefly, the MLP receives the query specification and image features as inputs and produces outputs for density and color. The image features are passed to each MLP layer as a residual. If multiple images are available, they are first encoded into a latent representation in each camera’s coordinate frame and combined in an intermediate layer before predicting the color and density. The models are trained using a reconstruction loss between a ground truth image and a view created through conventional volume rendering techniques. This approach caused instability on the test time, especially when you have more images on test time than training time. We choose three images from each scene to train the models on the DTU MVS dataset, while on test time, we chose different settings as explained in the main paper. Finally, the ideal performance is achieved when training and testing images are equal. 
A potential solution that can be considered to solve this issue is training MomentsNeRF on different numbers of images or considering different projection and interpolation methods. We are targeting this issue as a future work to have a more stable few-shot neural rendering performance. 
\begin{figure}
    \centering
    \includegraphics[width=.9\textwidth]{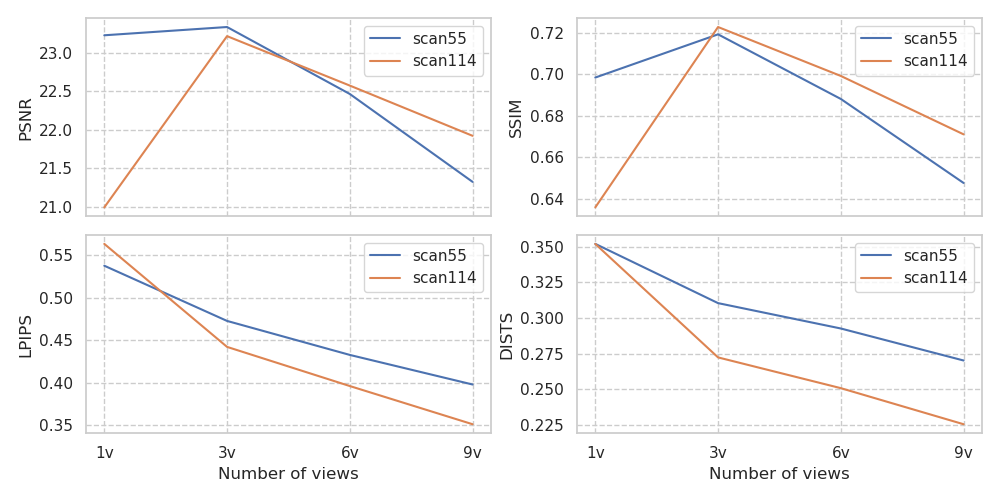}
    \caption{Results of Fig. 7 (main paper) scenes for each quality metric as a function of number of views using PixelNeRF.}
    \label{fig:ablation_study_qualitative_pixel}
\end{figure}

\begin{table}[htb]
\caption{A quantitative comparisons on the DTU dataset for each figure. The ID column refers to scene ID. The best results are marked with red, while the second best results marked with orange.}
\setlength{\tabcolsep}{1pt}
\centering
\begin{tabular}{c|c|c|cccc}
\hline
ID & Fig. & Views & PSNR↑   & SSIM↑  & LPIPS↓ & DISTS↓ \\
\hline
\multirow{2}{*}{21} & ~\ref{fig:scan21_1v_a} & 1 & \cellcolor{orange!25} 20.847 & \cellcolor{orange!25} 0.579 & \cellcolor{orange!25} 0.501 & \cellcolor{orange!25} 0.24 \\
& ~\ref{fig:scan21_1v_b} & 1 & \cellcolor{red!25} 21.572 & \cellcolor{red!25} 0.646 & \cellcolor{red!25} 0.456 & \cellcolor{red!25} 0.22 \\
\hline
\multirow{2}{*}{110} &~\ref{fig:scan110_3v_a} & 3 & \cellcolor{orange!25} 28.115 & \cellcolor{orange!25} 0.847 & \cellcolor{orange!25} 0.372 & \cellcolor{orange!25} 0.246 \\
& ~\ref{fig:scan110_3v_b} & 3 & \cellcolor{red!25} 31.292 & \cellcolor{red!25} 0.909 & \cellcolor{red!25} 0.335 & \cellcolor{red!25} 0.204 \\
\hline
\multirow{2}{*}{63} & ~\ref{fig:scan63_6v_a} & 6 & \cellcolor{orange!25} 25.531 & \cellcolor{orange!25} 0.918 & \cellcolor{orange!25} 0.193 & \cellcolor{orange!25} 0.235 \\
& ~\ref{fig:scan63_6v_b} & 6 & \cellcolor{red!25} 27.012 & \cellcolor{red!25} 0.925 & \cellcolor{red!25} 0.179 & \cellcolor{red!25} 0.212 \\
\hline
\multirow{2}{*}{114} & ~\ref{fig:scan114_9v_a} & 9 & \cellcolor{red!25} 28.874 & \cellcolor{orange!25} 0.859 & \cellcolor{orange!25} 0.317 & \cellcolor{orange!25} 0.20 \\
& ~\ref{fig:scan114_9v_b} & 9 & \cellcolor{orange!25} 28.253 & \cellcolor{red!25} 0.879 & \cellcolor{red!25} 0.279 & \cellcolor{red!25} 0.166 \\
\hline
\end{tabular}
\label{table:figs_quantitive_res}
\end{table}

 \begin{figure}[htb]
     \centering\setlength{\tabcolsep}{1pt}
    \begin{tabular}{cccc}
     Input & PixelNeRF & Ours & Reference \\ 
       \begin{subfigure}[b]{.19\linewidth}
      \captionsetup[figure]{font=footnotesize,labelfont=footnotesize}
         \centering
         \includegraphics[width=.675\textwidth]{comparisons_scan21_1v_000023.png}
         \setcounter{subfigure}{0}
         \caption{ }
         \label{fig:scene55_3v_in}
     \end{subfigure}
      &
      \begin{subfigure}[b]{.27\linewidth}
         \centering
         \includegraphics[width=\textwidth]{comparisons_scan63_6v_fig_sample_a.png}           \setcounter{subfigure}{1}%
         \caption{ }
         \label{fig:scan21_1v_a}
     \end{subfigure} &
     \begin{subfigure}[b]{.27\linewidth}
         \centering
         \includegraphics[width=\textwidth]{comparisons_scan63_6v_fig_sample_b.png}
         \setcounter{subfigure}{2}%
         \caption{ }
         \label{fig:scan21_1v_b}
     \end{subfigure} 
     &
     \begin{subfigure}[b]{.27\linewidth}
         \centering
         \includegraphics[width=\textwidth]{comparisons_scan63_6v_fig_sample_c.png}
         \setcounter{subfigure}{3}%
         \caption{}
         \label{fig:scan21_1v_c}
     \end{subfigure} \\
     \begin{subfigure}[b]{.19\linewidth}
         \centering
         \includegraphics[width=.675\textwidth]{comparisons_scan21_1v_000023.png}
          \includegraphics[width=.675\textwidth]{comparisons_scan63_6v_000032.png}
          \includegraphics[width=.675\textwidth]{comparisons_scan110_3v_000046.png}
         \setcounter{subfigure}{4}%
         \caption{ }
         \label{fig:scan110_3v_a}
     \end{subfigure} &
      \begin{subfigure}[b]{.27\linewidth}
         \centering
         \includegraphics[width=\textwidth]{comparisons_scan63_6v_fig_sample_a.png}
         \setcounter{subfigure}{5}%
         \caption{ }
         \label{fig:scan110_3v_a}
     \end{subfigure} &
     \begin{subfigure}[b]{.27\linewidth}
         \centering
         \includegraphics[width=\textwidth]{comparisons_scan63_6v_fig_sample_b.png}
         \setcounter{subfigure}{6}%
         \caption{ }
         \label{fig:scan110_3v_b}
     \end{subfigure} 
     &
     \begin{subfigure}[b]{.27\linewidth}
         \centering
         \includegraphics[width=\textwidth]{comparisons_scan63_6v_fig_sample_c.png}
         \setcounter{subfigure}{7}%
         \caption{ }
         \label{fig:scan110_3v_c}
     \end{subfigure} \\
     \begin{subfigure}[b]{.19\linewidth}
         \centering
         \includegraphics[width=.45\textwidth]{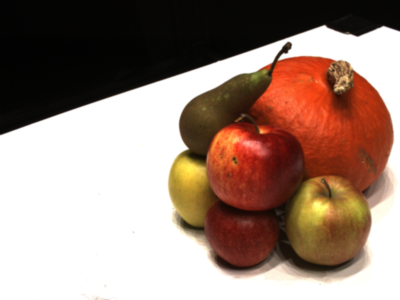}
         \includegraphics[width=.45\textwidth]{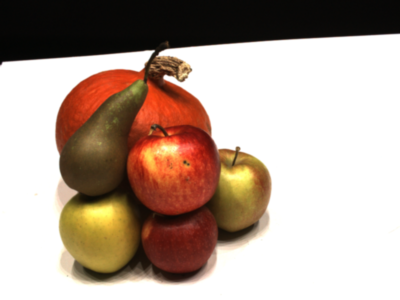}
         \includegraphics[width=.45\textwidth]{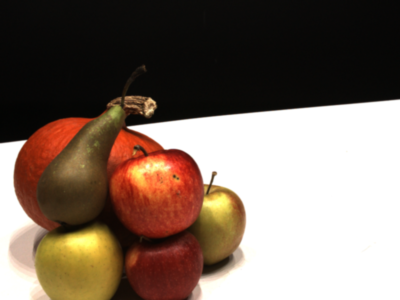}
          \includegraphics[width=.45\textwidth]{comparisons_scan63_6v_000032.png}
         \includegraphics[width=.45\textwidth]{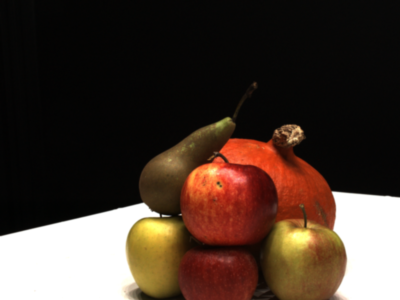}
         \includegraphics[width=.45\textwidth]{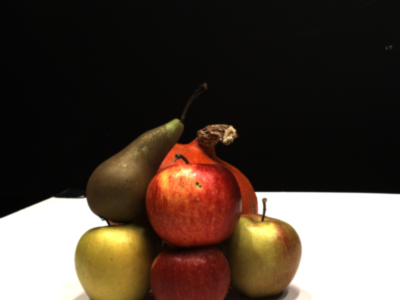}
         \setcounter{subfigure}{8}%
         \caption{ }
         \label{fig:scan63_6v_a}
     \end{subfigure} &
     \begin{subfigure}[b]{.27\linewidth}
         \centering
         \includegraphics[width=\textwidth]{comparisons_scan63_6v_fig_sample_a.png}
         \setcounter{subfigure}{9}%
         \caption{ }
         \label{fig:scan63_6v_a}
     \end{subfigure} &
     \begin{subfigure}[b]{.27\linewidth}
         \centering
         \includegraphics[width=\textwidth]{comparisons_scan63_6v_fig_sample_b.png}
         \setcounter{subfigure}{10}%
         \caption{ }
         \label{fig:scan63_6v_b}
     \end{subfigure} 
     &
     \begin{subfigure}[b]{.27\linewidth}
         \centering
         \includegraphics[width=\textwidth]{comparisons_scan63_6v_fig_sample_c.png}
         \setcounter{subfigure}{11}%
         \caption{ }
         \label{fig:scan63_6v_c}
     \end{subfigure} \\
     \begin{subfigure}[b]{.19\linewidth}
         \centering
         \includegraphics[width=.3\textwidth]{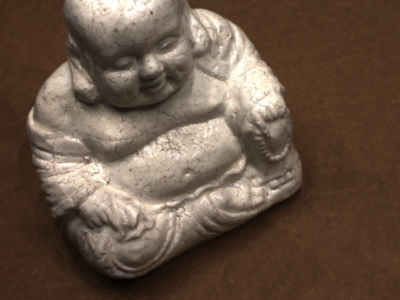}
         \includegraphics[width=.3\textwidth]{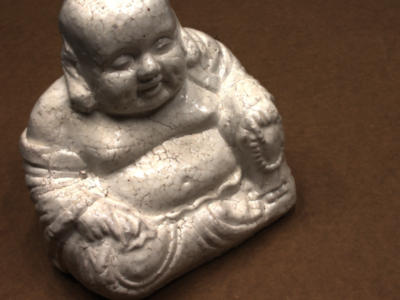}
          \includegraphics[width=.3\textwidth]{comparisons_scan21_1v_000023.png}
          \includegraphics[width=.3\textwidth]{comparisons_scan63_6v_000026.png}
          \includegraphics[width=.3\textwidth]{comparisons_scan63_6v_000029.png}
          \includegraphics[width=.3\textwidth]{comparisons_scan63_6v_000032.png}
          \includegraphics[width=.3\textwidth]{comparisons_scan63_6v_000043.png}
          \includegraphics[width=.3\textwidth]{comparisons_scan63_6v_000045.png}
           \includegraphics[width=.3\textwidth]{103b75dfd146976563ed57e35c972b4b_gt_000048.png}
         \setcounter{subfigure}{12}%
         \caption{ }
         \label{fig:scan114_9v_a}
     \end{subfigure} &
     \begin{subfigure}[b]{.27\linewidth}
         \centering
         \includegraphics[width=\textwidth]{comparisons_scan63_6v_fig_sample_a.png}
         \setcounter{subfigure}{13}%
         \caption{ }
         \label{fig:scan114_9v_a}
     \end{subfigure} &
     \begin{subfigure}[b]{.27\linewidth}
         \centering
         \includegraphics[width=\textwidth]{comparisons_scan63_6v_fig_sample_b.png}
         \setcounter{subfigure}{14}%
         \caption{ }
         \label{fig:scan114_9v_b}
     \end{subfigure} 
     &
     \begin{subfigure}[b]{.27\linewidth}
         \centering
         \includegraphics[width=\textwidth]{comparisons_scan63_6v_fig_sample_c.png}
         \setcounter{subfigure}{15}%
         \caption{ }
         \label{fig:scan114_9v_c}
     \end{subfigure} 

\end{tabular}
        \caption{Four qualitative comparisons on DTU dataset. Each comparison shows the difference between PixelNeRF, MomentsNeRF, and the Reference image (in order from left to right). Our results show better details for the texture, and reducing noise and artifacts, and completing the missing parts of the scene. The quantitative results can be found in Table. ~\ref{table:figs_quantitive_res}.
        }
        \label{fig:selective_qualitative_comparison_on_DTU}
\end{figure}

\section{Positional Encoding}
We omit $pi$ in positional encoding function similar to NeRF \cite{mildenhall2021nerf} implementation:
\begin{equation}
    \gamma(p) = (sin(2^0p), cos(2^0p), ..., sin(2^{L-1}p), cos(2^{L-1}p)).
\end{equation}
where the function $\gamma (.)$ is used on each of the three coordinate values in $\mbb{x}$, which have been normalized to the range of $[-1.5, 1.5]$. It is also applied to the three components of the Cartesian viewing direction unit vector $\mbb{d}$, which also lie within the range of $[-1.5, 1.5]$.

\section{Super Gaussian Activation}
Our MLP uses the Super Gaussian Activation function \cite{ramasinghe2022beyond} instead of ReLU. The Super Gaussian Activation function is a specific activation function derived from the Gaussian function. The key difference is that it has heavier tails due to a higher degree of kurtosis. This function is commonly used in neural networks to introduce non-linearity into the model and improve its accuracy. More briefly, the Gaussian function is defined as:
\begin{equation}
    f(t) = e^{-\frac{t^2}{2\bar\sigma^2}}
\end{equation}
where $t$ is a given input tensor; $\bar \sigma$ is the spread of the function. It is akin to the standard deviation in the Gaussian function. Furthermore, the Super Gaussian function can be defined as:
\begin{equation}
    f(t) = e^{-\left(\frac{t}{\bar\sigma}\right)^{2p}}
\end{equation}
where $p$ is the degree of kurtosis or the shape of the Gaussian function. Notably, a higher value of $p$ leads to heavier tails, making the function more peaked. Fig. \ref{fig:normalized_activation_functions} compares the activation functions. We set $p = 4$ in our implementation.

\subsection{Evaluation Protocol}
Similarly to PixelNeRF, to evaluate our system, we individually used 1, 3, 6, and 9 informative input views and calculated image metrics with the remaining views. We selected views 25, 22, 28, 40, 44, 48, 0, 8, and 13 for input, and when less than 9 views were used, we took a prefix of this list. Furthermore, during testing, we did exclude the views with bad exposure from MVS DTU dataset, such as (3, 4, 5, 6, 7, 16, 17, 18, 19, 20, 21, 36, 37, 38, 39). Our results show that MomentsNeRF works perfectly with these bad exposures. Moreover, PixelNeRF results in our paper are reproduced to accept the new quality metric.  
For a deeper analysis, we randomly changed the view settings for both PixelNeRF and MomentsNeRF; we used fixed settings for the views as: `23' for one view, `23 32 46' for three views, `23 26 29 32 43 45' for six views, and `23 26 29 32 43 45 48 9 13' for nine views settings. Table 4 (main paper)
shows a slight degradation in overall performance for both Ours and PixelNeRF.

\subsection{Quality Metrics}
Objective quality metrics such as Peak Signal-to-Noise Ratio (PSNR), Structural Similarity Index (SSIM), Learned Perceptual Image Patch Similarity (LPIPS), and DISTS (Deep Image Structure and Texture Similarity) are commonly used to assess image quality. PSNR measures image distortion based on differences in pixel intensity and provides a logarithmic scale to represent the error. SSIM evaluates structural similarity by considering luminance, contrast, and structure components. LPIPS uses deep learning techniques to create a perceptual similarity metric, which captures complex visual properties and aligns well with human perception. DISTS combines deep learning with a new architecture to measure structure and texture similarity, improving the perceptual relevance of the metric. These metrics are valuable in various image processing applications, helping quantify and optimize image quality.
For all evaluations, we provide the standard metrics for image quality, including PSNR, SSIM, and LPIPS. Additionally, we incorporate DISTS to represent the human perception of texture accurately. In this scenario, we adhere to the PixelNeRF protocol to maintain consistency with prior works.

\subsection{Datasets}
\subsubsection{The DTU MVS dataset} 
\begin{figure}[htb]
     \centering\setlength{\tabcolsep}{1pt}
     \captionsetup{justification=centering,singlelinecheck=false}
    \begin{tabular}{ccccc}
             \begin{subfigure}[b]{.15\linewidth}
         \centering
         \includegraphics[width=\textwidth]{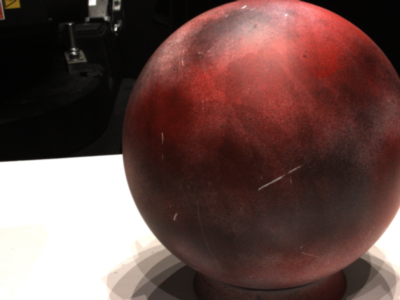}
         \setcounter{subfigure}{0}%
         \caption{scan8}
         \label{fig:scene8}
     \end{subfigure} 
     & 
    \begin{subfigure}[b]{.15\linewidth}
         \centering
         \includegraphics[width=\textwidth]{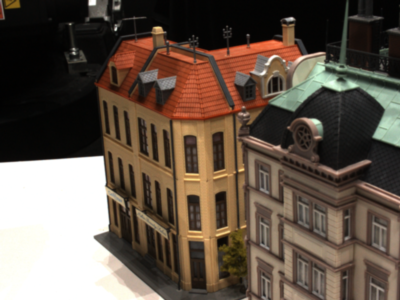}
         \setcounter{subfigure}{1}%
         \caption{scan21}
         \label{fig:scene21}
     \end{subfigure} 
     & 
    \begin{subfigure}[b]{.15\linewidth}
         \centering
         \includegraphics[width=\textwidth]{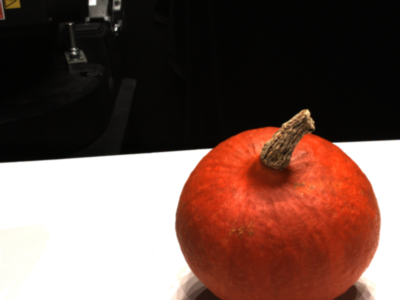}
         \setcounter{subfigure}{2}%
         \caption{scan30}
         \label{fig:scene30}
     \end{subfigure} 
     & 
     \begin{subfigure}[b]{.15\linewidth}
         \centering
         \includegraphics[width=\textwidth]{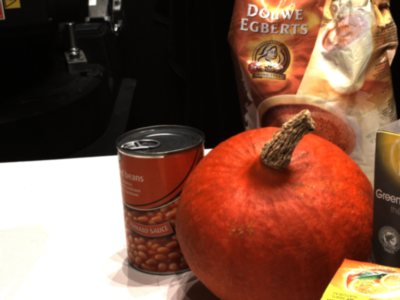}
         \setcounter{subfigure}{3}%
         \caption{scan31}
         \label{fig:scene31}
     \end{subfigure} 
     &
     \begin{subfigure}[b]{.15\linewidth}
         \centering
         \includegraphics[width=\textwidth]{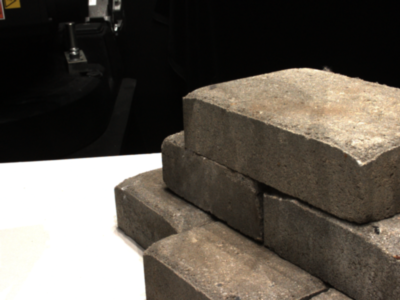}
         \setcounter{subfigure}{4}%
         \caption{scan34}
         \label{fig:scene8}
     \end{subfigure} 
     \\
     \begin{subfigure}[b]{.15\linewidth}
         \centering
         \includegraphics[width=\textwidth]{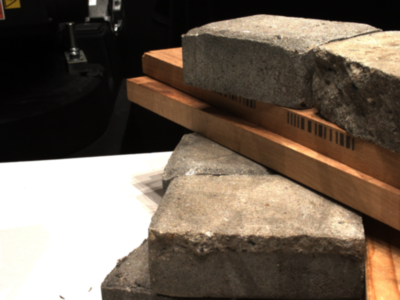}
         \setcounter{subfigure}{5}%
         \caption{scan38}
         \label{fig:scene38}
     \end{subfigure} 
     & 
    \begin{subfigure}[b]{.15\linewidth}
         \centering
         \includegraphics[width=\textwidth]{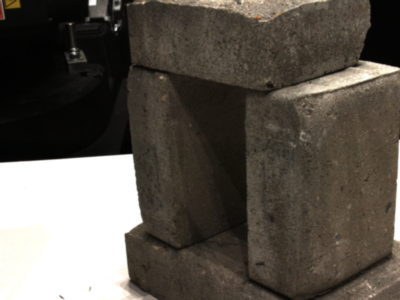}
         \setcounter{subfigure}{6}%
         \caption{scan40}
         \label{fig:scene40}
     \end{subfigure} 
     & 
    \begin{subfigure}[b]{.15\linewidth}
         \centering
         \includegraphics[width=\textwidth]{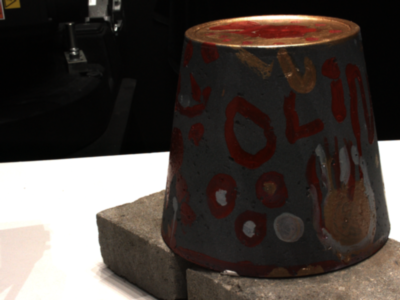}
         \setcounter{subfigure}{7}%
         \caption{scan41}
         \label{fig:scene41}
     \end{subfigure} 
    &
    \begin{subfigure}[b]{.15\linewidth}
         \centering
         \includegraphics[width=\textwidth]{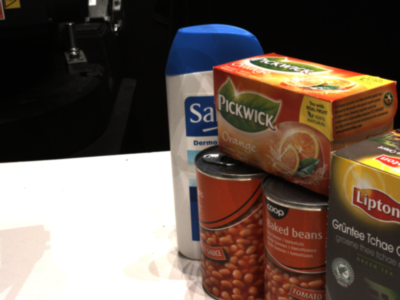}
         \setcounter{subfigure}{8}%
         \caption{scan45}
         \label{fig:scene45}
     \end{subfigure} 
     &
     \begin{subfigure}[b]{.15\linewidth}
         \centering
         \includegraphics[width=\textwidth]{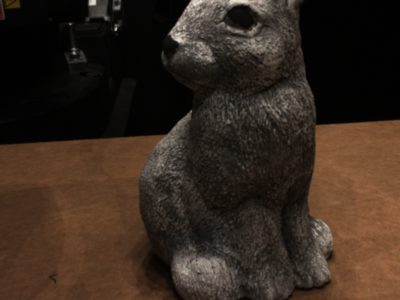}
         \setcounter{subfigure}{9}%
         \caption{scan55}
         \label{fig:scene55}
     \end{subfigure} 
     \\
    \begin{subfigure}[b]{.15\linewidth}
         \centering
         \includegraphics[width=\textwidth]{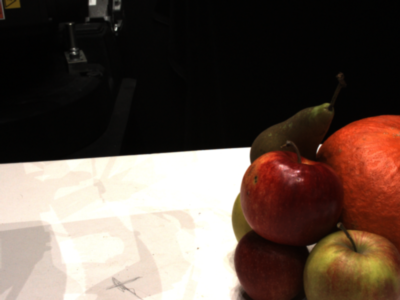}
         \setcounter{subfigure}{10}%
         \caption{scan63}
         \label{fig:scene63}
     \end{subfigure} 
     & 
    \begin{subfigure}[b]{.15\linewidth}
         \centering
         \includegraphics[width=\textwidth]{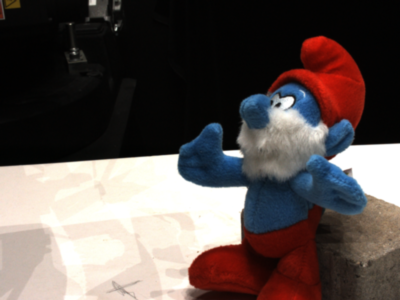}
         \setcounter{subfigure}{11}%
         \caption{scan82}
         \label{fig:scene82}
     \end{subfigure} 
     &
    \begin{subfigure}[b]{.15\linewidth}
         \centering
         \includegraphics[width=\textwidth]{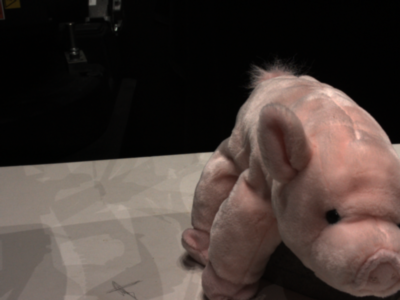}
         \setcounter{subfigure}{12}%
         \caption{scan103}
         \label{fig:scene45}
     \end{subfigure} 
     &
     \begin{subfigure}[b]{.15\linewidth}
         \centering
         \includegraphics[width=\textwidth]{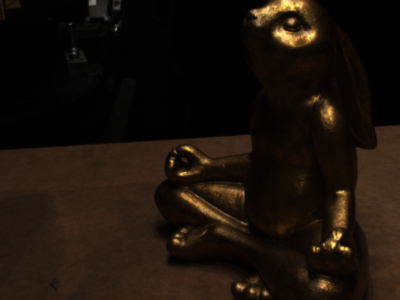}
         \setcounter{subfigure}{13}%
         \caption{scan110}
         \label{fig:scene110}
     \end{subfigure} 
     & 
    \begin{subfigure}[b]{.15\linewidth}
         \centering
         \includegraphics[width=\textwidth]{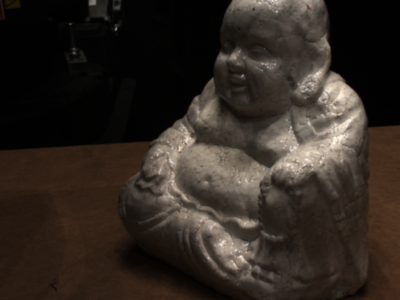}
         \setcounter{subfigure}{14}%
         \caption{scan114}
         \label{fig:scene114}
     \end{subfigure} 
\end{tabular}
        \caption{Examples from the DTU dataset for the selected 15 scenes.}
         \label{fig:example_dtu_datasets}
\end{figure}

comprises 124 scenes, including 22 scenes for testing, 79 for training, and 18 for validation. Each scene features a sophisticated central area and intricate background details. To capture these scenes, specific measures were taken to minimize photometric variations by fixing camera exposure settings, minimizing lighting changes, and avoiding moving objects. The examples of these scenes are illustrated in Fig.\ref{fig:example_dtu_datasets}.

\subsubsection{ShapeNet} 
\begin{figure}[htb]
     \centering\setlength{\tabcolsep}{1pt}
     \begin{center}
    \begin{tabular}{ccccccc}
      Input & \multicolumn{2}{c}{PixelNeRF} &  \multicolumn{2}{c}{Ours} & \multicolumn{2}{c}{Reference} 
           \\
    \begin{subfigure}[t]{0.13\linewidth}
         \centering
         \includegraphics[trim={1cm 0.55cm 1cm 1cm},clip,width=1\textwidth]{144d0880f61d813ef7b860bd772a37_gt_000064.png}
     \end{subfigure} 
     &
     \begin{subfigure}[t]{0.13\linewidth}
         \centering
         \includegraphics[trim={1cm 0.55cm 1cm 1cm},clip,width=1\textwidth]{17e916fc863540ee3def89b32cef8e45_gt_000000.png}
     \end{subfigure} 
     &
     \begin{subfigure}[t]{0.13\linewidth}
         \centering
         \includegraphics[trim={1cm 0.55cm 1cm 1cm},clip,width=1\textwidth]{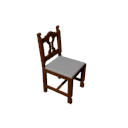}
     \end{subfigure} 
     &
      \begin{subfigure}[t]{0.13\linewidth}
         \centering
         \includegraphics[trim={1cm 0.55cm 1cm 1cm},clip,width=1\textwidth]{17e916fc863540ee3def89b32cef8e45_gt_000000.png}
     \end{subfigure}
     &
     \begin{subfigure}[t]{0.13\linewidth}
         \centering
         \includegraphics[trim={1cm 0.55cm 1cm 1cm},clip,width=1\textwidth]{17e916fc863540ee3def89b32cef8e45_gt_000086.png}
     \end{subfigure}
     &
     \begin{subfigure}[t]{0.13\linewidth}
         \centering
         \includegraphics[trim={1cm 0.55cm 1cm 1cm},clip,width=1\textwidth]{17e916fc863540ee3def89b32cef8e45_gt_000000.png}
     \end{subfigure} 
     &
     \begin{subfigure}[t]{0.13\linewidth}
         \centering
         \includegraphics[trim={1cm 0.55cm 1cm 1cm},clip,width=1\textwidth]{17e916fc863540ee3def89b32cef8e45_gt_000086.png}
     \end{subfigure} 
     \\
     \begin{subfigure}[t]{0.13\linewidth}
         \centering
         \includegraphics[trim={1cm 0.55cm 1cm 1cm},clip,width=1\textwidth]{144d0880f61d813ef7b860bd772a37_gt_000064.png}
     \end{subfigure} 
     &
     \begin{subfigure}[t]{0.13\linewidth}
         \centering
         \includegraphics[trim={1cm 0.55cm 1cm 1cm},clip,width=1\textwidth]{17e916fc863540ee3def89b32cef8e45_gt_000000.png}
     \end{subfigure} 
     &
     \begin{subfigure}[t]{0.13\linewidth}
         \centering
         \includegraphics[trim={1cm 0.55cm 1cm 1cm},clip,width=1\textwidth]{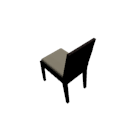}
     \end{subfigure} 
     &
      \begin{subfigure}[t]{0.13\linewidth}
         \centering
         \includegraphics[trim={1cm 0.55cm 1cm 1cm},clip,width=1\textwidth]{17e916fc863540ee3def89b32cef8e45_gt_000000.png}
     \end{subfigure}
     &
     \begin{subfigure}[t]{0.13\linewidth}
         \centering
         \includegraphics[trim={1cm 0.55cm 1cm 1cm},clip,width=1\textwidth]{195464ae11f6bfe1cba091e036bf65ed_gt_000122.png}
     \end{subfigure}
     &
     \begin{subfigure}[t]{0.13\linewidth}
         \centering
         \includegraphics[trim={1cm 0.55cm 1cm 1cm},clip,width=1\textwidth]{17e916fc863540ee3def89b32cef8e45_gt_000000.png}
     \end{subfigure} 
     &
     \begin{subfigure}[t]{0.13\linewidth}
         \centering
         \includegraphics[trim={1cm 0.55cm 1cm 1cm},clip,width=1\textwidth]{195464ae11f6bfe1cba091e036bf65ed_gt_000122.png}
     \end{subfigure} 
     \\
     \begin{subfigure}[t]{0.13\linewidth}
         \centering
         \includegraphics[trim={0.55cm 0.75cm 0.75cm 0.75cm},clip,width=1\textwidth]{144d0880f61d813ef7b860bd772a37_gt_000064.png}
     \end{subfigure} 
     &
     \begin{subfigure}[t]{0.13\linewidth}
         \centering
         \includegraphics[trim={0.55cm 0.75cm 0.75cm 0.75cm},clip,width=1\textwidth]{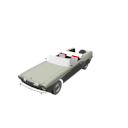}
     \end{subfigure} 
     &
     \begin{subfigure}[t]{0.13\linewidth}
         \centering
         \includegraphics[trim={0.55cm 0.75cm 0.5cm 0.75cm},clip,width=1\textwidth]{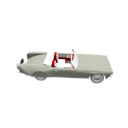}
     \end{subfigure} 
     &
      \begin{subfigure}[t]{0.13\linewidth}
         \centering
         \includegraphics[trim={0.55cm 0.75cm 0.75cm 0.75cm},clip,width=1\textwidth]{110ab054cfc351f46a2345809e2bb169_gt_000066.png}
     \end{subfigure}
     &
     \begin{subfigure}[t]{0.13\linewidth}
         \centering
         \includegraphics[trim={0.55cm 0.75cm 0.5cm 0.75cm},clip,width=1\textwidth]{110ab054cfc351f46a2345809e2bb169_gt_000092.png}
     \end{subfigure}
     &
     \begin{subfigure}[t]{0.13\linewidth}
         \centering
         \includegraphics[trim={0.55cm 0.75cm 0.75cm 0.75cm},clip,width=1\textwidth]{110ab054cfc351f46a2345809e2bb169_gt_000066.png}
     \end{subfigure} 
     &
     \begin{subfigure}[t]{0.13\linewidth}
         \centering
         \includegraphics[trim={0.55cm 0.75cm 0.5cm 0.75cm},clip,width=1\textwidth]{110ab054cfc351f46a2345809e2bb169_gt_000092.png}
     \end{subfigure} 
     \\
     \begin{subfigure}[t]{0.13\linewidth}
         \centering
         \includegraphics[trim={0.55cm 0.75cm 0.75cm 0.75cm},clip,width=1\textwidth]{144d0880f61d813ef7b860bd772a37_gt_000064.png}
     \end{subfigure} 
     &
     \begin{subfigure}[t]{0.13\linewidth}
         \centering
         \includegraphics[trim={0.55cm 0.75cm 0.75cm 0.75cm},clip,width=1\textwidth]{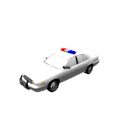}
     \end{subfigure} 
     &
     \begin{subfigure}[t]{0.13\linewidth}
         \centering
         \includegraphics[trim={0.55cm 0.75cm 0.5cm 0.75cm},clip,width=1\textwidth]{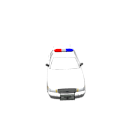}
     \end{subfigure} 
     &
      \begin{subfigure}[t]{0.13\linewidth}
         \centering
         \includegraphics[trim={0.55cm 0.75cm 0.75cm 0.75cm},clip,width=1\textwidth]{11a96098620b2ebac2f9fb5458a091d1_gt_000063.png}
     \end{subfigure}
     &
     \begin{subfigure}[t]{0.13\linewidth}
         \centering
         \includegraphics[trim={0.55cm 0.75cm 0.5cm 0.75cm},clip,width=1\textwidth]{11a96098620b2ebac2f9fb5458a091d1_gt_000074.png}
     \end{subfigure}
     &
     \begin{subfigure}[t]{0.13\linewidth}
         \centering
         \includegraphics[trim={0.55cm 0.75cm 0.75cm 0.75cm},clip,width=1\textwidth]{11a96098620b2ebac2f9fb5458a091d1_gt_000063.png}
     \end{subfigure} 
     &
     \begin{subfigure}[t]{0.13\linewidth}
         \centering
         \includegraphics[trim={0.55cm 0.75cm 0.5cm 0.75cm},clip,width=1\textwidth]{11a96098620b2ebac2f9fb5458a091d1_gt_000074.png}
     \end{subfigure} 
\end{tabular}
        \caption{Category-specific one-shot neural rendering benchmark. We trained two separate models for chairs and cars and compared them to the baseline.}
        \label{fig:1v_cars_chairs_results}
        \end{center}
\end{figure}

\begin{figure}[htb]
     \centering\setlength{\tabcolsep}{1pt}
    \begin{tabular}{cccccccc}
       \multicolumn{2}{c}{Input} & \multicolumn{2}{c}{PixelNeRF} &  \multicolumn{2}{c}{Ours} & \multicolumn{2}{c}{Reference} 
       \\
    \begin{subfigure}[t]{0.11\linewidth}
         \centering
         \includegraphics[trim={1cm 0.55cm 1cm 1cm},clip,width=1\textwidth]{144d0880f61d813ef7b860bd772a37_gt_000064.png}
     \end{subfigure}
     &
      \begin{subfigure}[t]{0.11\linewidth}
         \centering
         \includegraphics[trim={1cm 0.55cm 1cm 1cm},clip,width=1\textwidth]{144d0880f61d813ef7b860bd772a37_gt_000104.png}
     \end{subfigure} 
     &
     \begin{subfigure}[t]{0.11\linewidth}
         \centering
         \includegraphics[trim={1cm 0.55cm 1cm 1cm},clip,width=1\textwidth]{158a95b4da25aa1fa37f3fc191551700_gt_000172.png}
     \end{subfigure} 
     &
     \begin{subfigure}[t]{0.11\linewidth}
         \centering
         \includegraphics[trim={1cm 0.55cm 1cm 1cm},clip,width=1\textwidth]{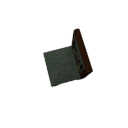}
     \end{subfigure} 
     &
      \begin{subfigure}[t]{0.11\linewidth}
         \centering
         \includegraphics[trim={1cm 0.55cm 1cm 1cm},clip,width=1\textwidth]{158a95b4da25aa1fa37f3fc191551700_gt_000172.png}
     \end{subfigure}
     &
     \begin{subfigure}[t]{0.11\linewidth}
         \centering
         \includegraphics[trim={1cm 0.55cm 1cm 1cm},clip,width=1\textwidth]{17aeeadccf0e560e274b862d3a151946_gt_000214.png}
     \end{subfigure}
     &
     \begin{subfigure}[t]{0.11\linewidth}
         \centering
         \includegraphics[trim={1cm 0.55cm 1cm 1cm},clip,width=1\textwidth]{158a95b4da25aa1fa37f3fc191551700_gt_000172.png}
     \end{subfigure} 
     &
     \begin{subfigure}[t]{0.11\linewidth}
         \centering
         \includegraphics[trim={1cm 0.55cm 1cm 1cm},clip,width=1\textwidth]{17aeeadccf0e560e274b862d3a151946_gt_000214.png}
     \end{subfigure} 
     \\
      \begin{subfigure}[t]{0.11\linewidth}
         \centering
         \includegraphics[trim={1cm 0.55cm 1cm 1cm},clip,width=1\textwidth]{144d0880f61d813ef7b860bd772a37_gt_000064.png}
     \end{subfigure} 
     &
     \begin{subfigure}[t]{0.11\linewidth}
         \centering
         \includegraphics[trim={1cm 0.55cm 1cm 1cm},clip,width=1\textwidth]{144d0880f61d813ef7b860bd772a37_gt_000104.png}
     \end{subfigure} 
     &
     \begin{subfigure}[t]{0.11\linewidth}
         \centering
         \includegraphics[trim={1cm 0.55cm 1cm 1cm},clip,width=1\textwidth]{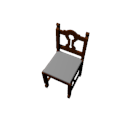}
     \end{subfigure} 
     &
     \begin{subfigure}[t]{0.11\linewidth}
         \centering
         \includegraphics[trim={1cm 0.55cm 1cm 1cm},clip,width=1\textwidth]{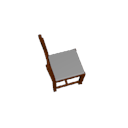}
     \end{subfigure} 
     &
      \begin{subfigure}[t]{0.11\linewidth}
         \centering
         \includegraphics[trim={1cm 0.55cm 1cm 1cm},clip,width=1\textwidth]{17e916fc863540ee3def89b32cef8e45_gt_000146.png}
     \end{subfigure}
     &
     \begin{subfigure}[t]{0.11\linewidth}
         \centering
         \includegraphics[trim={1cm 0.55cm 1cm 1cm},clip,width=1\textwidth]{17e916fc863540ee3def89b32cef8e45_gt_000175.png}
     \end{subfigure}
     &
     \begin{subfigure}[t]{0.11\linewidth}
         \centering
         \includegraphics[trim={1cm 0.55cm 1cm 1cm},clip,width=1\textwidth]{17e916fc863540ee3def89b32cef8e45_gt_000146.png}
     \end{subfigure} 
     &
     \begin{subfigure}[t]{0.11\linewidth}
         \centering
         \includegraphics[trim={1cm 0.55cm 1cm 1cm},clip,width=1\textwidth]{17e916fc863540ee3def89b32cef8e45_gt_000175.png}
     \end{subfigure} 
     \\
     \begin{subfigure}[t]{0.11\linewidth}
         \centering
         \includegraphics[trim={0.55cm 0.75cm 0.5cm 0.75cm},clip,width=1\textwidth]{144d0880f61d813ef7b860bd772a37_gt_000064.png}
     \end{subfigure} 
     &
     \begin{subfigure}[t]{0.11\linewidth}
         \centering
         \includegraphics[trim={0.55cm 0.75cm 0.5cm 0.75cm},clip,width=1\textwidth]{144d0880f61d813ef7b860bd772a37_gt_000104.png}
     \end{subfigure} 
     &
     \begin{subfigure}[t]{0.11\linewidth}
         \centering
         \includegraphics[trim={0.55cm 0.75cm 0.5cm 0.75cm},clip,width=1\textwidth]{144d0880f61d813ef7b860bd772a37_gt_000065.png}
     \end{subfigure} 
     &
     \begin{subfigure}[t]{0.11\linewidth}
         \centering
         \includegraphics[trim={0.55cm 0.75cm 0.5cm 0.75cm},clip,width=1\textwidth]{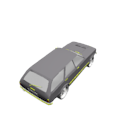}
     \end{subfigure} 
     &
      \begin{subfigure}[t]{0.11\linewidth}
         \centering
         \includegraphics[trim={0.55cm 0.75cm 0.5cm 0.75cm},clip,width=1\textwidth]{144d0880f61d813ef7b860bd772a37_gt_000065.png}
     \end{subfigure}
     &
     \begin{subfigure}[t]{0.11\linewidth}
         \centering
         \includegraphics[trim={0.55cm 0.75cm 0.5cm 0.75cm},clip,width=1\textwidth]{156d4748560997c9a848f24544821b25_gt_000105.png}
     \end{subfigure}
     &
     \begin{subfigure}[t]{0.11\linewidth}
         \centering
         \includegraphics[trim={0.55cm 0.75cm 0.5cm 0.75cm},clip,width=1\textwidth]{144d0880f61d813ef7b860bd772a37_gt_000065.png}
     \end{subfigure} 
     &
     \begin{subfigure}[t]{0.11\linewidth}
         \centering
         \includegraphics[trim={0.55cm 0.75cm 0.5cm 0.75cm},clip,width=1\textwidth]{144d0880f61d813ef7b860bd772a37_gt_000104.png}
     \end{subfigure} 
     \\
     \begin{subfigure}[t]{0.11\linewidth}
         \centering
         \includegraphics[trim={0.55cm 0.75cm 0.5cm 0.75cm},clip,width=1\textwidth]{144d0880f61d813ef7b860bd772a37_gt_000064.png}
     \end{subfigure} 
     &
     \begin{subfigure}[t]{0.11\linewidth}
         \centering
         \includegraphics[trim={0.55cm 0.75cm 0.5cm 0.75cm},clip,width=1\textwidth]{144d0880f61d813ef7b860bd772a37_gt_000104.png}
     \end{subfigure} 
     &
     \begin{subfigure}[t]{0.11\linewidth}
         \centering
         \includegraphics[trim={0.55cm 0.75cm 0.5cm 0.75cm},clip,width=1\textwidth]{158a95b4da25aa1fa37f3fc191551700_gt_000172.png}
     \end{subfigure} 
     &
     \begin{subfigure}[t]{0.11\linewidth}
         \centering
         \includegraphics[trim={0.55cm 0.75cm 0.5cm 0.75cm},clip,width=1\textwidth]{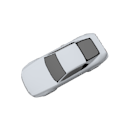}
     \end{subfigure} 
     &
      \begin{subfigure}[t]{0.11\linewidth}
         \centering
         \includegraphics[trim={0.55cm 0.75cm 0.5cm 0.75cm},clip,width=1\textwidth]{158a95b4da25aa1fa37f3fc191551700_gt_000172.png}
     \end{subfigure}
     &
     \begin{subfigure}[t]{0.11\linewidth}
         \centering
         \includegraphics[trim={0.55cm 0.75cm 0.5cm 0.75cm},clip,width=1\textwidth]{158a95b4da25aa1fa37f3fc191551700_gt_000217.png}
     \end{subfigure}
     &
     \begin{subfigure}[t]{0.11\linewidth}
         \centering
         \includegraphics[trim={0.55cm 0.75cm 0.5cm 0.75cm},clip,width=1\textwidth]{144d0880f61d813ef7b860bd772a37_gt_000104.png}
     \end{subfigure} 
     &
     \begin{subfigure}[t]{0.11\linewidth}
         \centering
         \includegraphics[trim={0.55cm 0.75cm 0.5cm 0.75cm},clip,width=1\textwidth]{158a95b4da25aa1fa37f3fc191551700_gt_000172.png}
     \end{subfigure} 
     
     \\
          \begin{subfigure}[t]{0.11\linewidth}
         \centering
         \includegraphics[trim={0.55cm 1cm 0.5cm 1cm},clip,width=1\textwidth]{144d0880f61d813ef7b860bd772a37_gt_000064.png}
     \end{subfigure} 
     &
     \begin{subfigure}[t]{0.11\linewidth}
         \centering
         \includegraphics[trim={0.55cm 1cm 0.5cm 1cm},clip,width=1\textwidth]{144d0880f61d813ef7b860bd772a37_gt_000104.png}
     \end{subfigure} 
     &
     \begin{subfigure}[t]{0.11\linewidth}
         \centering
         \includegraphics[trim={0.55cm 1cm 0.5cm 1cm},clip,width=1\textwidth]{11a96098620b2ebac2f9fb5458a091d1_gt_000060.png}
     \end{subfigure} 
     &
     \begin{subfigure}[t]{0.11\linewidth}
         \centering
         \includegraphics[trim={0.55cm 1cm 0.5cm 1cm},clip,width=1\textwidth]{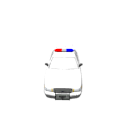}
     \end{subfigure} 
     &
      \begin{subfigure}[t]{0.11\linewidth}
         \centering
         \includegraphics[trim={0.55cm 1cm 0.5cm 1cm},clip,width=1\textwidth]{11a96098620b2ebac2f9fb5458a091d1_gt_000060.png}
     \end{subfigure}
     &
     \begin{subfigure}[t]{0.11\linewidth}
         \centering
         \includegraphics[trim={0.55cm 1cm 0.5cm 1cm},clip,width=1\textwidth]{11a96098620b2ebac2f9fb5458a091d1_gt_000073.png}
     \end{subfigure}
     &
     \begin{subfigure}[t]{0.11\linewidth}
         \centering
         \includegraphics[trim={0.55cm 1cm 0.5cm 1cm},clip,width=1\textwidth]{11a96098620b2ebac2f9fb5458a091d1_gt_000060.png}
     \end{subfigure} 
     &
     \begin{subfigure}[t]{0.11\linewidth}
         \centering
         \includegraphics[trim={0.55cm 1cm 0.5cm 1cm},clip,width=1\textwidth]{11a96098620b2ebac2f9fb5458a091d1_gt_000073.png}
     \end{subfigure} 
\end{tabular}
        \caption{Category-specific few-shot neural rendering benchmark. We trained two separate models for chairs and cars and compared them to the baseline. The input is two view images.}
        \label{fig:2v_cars_chairs_results}
\end{figure}

is a vast collection of 3D models with over 51,000 unique shapes belonging to 55 categories, such as chairs, cars, and airplanes. The dataset provides comprehensive information for each object, including category labels, part segmentations, and alignment information. This makes it an excellent resource for training and evaluating algorithms in 3D object recognition, reconstruction, and other related tasks. The input column in Fig. \ref{fig:1v_cars_chairs_results} and Fig. \ref{fig:2v_cars_chairs_results} are examples of the ShapeNet dataset.

\end{document}